\documentclass[10pt,twocolumn,letterpaper]{article}

\pdfoutput=1

\usepackage[usenames,dvipsnames]{color}
\usepackage{booktabs}
\usepackage{cvpr}
\usepackage{times}
\usepackage{epsfig}
\usepackage{graphicx}
\usepackage{amsmath}
\usepackage{amssymb}
\usepackage{subfigure} 
\usepackage{algorithm}
\usepackage{multirow}
\usepackage[english]{babel}
\usepackage{tabularx}
\usepackage{array}
\usepackage{caption}
\usepackage{algorithmic}
\usepackage{enumitem}

\usepackage{color}

\newcommand{\comment}[1]{}

\newcommand{\note}[1]{{\color{red}{\noindent\framebox{\begin{minipage}{8cm}{#1}\end{minipage}}\\ }}}
\newcommand{\ans}[1]{{\color{blue}{\noindent\framebox{\begin{minipage}{8cm}{#1}\end{minipage}}\\ }}}

\newcommand{\vincent}[1]{{\color{OliveGreen}{#1}}}
\newcommand{\vincentrmk}[1]{{\color{OliveGreen}{\bf #1}}}
\newcommand{\confirmed}[1]{{\color{black}{#1}}}
\newcommand{\artem}[1]{{\color{black}{#1}}}

\newcommand{\mycaption}[1]{\vspace{-0em}\caption{#1}\vspace{0em}}
\newcommand{\myparagraph}[1]{\vspace{-0em}\paragraph{#1}}

\newcommand{\stcb}[0]{st-cube}
\newcommand{\Stcb}[0]{St-cube}

\newcommand{\halfimsz}{1.6in}
\newcommand{\dbimsz}{0.31in}
\newcommand{\mdsz}{0.35in}
\newcommand{\imsz}{0.60in}

\newcommand{\fig}[1]{Fig.~\ref{fig:#1}}

\newcommand{\tbl}[1]{Table~\ref{tbl:#1}}
\newcommand{\eqt}[1]{Eq.~\ref{eq:#1}}
\newcommand{\alg}[1]{Algorithm~\ref{alg:#1}}
\newcolumntype{M}[1]{>{\centering\arraybackslash}m{#1}}
\newcolumntype{P}[1]{>{\centering\arraybackslash}p{#1}}

\newcommand{\mynote}[1]{\renewcommand{\thefootnote}{\alph{footnote}}\footnotemark[#1]}

\cvprfinalcopy 

\def\cvprPaperID{2000} 

\begin{document}

\newcommand{\question}[1]{\emph{\bf #1}}
\newcommand{\answer}[1]{$\rightarrow$ \emph{\bf #1}}

\title{Flying Objects Detection from a Single Moving Camera}

\author{
	$\qquad$ Artem Rozantsev\mynote{1} $\qquad$ Vincent Lepetit  \mynote{1}\hspace{2pt}$\hspace{3pt} ^{,}$\mynote{2} $\qquad$ Pascal Fua\mynote{1} \\
\mynote{1}{\hspace{2pt} Computer Vision Laboratory, \'{E}cole Polytechnique F\'{e}d\'{e}rale de Lausanne (EPFL)}\\
\mynote{2}{\hspace{2pt} Institute for Computer Graphics and Vision, Graz University of Technology} \\
{\tt\small \{artem.rozantsev, pascal.fua\}@epfl.ch, lepetit@icg.tugraz.at} \\
}

\maketitle


\begin{abstract}

We propose  an approach to detect  flying objects such as  UAVs and aircrafts
when they  occupy  a small  portion of the  field of  view, possibly moving
against complex backgrounds, and are filmed by a camera that itself moves.

Solving such a  difficult problem requires combining both  appearance and motion
cues.   To   this  end  we   propose  a  regression-based  approach   to  motion
stabilization  of  local image  patches  that  allows  us to  achieve  effective
classification on spatio-temporal image cubes and outperform state-of-the-art techniques.

As the problem is relatively new, we collected two challenging datasets for UAVs and Aircrafts, which can be used as benchmarks for flying objects detection and vision-guided collision avoidance.

\comment{We use two  very challenging  datasets to demostrate that our  approach outperforms state-of-the-art techniques.}

\end{abstract}



\section{Introduction}

We are headed for a world in which  the skies are occupied not only by birds and
planes but also by unmanned drones ranging from relatively large Unmanned Aerial
Vehicles  (UAVs)  to  much  smaller  consumer  ones.   Some  of  these  will  be
instrumented and able to communicate with each other to avoid collisions but not
all. Therefore, the ability to use  inexpensive and light sensors such as cameras
for collision-avoidance purposes will become increasingly important.

\begin{figure}
\centering
\begin{tabular}{cc}
\includegraphics[width = \halfimsz]{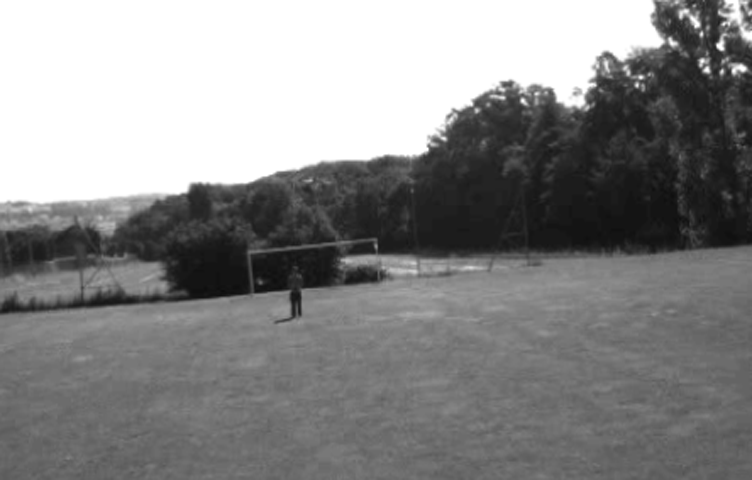} &
\hspace{-0.3cm}\includegraphics[width = \halfimsz]{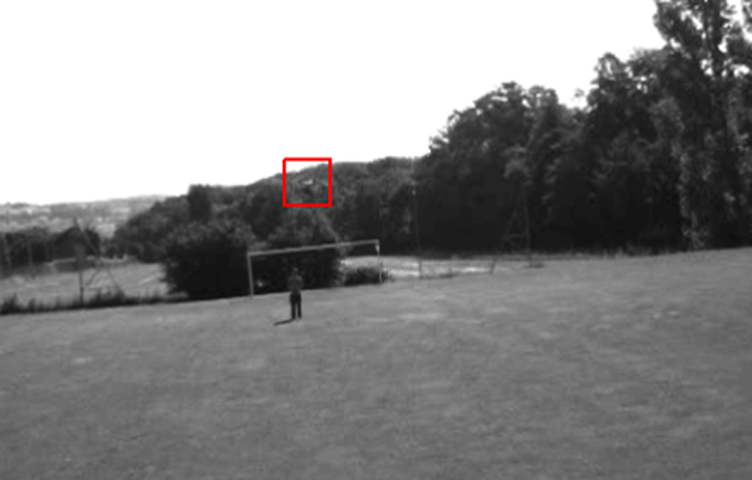}\\
\end{tabular}
\mycaption{Detecting a small drone against  a complex moving background. (Left) It
  is  almost invisible  to  the human  eye  and  hard to  detect  from a  single
  image. (Right) Yet, our algorithm can find it by using motion clues.}
\label{fig:drone}
\end{figure}

This problem  has been  tackled successfully in  the automotive  world and
there are now  commercial products~\cite{IntelligentDrive,Mobileeye} designed to
sense and  avoid both pedestrians and  other cars. In the world of flying machines most of the progress is achieved in the accurate position estimation and navigation from single or multiple cameras~\cite{Conte08,Martinez11,Meier11,Hane11,weiss13,lynen13,Forster14}, while not so much is done in the field of visual-guided collision avoidance~\cite{Zsedrovits11}. On the other hand, it is not possible to simply extend the algorithms used for pedestrian and automobile detection to the world of aircrafts and drones, as flying object detection poses some unique challenges:
\begin{itemize}[topsep=3pt, partopsep=5pt]
\setlength\itemsep{5pt}
\setlength{\parskip}{0pt}

\item The environment is fully 3D dimensional, which makes the motions more complex.
  
\item Flying objects have very diverse shapes and can be seen against either the
  ground or  the sky, which produces  complex and changing backgrounds, as shown
  in \fig{drone}.
  
 \item Given the speeds involved, potentially dangerous objects must be detected when they  are still far away,  which means they may still be very small  in the images.

\end{itemize}
As  a  result, motion  cues  become  crucial  for  detection. However,  they  are difficult to exploit when the images are acquired by a moving camera and feature backgrounds that are difficult to stabilize because they are non-planar and fast changing. Furthermore,  since there can  be other  moving objects in  the scene, for example, the  person  in \fig{drone}, motion  by itself  is  not enough  and appearance   must  also   be   taken  into   account.    In  these   situations, state-of-the-art  techniques  that  rely  on either  image  flow  or  background stabilization lose much of their effectiveness.

In this  paper, we detect  whether an object of  interest is present and  constitutes a  danger by classifying  3D descriptors  computed from spatio-temporal image  cubes. We will refer to them as \stcb{}s.  These   \stcb{}s  are   formed  by   stacking motion-stabilized  image windows  over several  consecutive frames,  which gives more  information than  using a  single image.   What makes  this approach  both practical   and    effective   is   a    regression-based   motion-stabilization algorithm. Unlike  those that rely  on optical  flow, it remains  effective even when the shape of  the  object to  be  detected  is blurry  or  barely visible,  as illustrated by \fig{stabilize}.

\Stcb{}s of image intensities  have   been  routinely used, for   action   recognition purposes~\cite{Laptev05b,Weinland10} using  a single fixed camera.  In contrast, most current  detection algorithms  work on a  single frame, or integrate the information from two of them, which might not be consecutive, by taking into account optical flow from one frame to another.  Our approach  can therefore be seen as a \comment{combination of these two strands of research} way to combine both the appearance and motion information to achieve effective detection in a very challenging context.



\section{Related work}
\label{sec:related}

\comment{\note{I hope the last sentence of the paragraph is true ;-)}

\ans{It is almost true. There is this work for pedestrians~\cite{Park13}, where they extended this to using information from not only consecutive frames, however the features that they used were just difference between two frames, so in terms of features - yes only two frames were involved, but generally their approach can use information not only from the previous frame, but for example 2-3 frames before.}

}
Approaches  to  detecting moving  objects  can  be  classified into  three  main
categories, those that rely on appearance  in individual frames, those that rely
primarily on motion  information across frames, and those that  combine the two.
We  briefly review  all  three types  in this section.  In the  results  section, we  will demonstrate that we can outperform state-of-the-art representatives of each.

{\bf Appearance-based  methods} rely  on Machine Learning  and have  proved to be
powerful  even in  the  presence  of complex  lighting  variations or  cluttered
background.     They     are    typically     based    on     Deformable    Part
Models~(DPM)~\cite{Felzenszwalb10},   Convolutional  Neural   Networks~(CNN)~\cite{Serre07} and Random Forests~\cite{Bosch07}. We will evaluate our approach in comparison with all of these methods and the another, which relies on an Aggregate Channel Features~(ACF)~\cite{Dollar09b}, as it is widely considered to be among the best.

\comment{and Boosting approaches with Haar-like features. Among the latter, the one that  relies on an Aggregate Channel Features detector~(ACF)~\cite{Dollar09b} is widely considered to be one of the best.}

\comment{
\note{No decision forests?}
\ans{added}
}

However,  they work  best when  the target  objects are  sufficiently large  and
clearly  visible in  individual  images, which  is  often not  the  case in  our
applications.  For example, in the  image of Fig.~\ref{fig:drone}, the object is
small and it is almost impossible to make out from the background without motion
cues.

{\bf Motion-based approaches} can themselves  be subdivided into two subclasses.
The first  comprises those that  rely on background  subtraction~\cite{Oliver00,SeungJong12,bgslibrary13}  and
detect objects as  groups of pixels that are different  from the background. The
second  includes  those   that  depend  on  optical   flow  between  consecutive
images~\cite{Brox11,Lucas81}. Background subtraction  works best when the  camera is static or its motion is small  enough to be easily compensated for, which  is not the case for the  on-board camera of a  fast moving vehicle. Flow-based  methods are more reliable in such  situations but are critically dependent on  the quality of the flow  vectors, which  tends to  be low  when the  target objects  are small  and blurry.

{\bf Hybrid approaches}  combine information about object  appearance and motion
patterns and are  therefore closest in spirit to what  we propose.  For example,
in~\cite{Walk10}, histograms of flow vectors are used as features in conjunction
with  more  standard appearance  features  and  fed  to a  statistical  learning
method. This approach was refined in~\cite{Park13} by first aligning the patches to compensate for  motion and then using  the differences of frames  that may or may not  be consecutive  as additional  features.  The  alignment relies  on the Lucas-Kanade  optical flow  algorithm~\cite{Lucas81}.   The resulting  algorithm works  very  well   for  pedestrian  detection  and  outperforms   most  of  the single-frame ones. However when the target  objects become smaller and harder to see, the  flow estimates become  unreliable and  this approach, like  the purely flow-based ones, becomes less effective.



\section{Approach}
\label{sec:approach}

\begin{figure*}[ht!]
\centering
\begin{tabular}{cccccccccccccccc}
\toprule
\multicolumn{8}{c}{\bf UAVs} & \multicolumn{8}{c}{\bf Aircrafts} \\
\multicolumn{4}{c}{\scriptsize{Uniform background}} &
\multicolumn{4}{c}{\scriptsize{Very noisy background}} & 
\multicolumn{4}{c}{\scriptsize{Non-uniform background}} & \multicolumn{4}{c}{\scriptsize{Noisy background}}\\
\midrule
&&&&&&\multicolumn{4}{c}{No motion compensation} &&&&&&\\
\cmidrule(r){7-10} 
\includegraphics[width = \dbimsz]{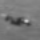} & 
  \hspace{-0.3cm}\includegraphics[width = \dbimsz]{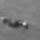} & 
  \hspace{-0.3cm}\includegraphics[width = \dbimsz]{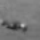} & 
  \hspace{-0.3cm}\includegraphics[width = \dbimsz]{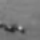} &
\includegraphics[width = \dbimsz]{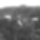} & 
  \hspace{-0.3cm}\includegraphics[width = \dbimsz]{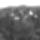} & 
  \hspace{-0.3cm}\includegraphics[width = \dbimsz]{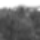} & 
  \hspace{-0.3cm}\includegraphics[width = \dbimsz]{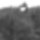} &
  \includegraphics[width = \dbimsz]{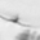} & 
  \hspace{-0.3cm}\includegraphics[width = \dbimsz]{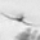} & 
  \hspace{-0.3cm}\includegraphics[width = \dbimsz]{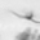} & 
  \hspace{-0.3cm}\includegraphics[width = \dbimsz]{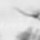} & 
  \includegraphics[width = \dbimsz]{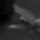} & 
  \hspace{-0.3cm}\includegraphics[width = \dbimsz]{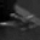} & 
  \hspace{-0.3cm}\includegraphics[width = \dbimsz]{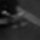} & 
  \hspace{-0.3cm}\includegraphics[width = \dbimsz]{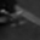} \\
  \multicolumn{4}{c}{
  	\begin{tabular}{ccc}
  		\includegraphics[height = \dbimsz]{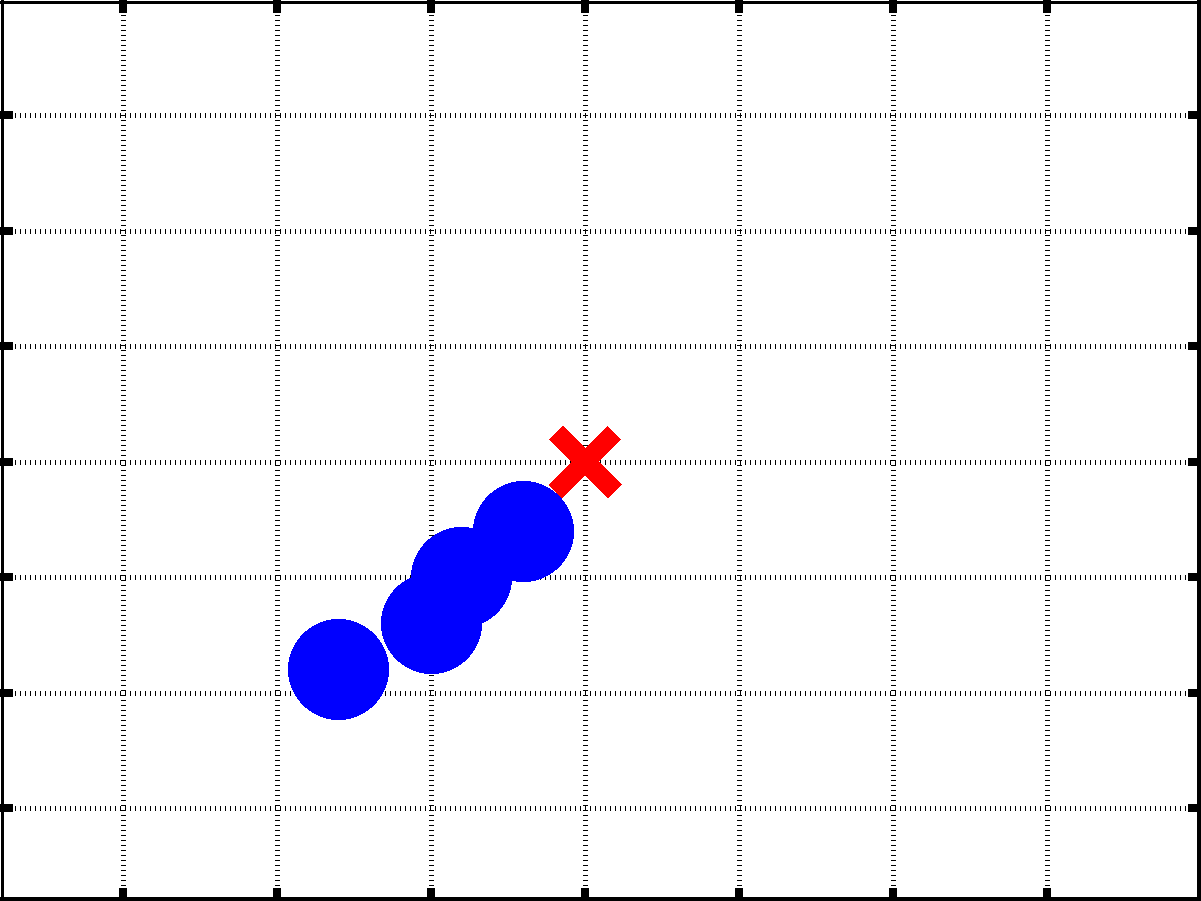} & 
  		\hspace{-0.3cm}\includegraphics[height = \dbimsz]{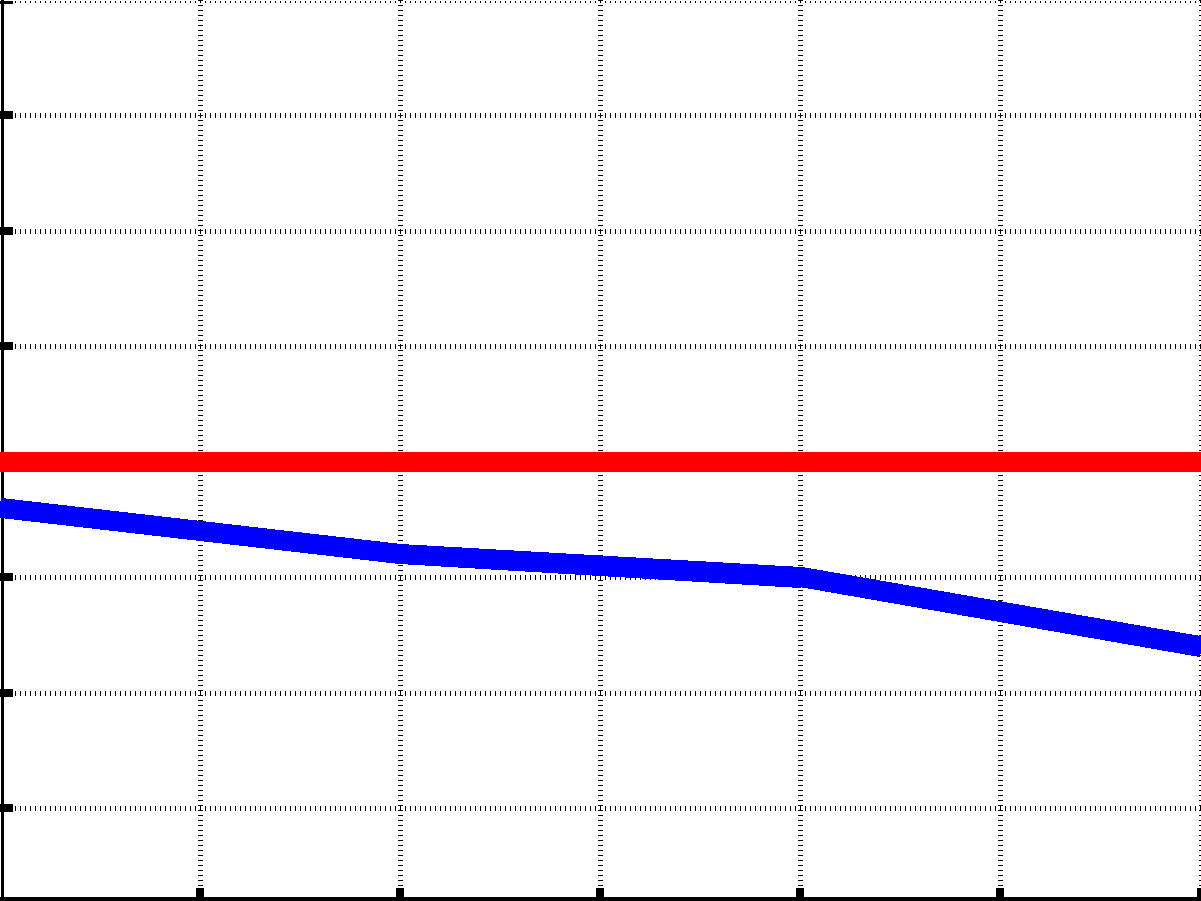} & 
  		\hspace{-0.3cm}\includegraphics[height = \dbimsz]{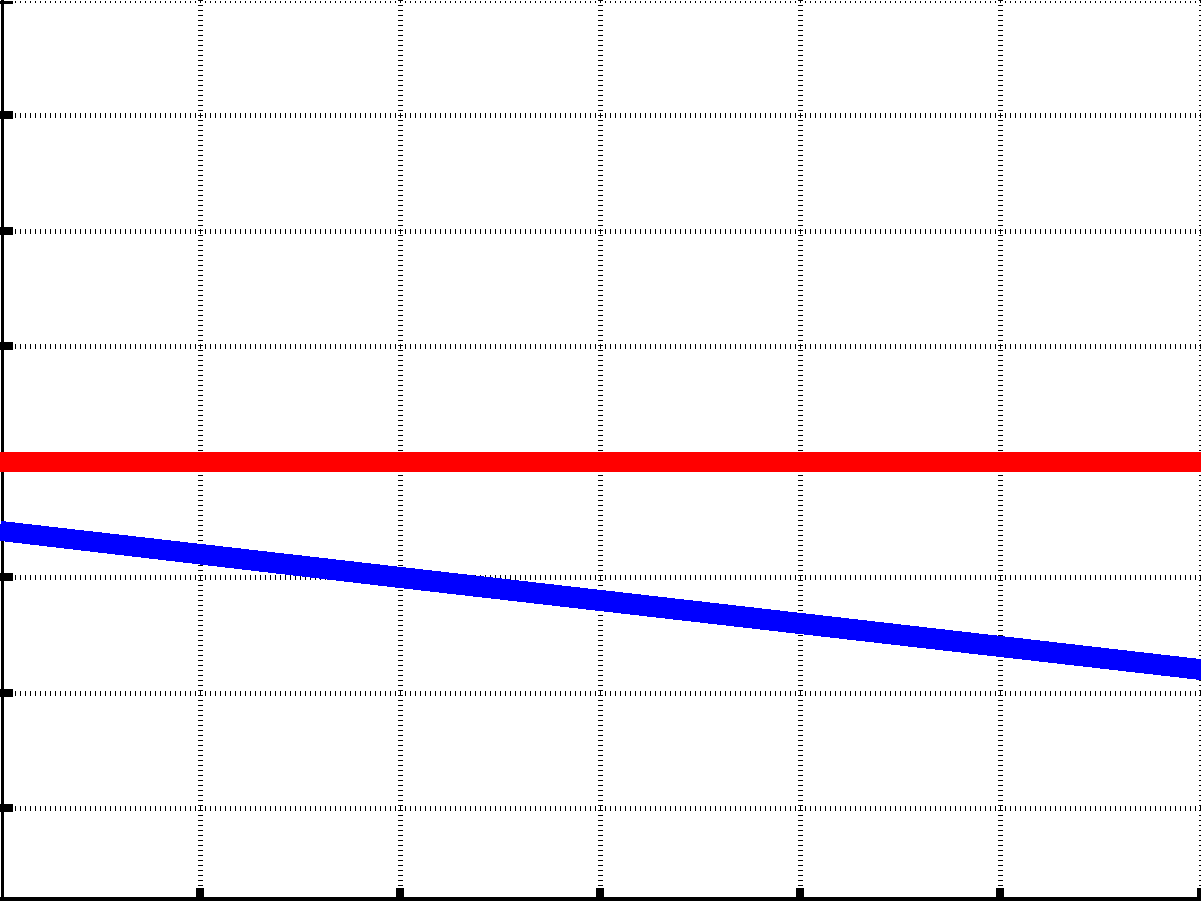} \\   
  	\end{tabular}
  }&
  \multicolumn{4}{c}{
  	\begin{tabular}{ccc}
  		\includegraphics[height = \dbimsz]{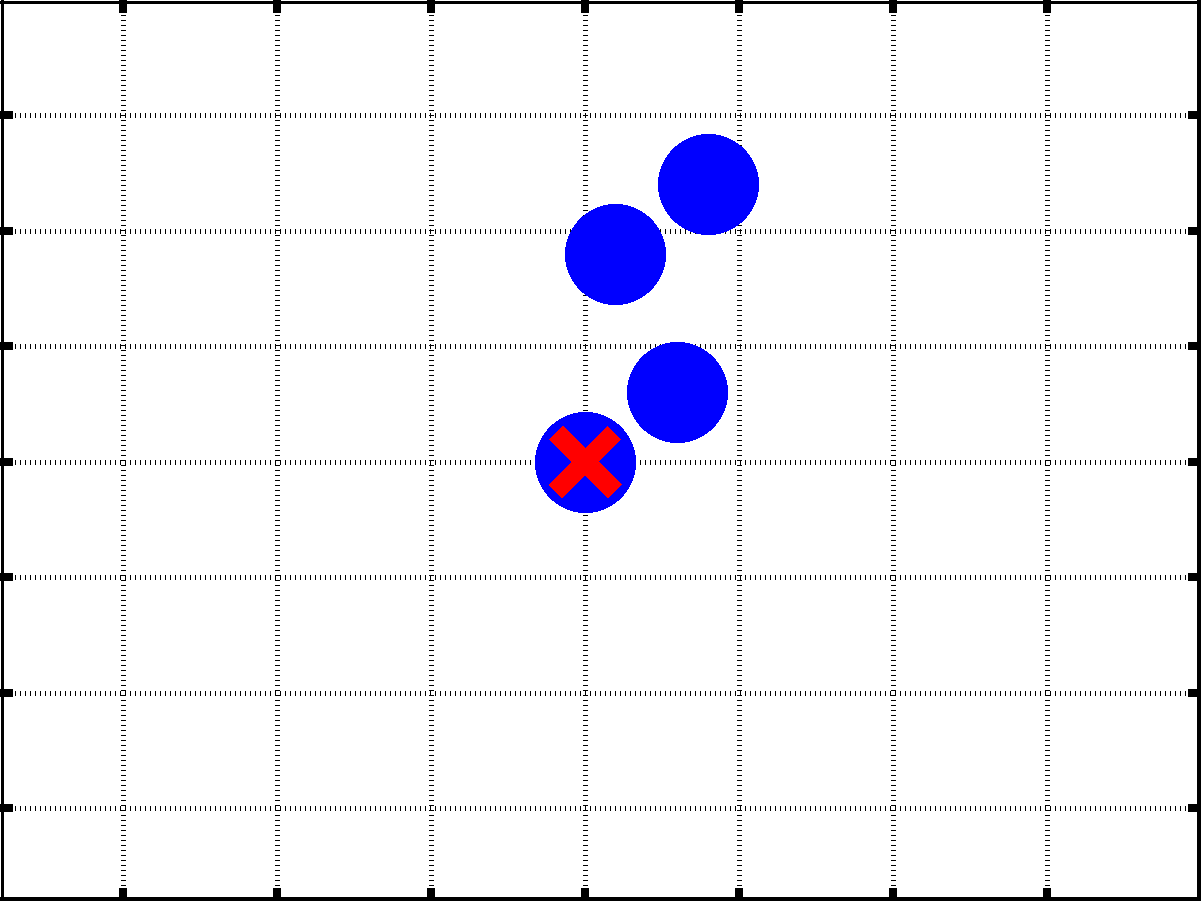} & 
  		\hspace{-0.3cm}\includegraphics[height = \dbimsz]{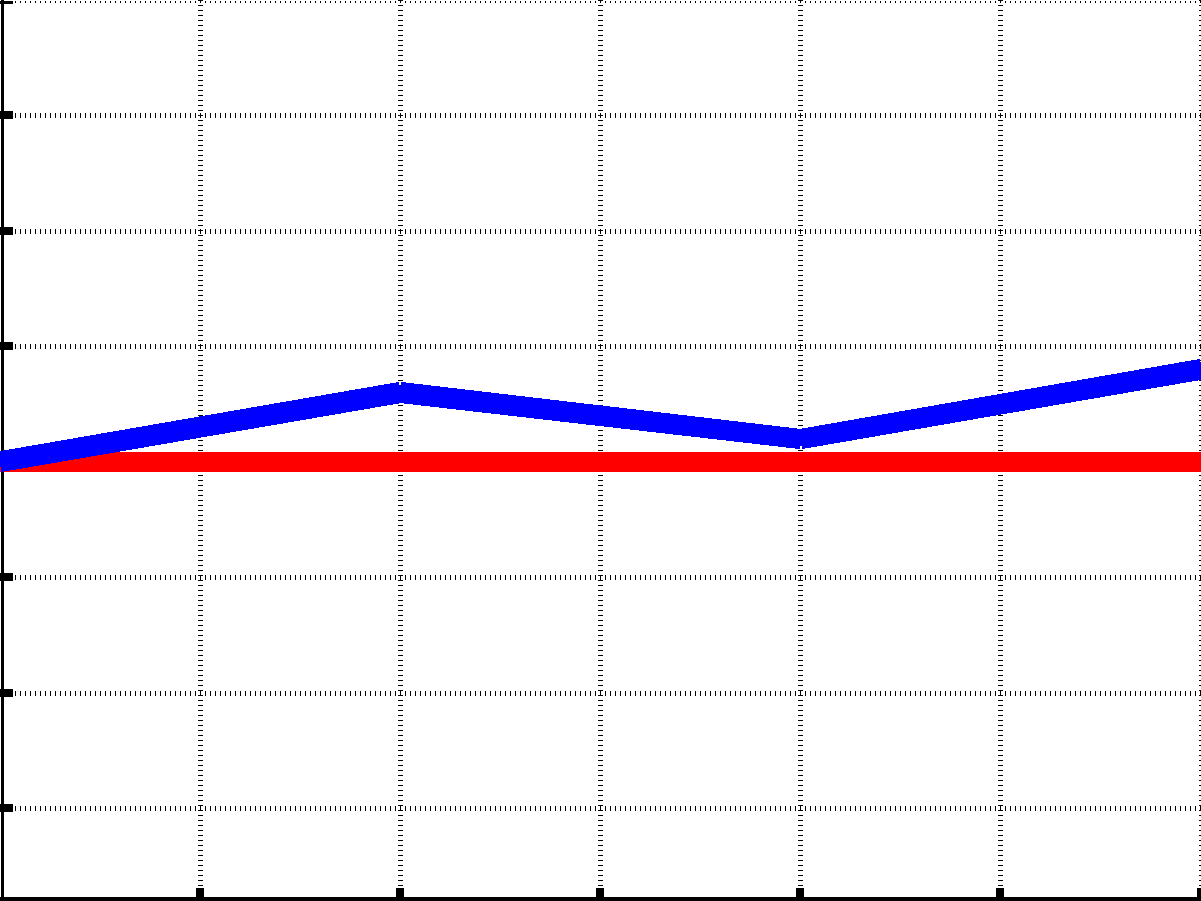} & 
  		\hspace{-0.3cm}\includegraphics[height = \dbimsz]{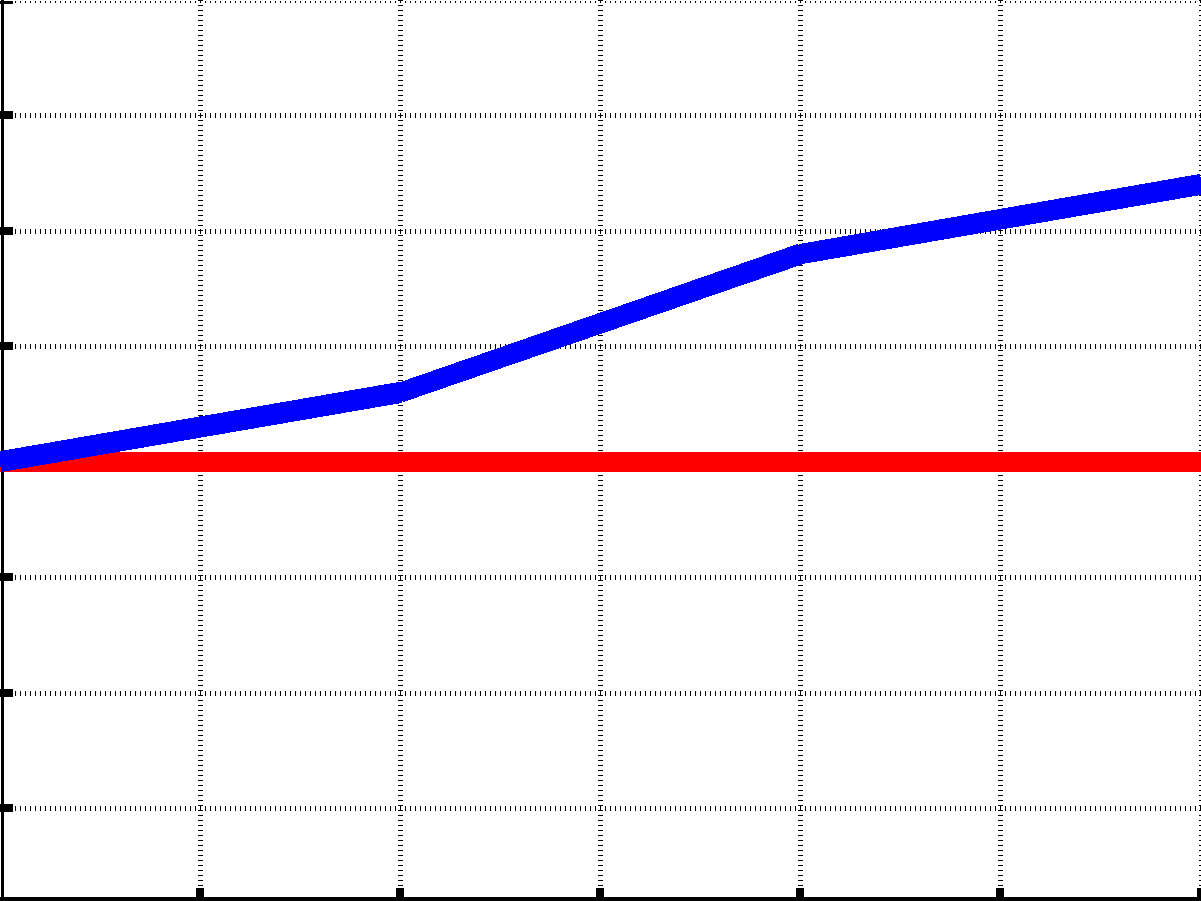} \\
  	\end{tabular}
  } &
  \multicolumn{4}{c}{
  	\begin{tabular}{ccc}
  		\includegraphics[height = \dbimsz]{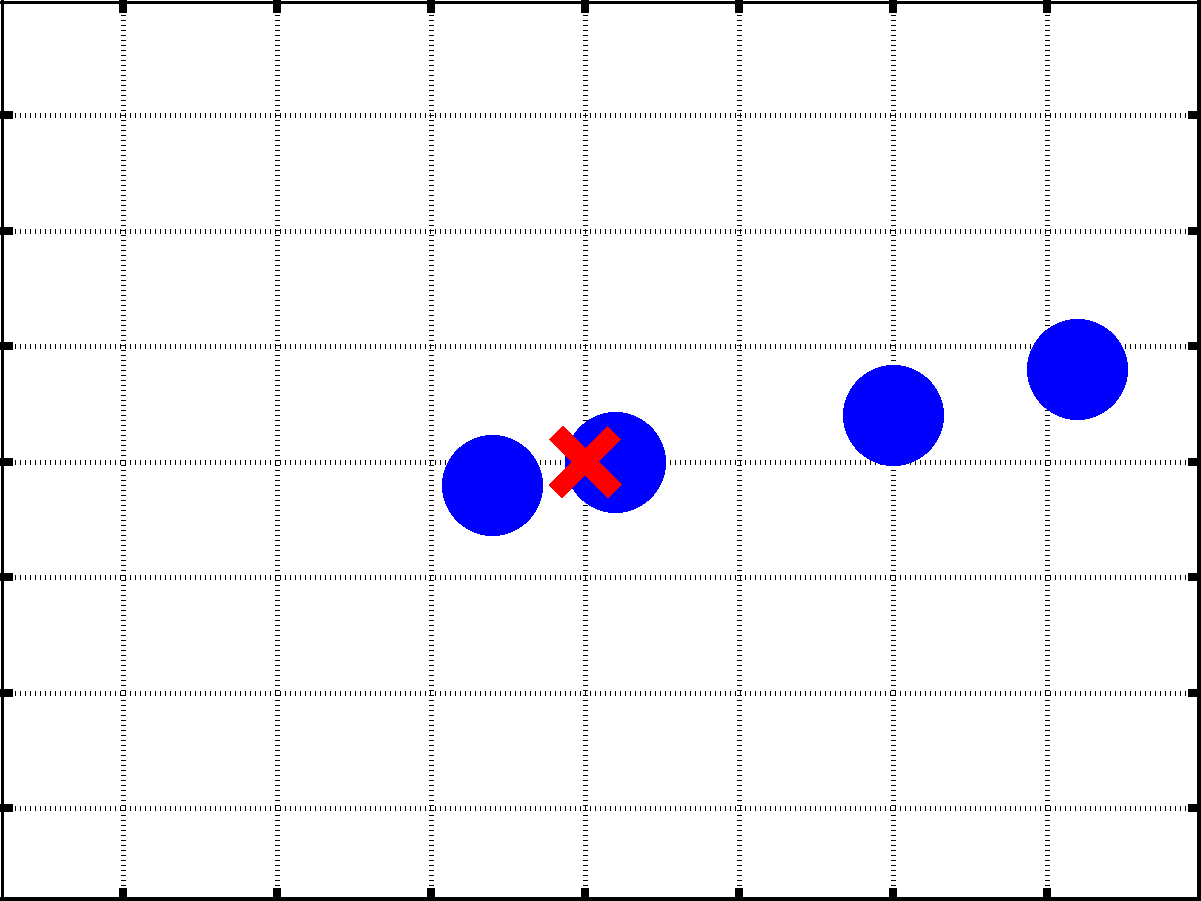} & 
  		\hspace{-0.3cm}\includegraphics[height = \dbimsz]{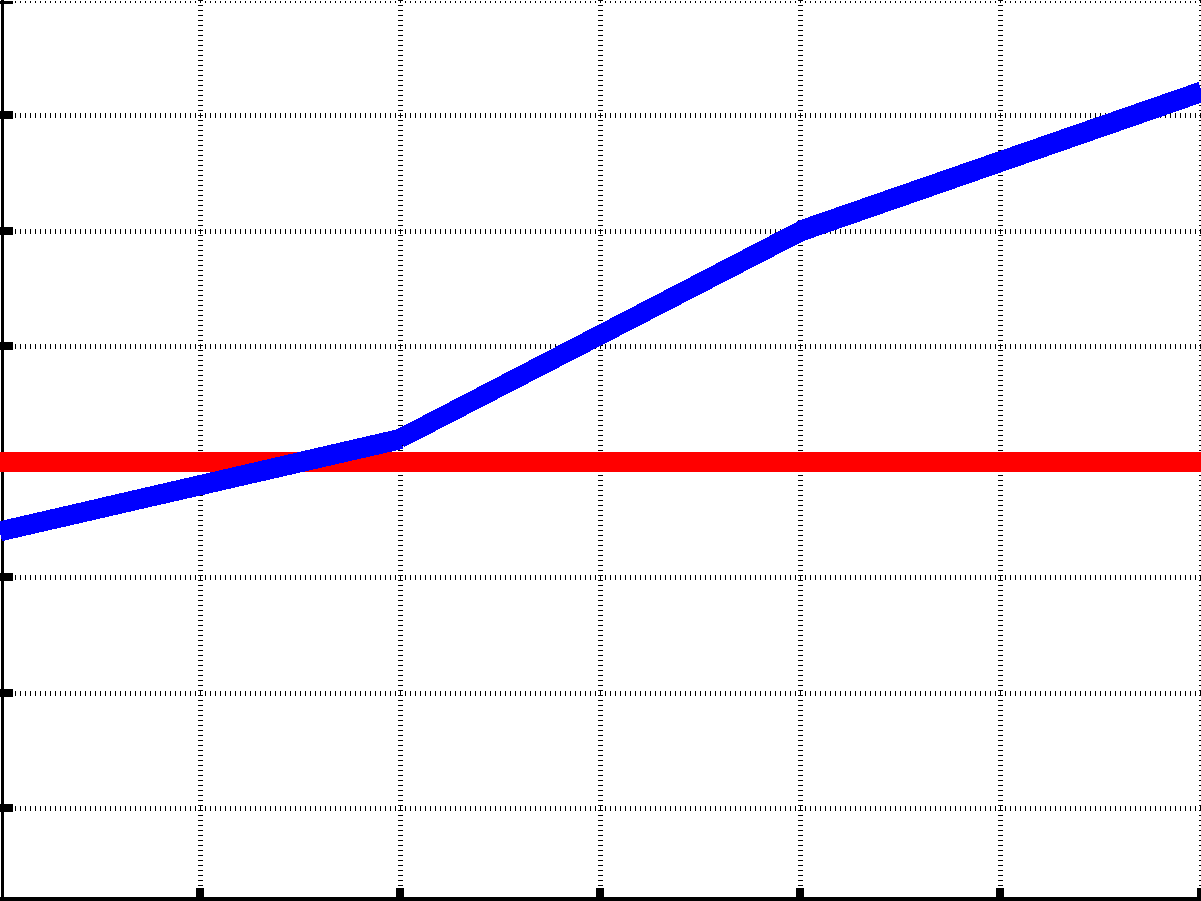} & 
  		\hspace{-0.3cm}\includegraphics[height = \dbimsz]{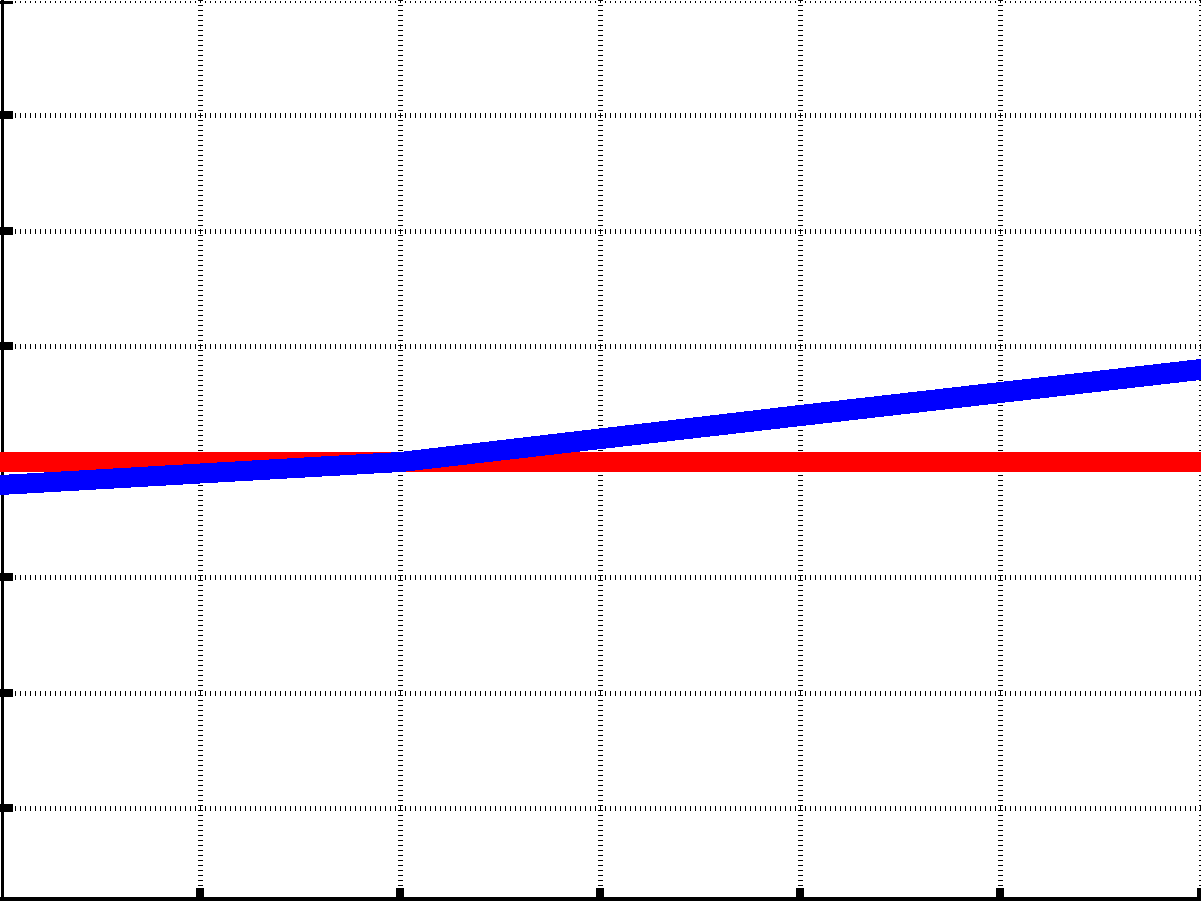} \\   
  	\end{tabular}
  }&
  \multicolumn{4}{c}{
  	\begin{tabular}{ccc}
  		\includegraphics[height = \dbimsz]{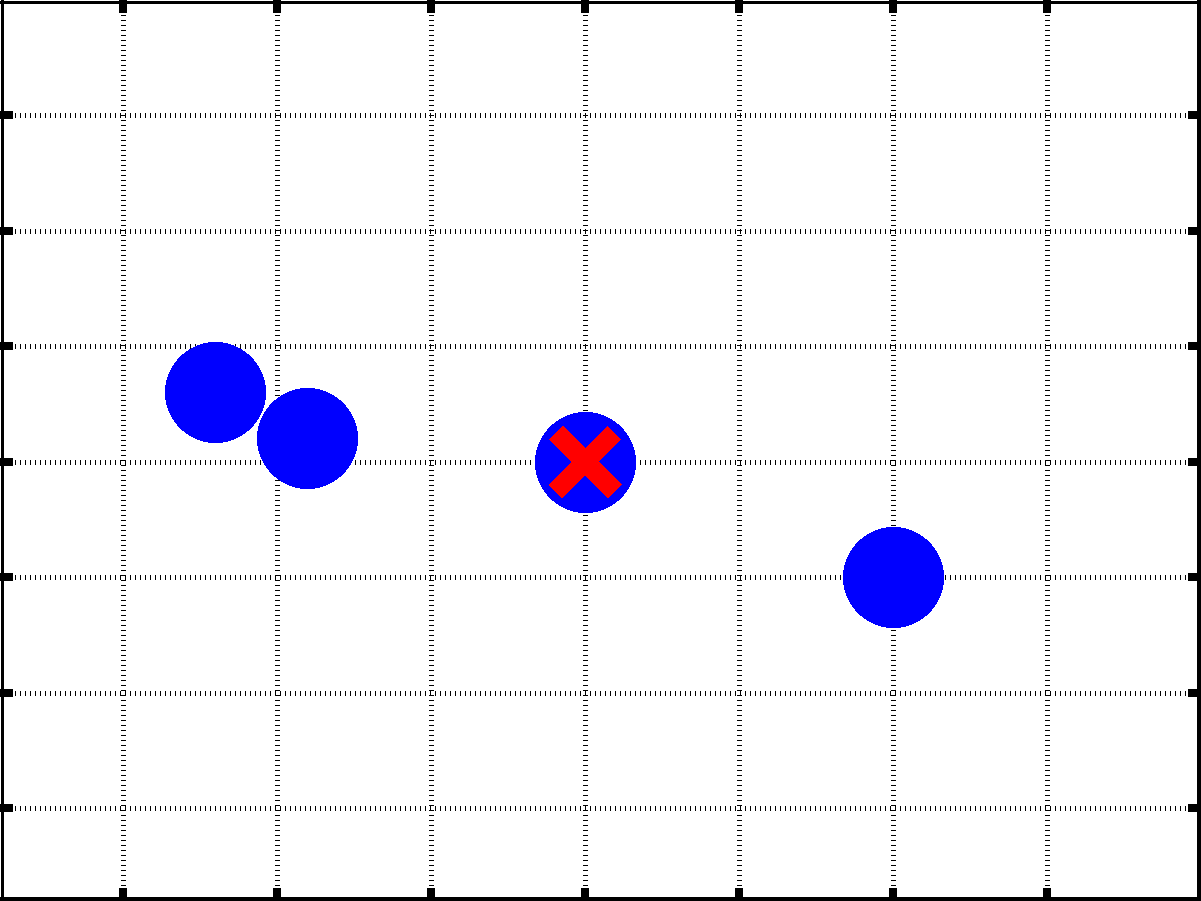} & 
  		\hspace{-0.3cm}\includegraphics[height = \dbimsz]{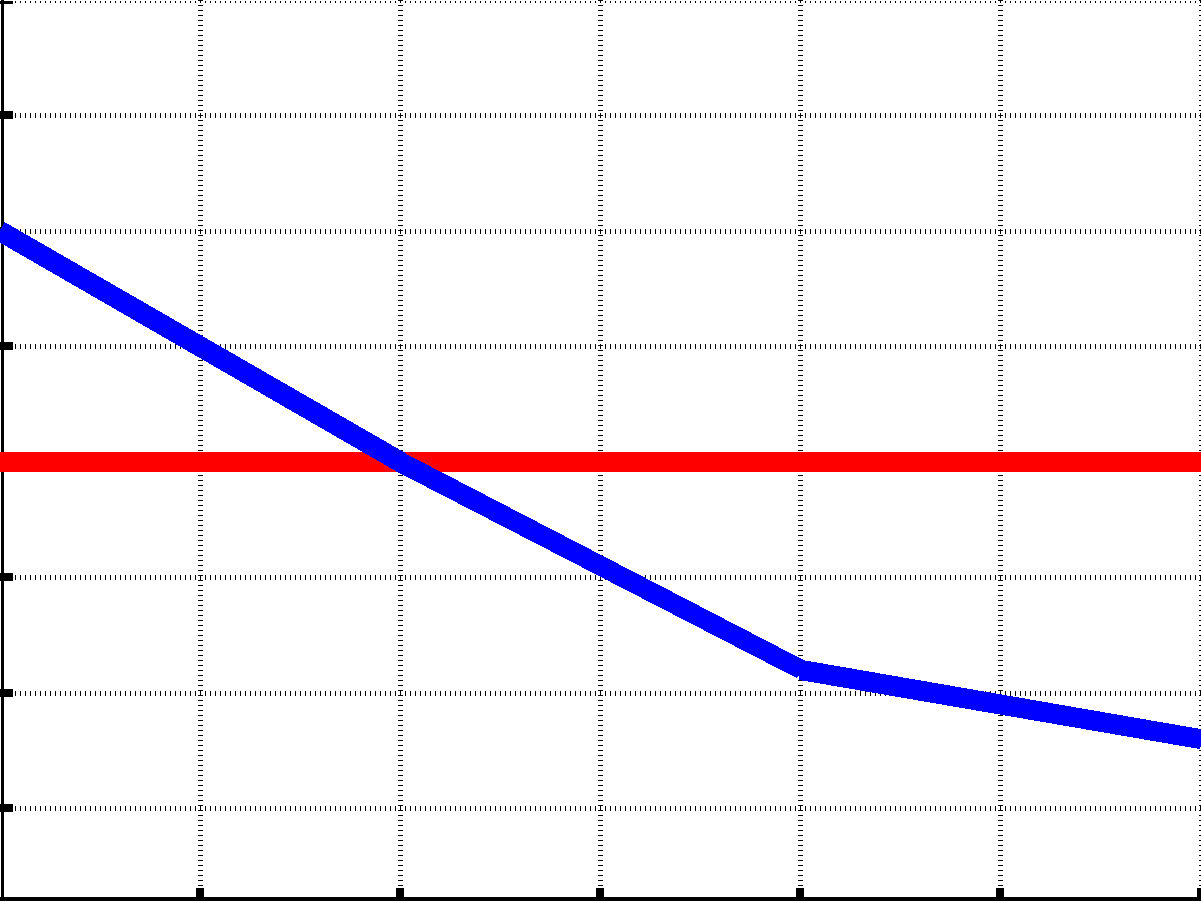} & 
  		\hspace{-0.3cm}\includegraphics[height = \dbimsz]{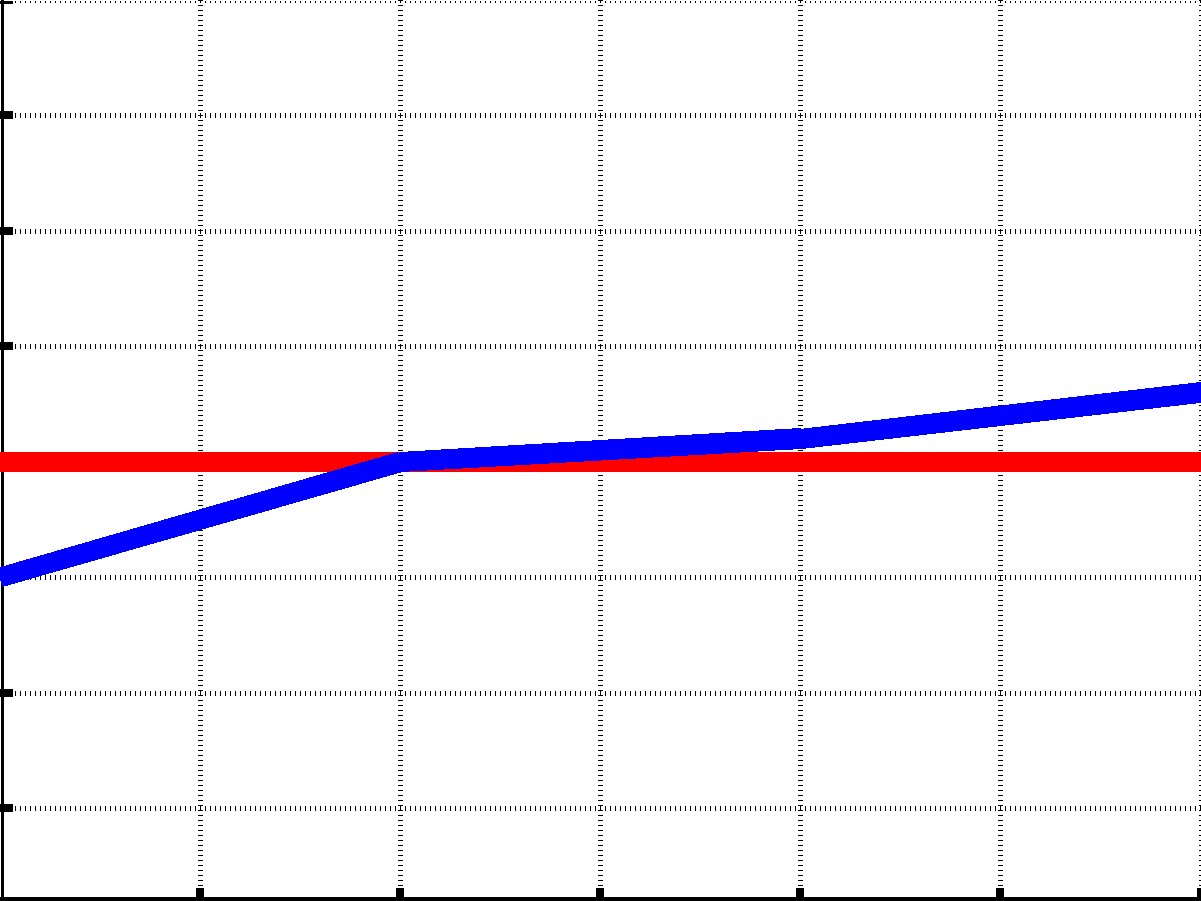} \\   
  	\end{tabular}
  } \\ 
\midrule
&&&&&&\multicolumn{4}{c}{Lucas-Kanade optical flow} &&&&&&\\
\cmidrule(r){7-10} 
\includegraphics[width = \dbimsz]{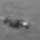} & 
  \hspace{-0.3cm}\includegraphics[width = \dbimsz]{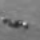} & 
  \hspace{-0.3cm}\includegraphics[width = \dbimsz]{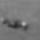} & 
  \hspace{-0.3cm}\includegraphics[width = \dbimsz]{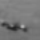} &
\includegraphics[width = \dbimsz]{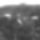} & 
  \hspace{-0.3cm}\includegraphics[width = \dbimsz]{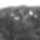} & 
  \hspace{-0.3cm}\includegraphics[width = \dbimsz]{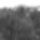} & 
  \hspace{-0.3cm}\includegraphics[width = \dbimsz]{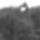} &
  \includegraphics[width = \dbimsz]{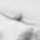} & 
  \hspace{-0.3cm}\includegraphics[width = \dbimsz]{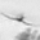} & 
  \hspace{-0.3cm}\includegraphics[width = \dbimsz]{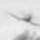} & 
  \hspace{-0.3cm}\includegraphics[width = \dbimsz]{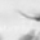} & 
  \includegraphics[width = \dbimsz]{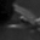} & 
  \hspace{-0.3cm}\includegraphics[width = \dbimsz]{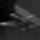} & 
  \hspace{-0.3cm}\includegraphics[width = \dbimsz]{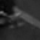} & 
  \hspace{-0.3cm}\includegraphics[width = \dbimsz]{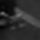} \\
  \multicolumn{4}{c}{
  \begin{tabular}{ccc}
\includegraphics[height = \dbimsz]{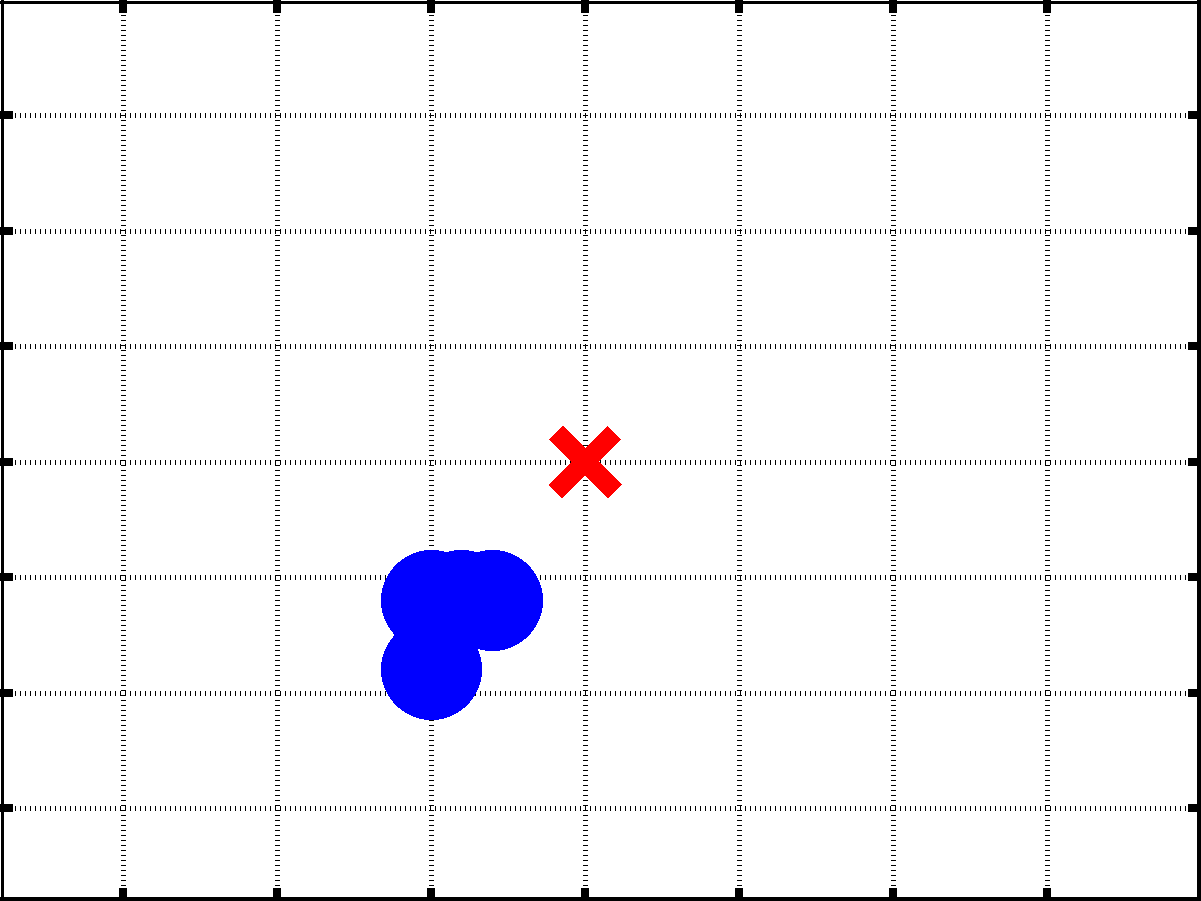} & 
  \hspace{-0.3cm}\includegraphics[height = \dbimsz]{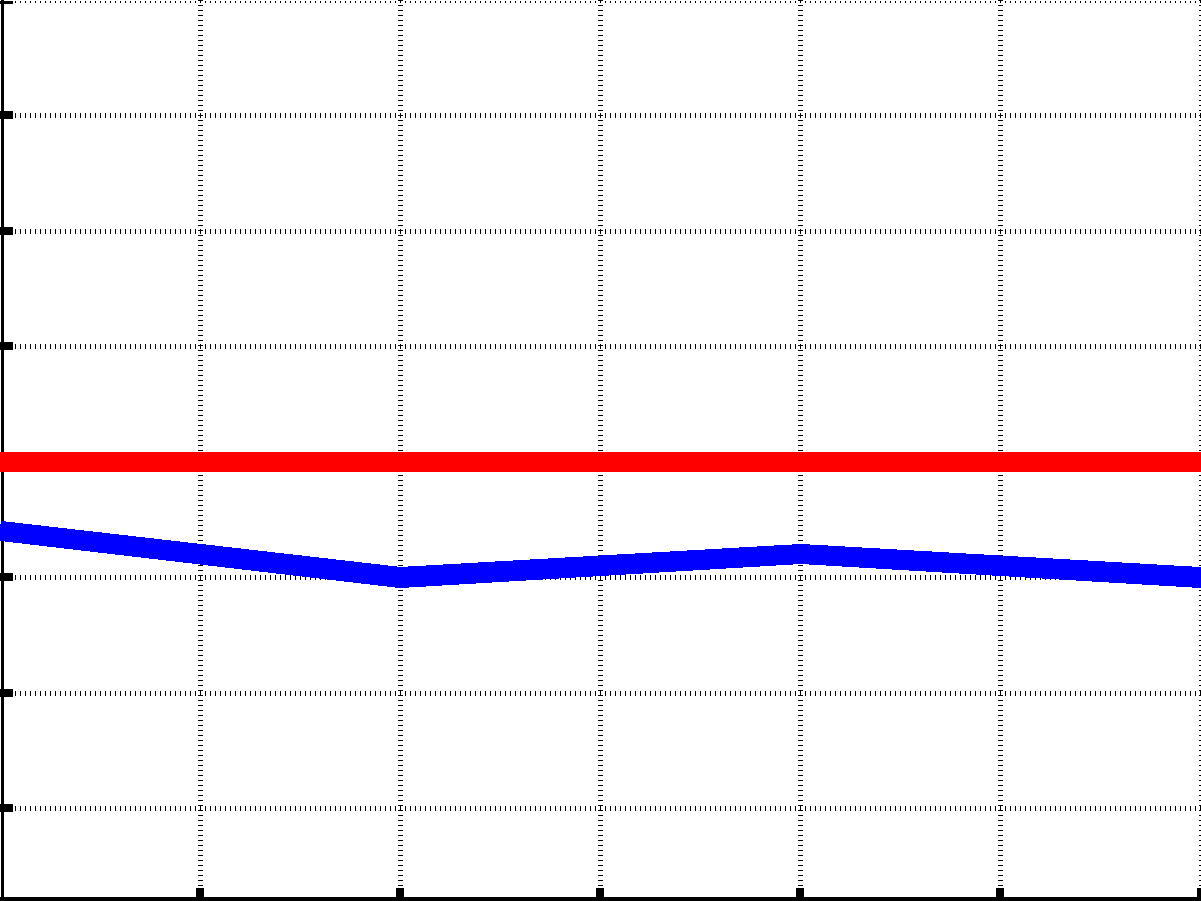} & 
  \hspace{-0.3cm}\includegraphics[height = \dbimsz]{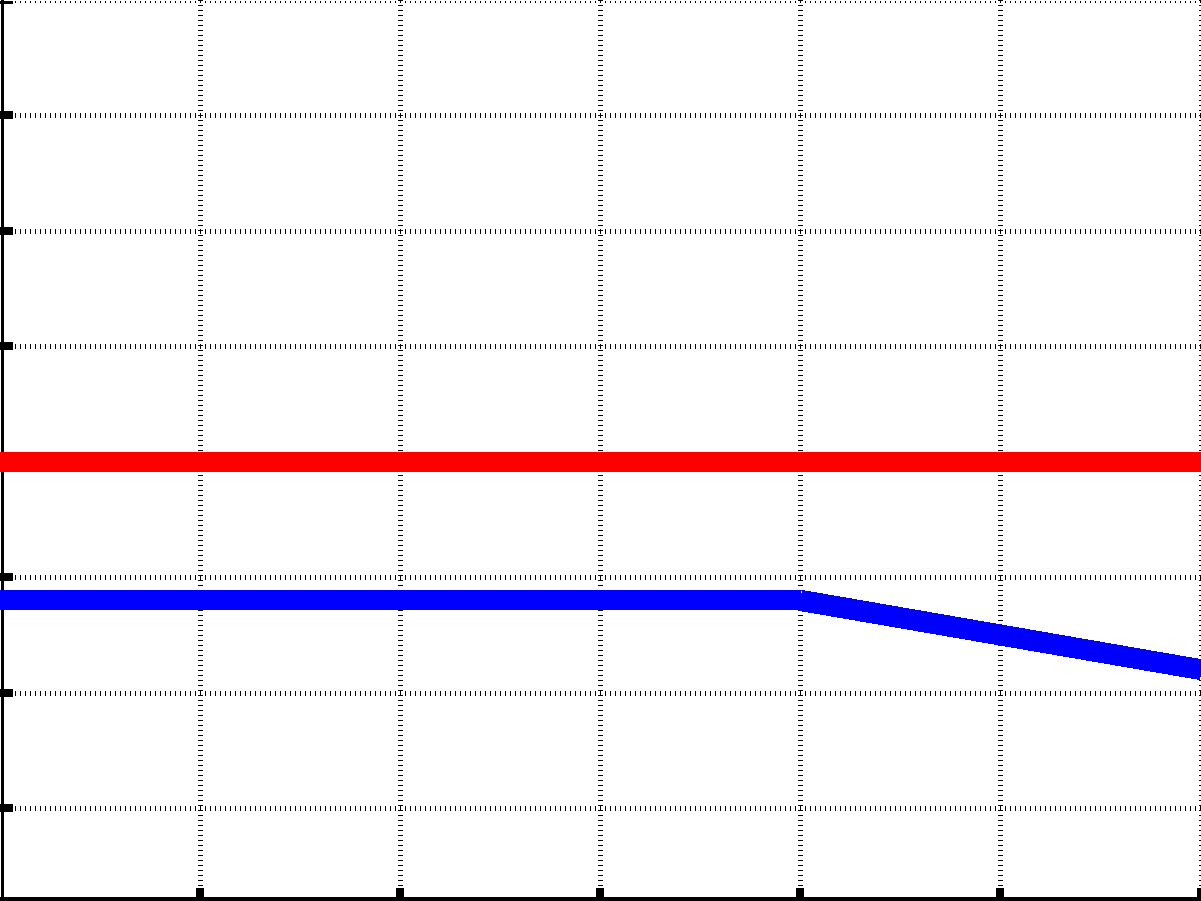} \\   
  \end{tabular}
    }&
    \multicolumn{4}{c}{
  \begin{tabular}{ccc}
\includegraphics[height = \dbimsz]{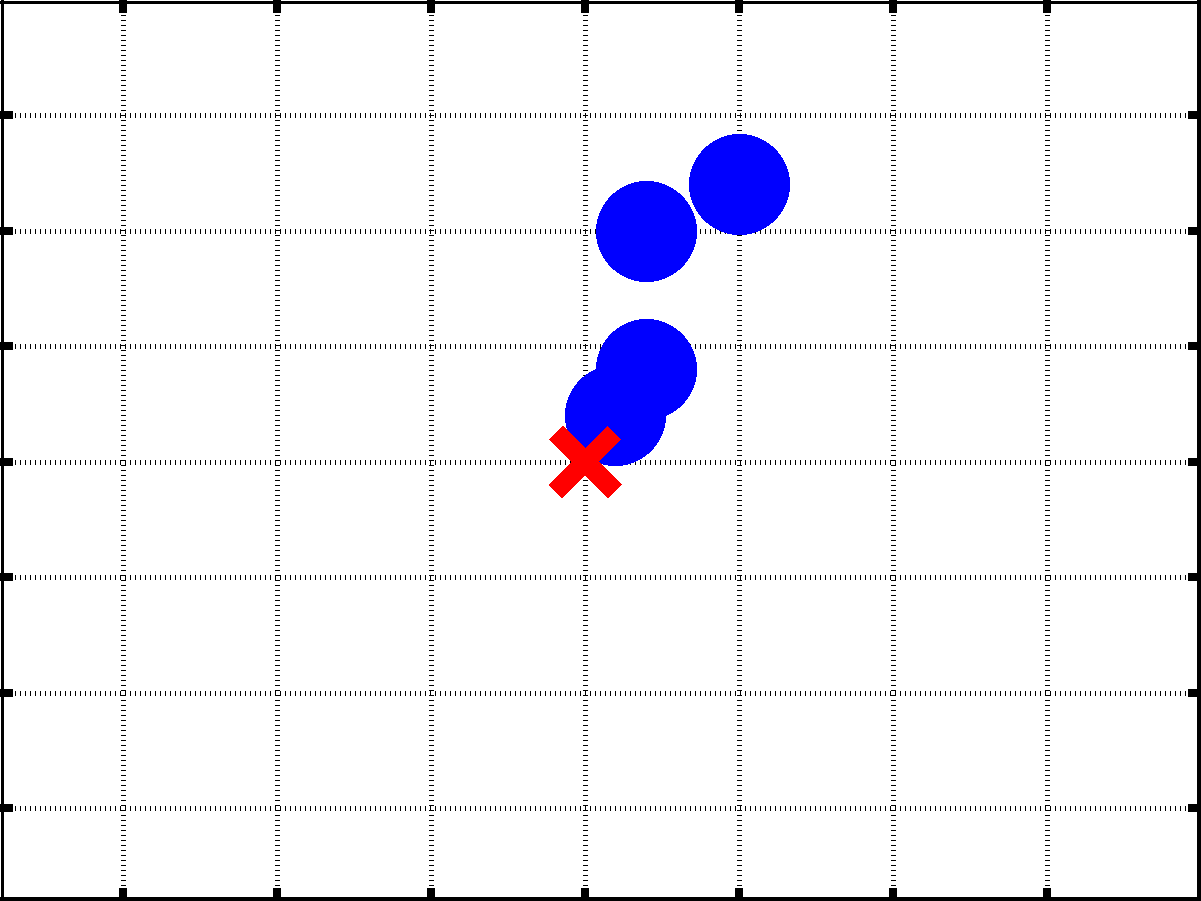} & 
  \hspace{-0.3cm}\includegraphics[height = \dbimsz]{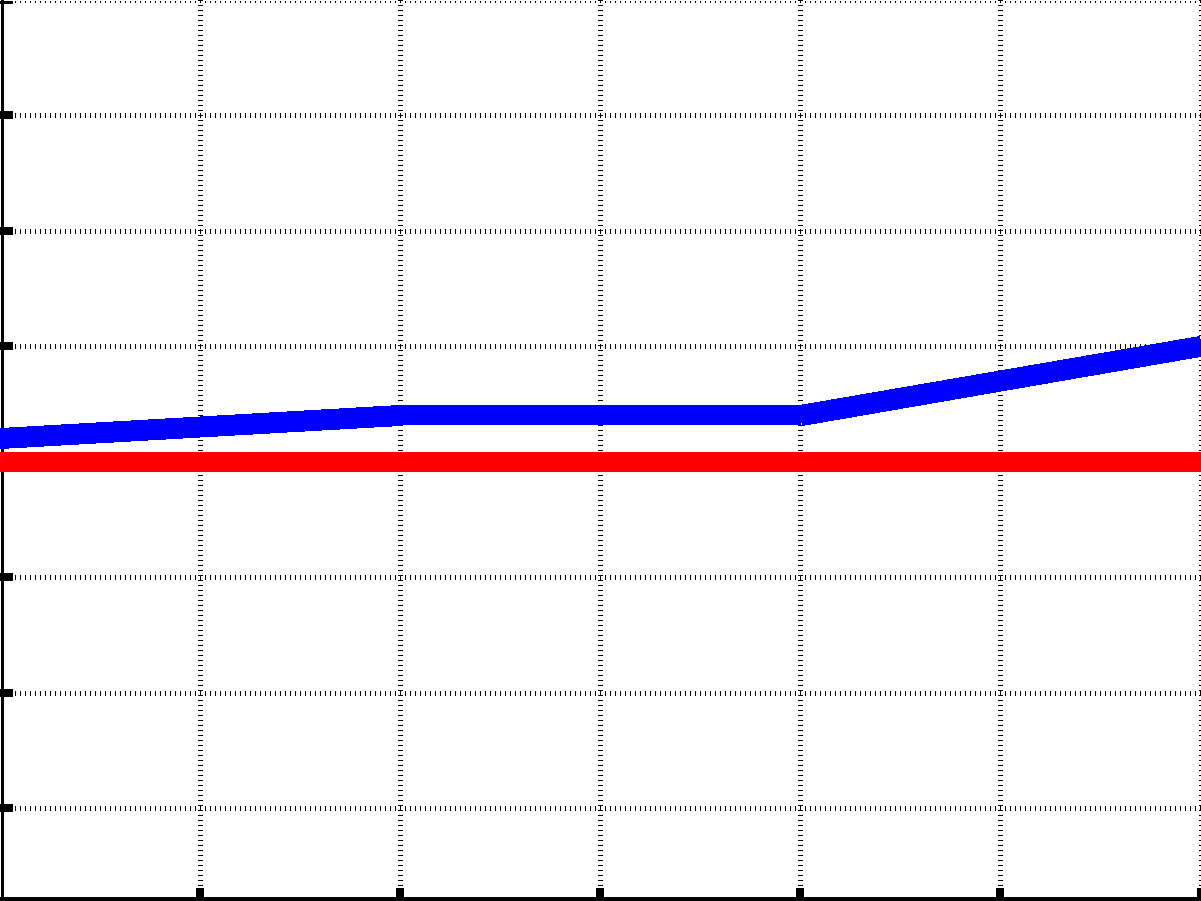} & 
  \hspace{-0.3cm}\includegraphics[height = \dbimsz]{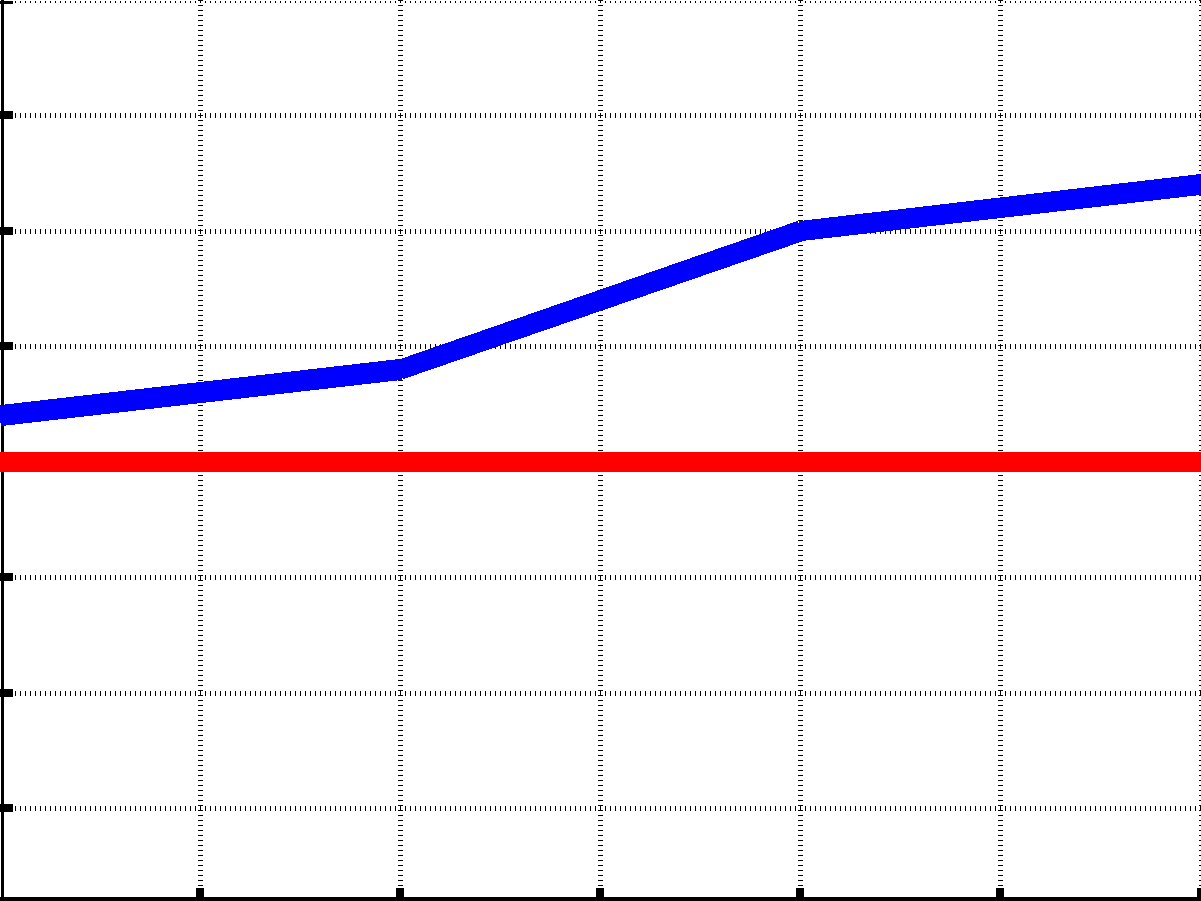} \\ \end{tabular}
} &

  \multicolumn{4}{c}{
  	\begin{tabular}{ccc}
  		\includegraphics[height = \dbimsz]{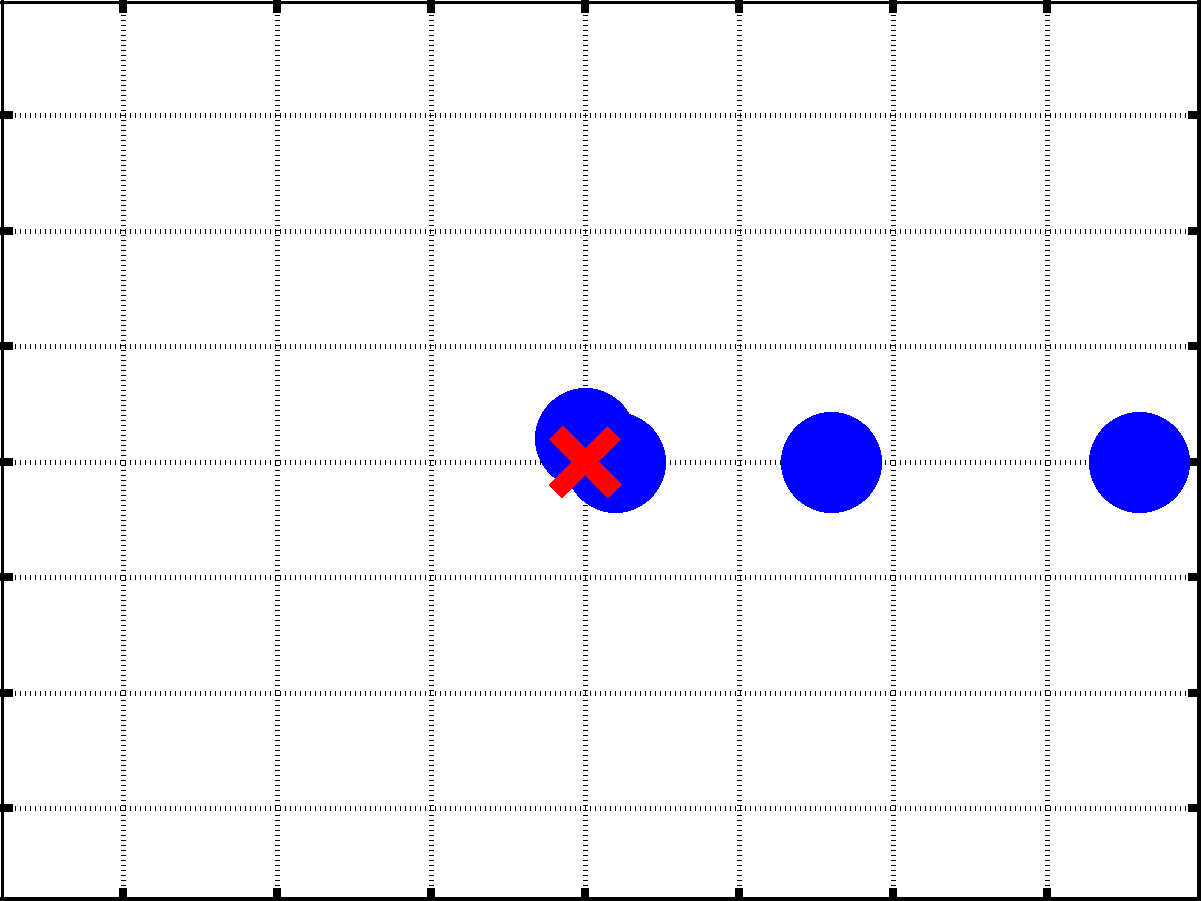} & 
  		\hspace{-0.3cm}\includegraphics[height = \dbimsz]{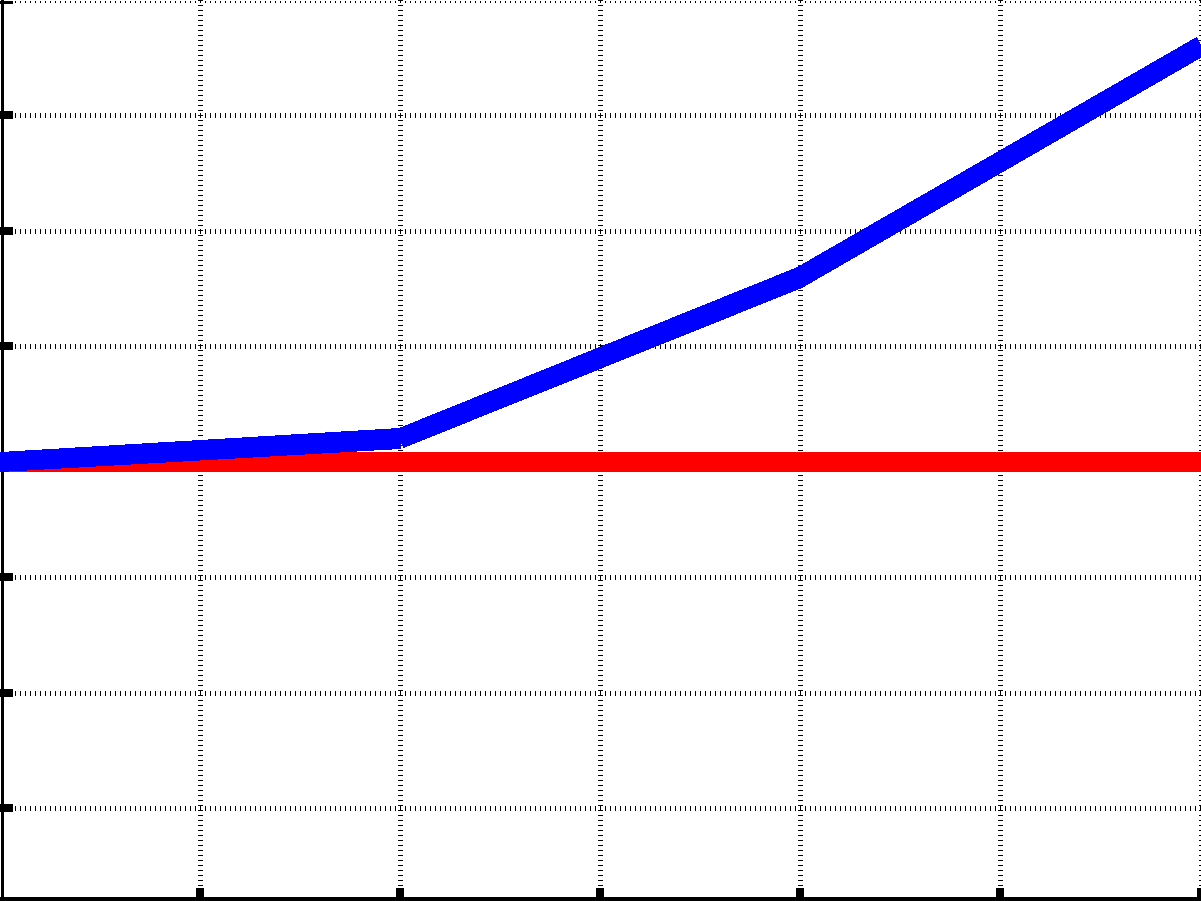} & 
  		\hspace{-0.3cm}\includegraphics[height = \dbimsz]{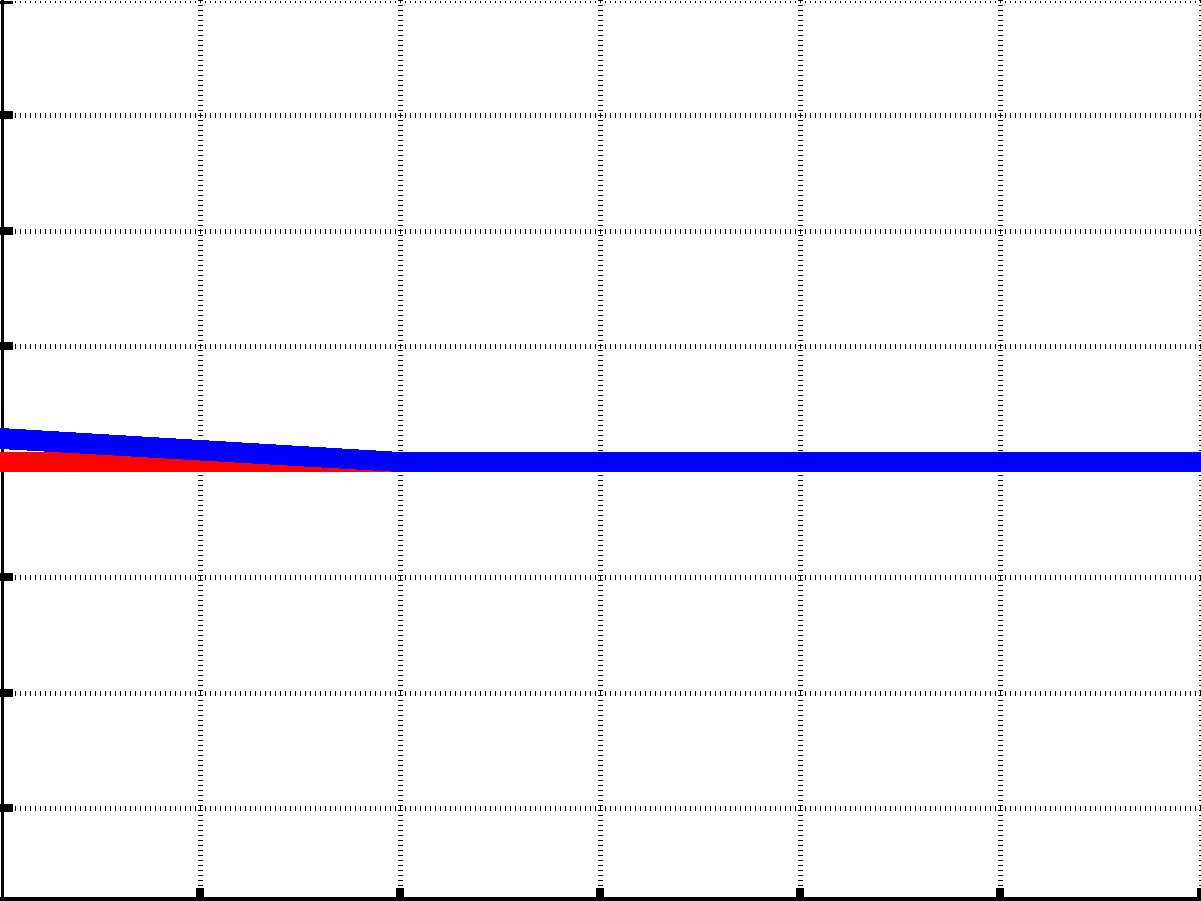} \\   
  	\end{tabular}
  }&
  \multicolumn{4}{c}{
  	\begin{tabular}{ccc}
  		\includegraphics[height = \dbimsz]{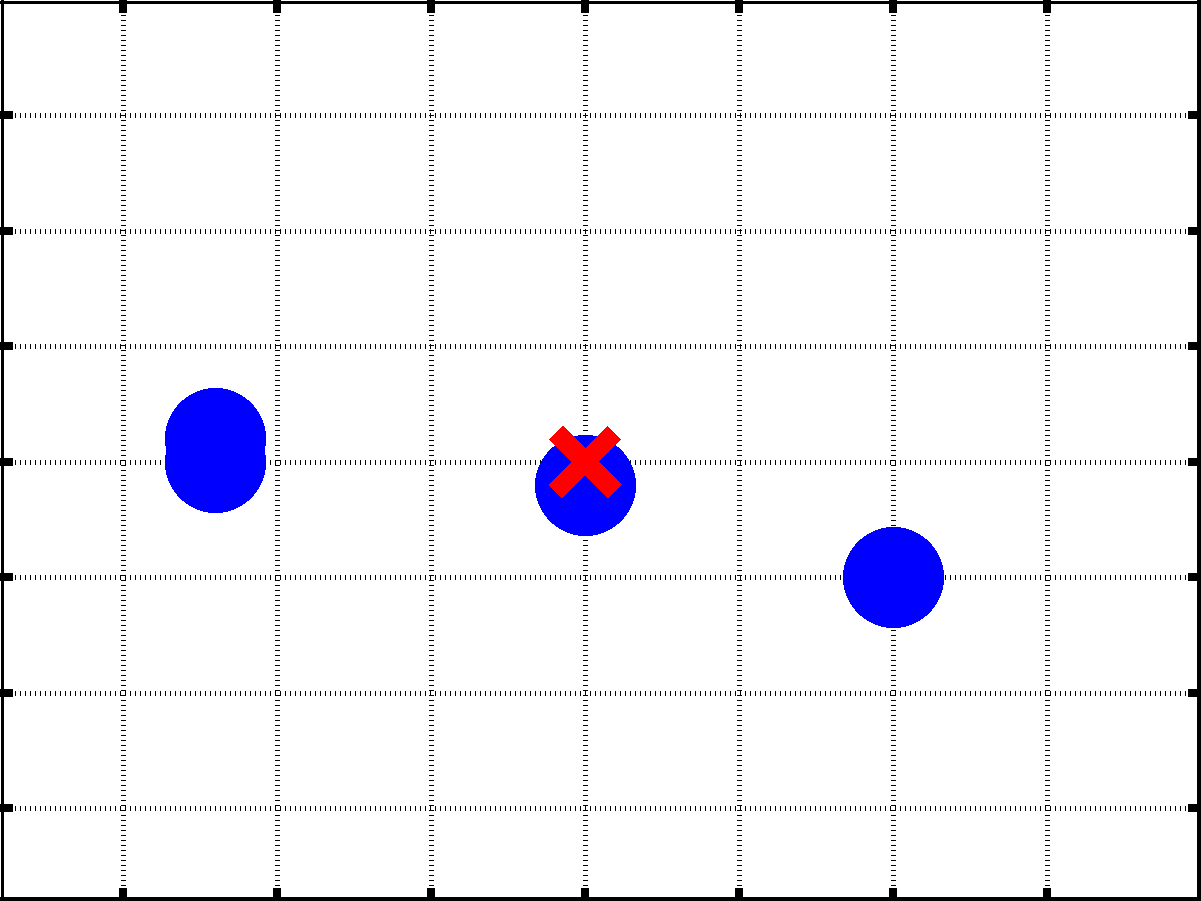} & 
  		\hspace{-0.3cm}\includegraphics[height = \dbimsz]{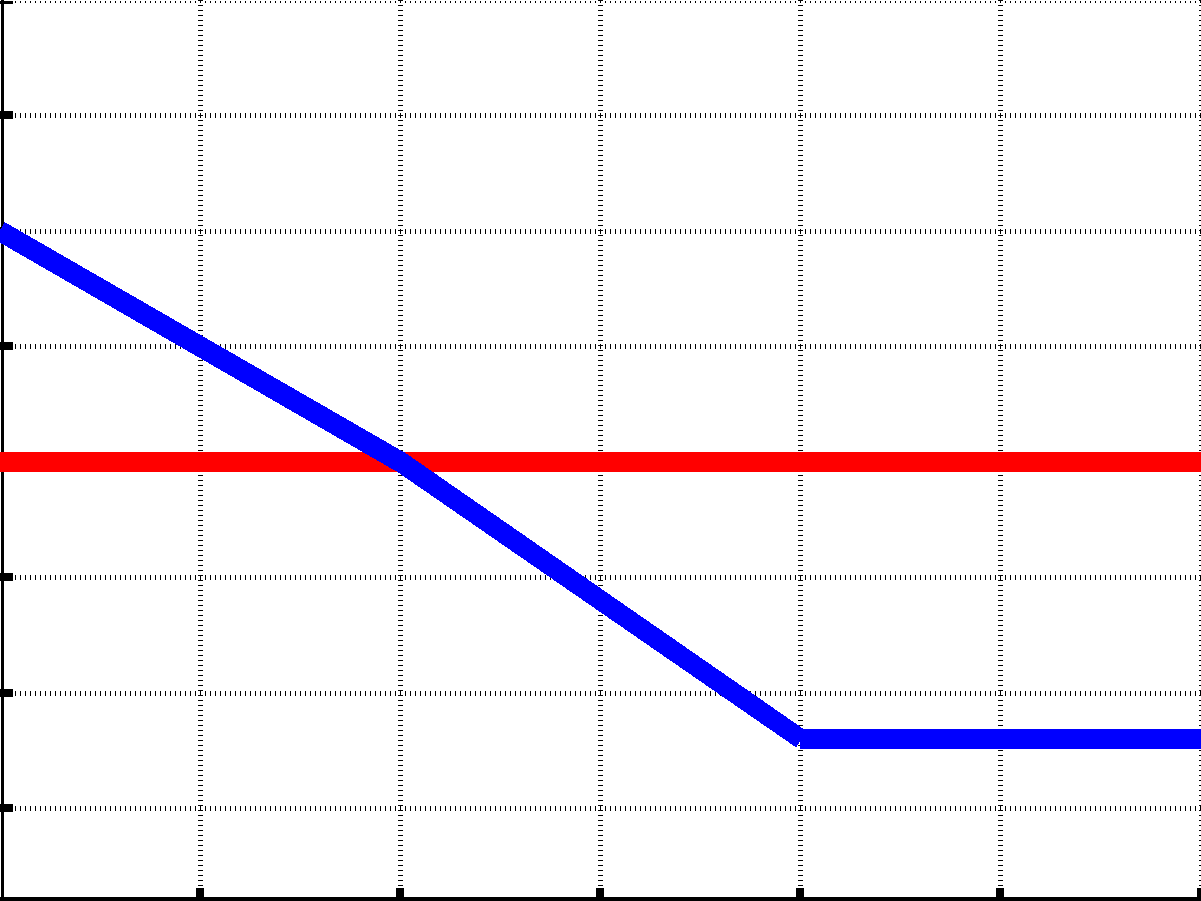} & 
  		\hspace{-0.3cm}\includegraphics[height = \dbimsz]{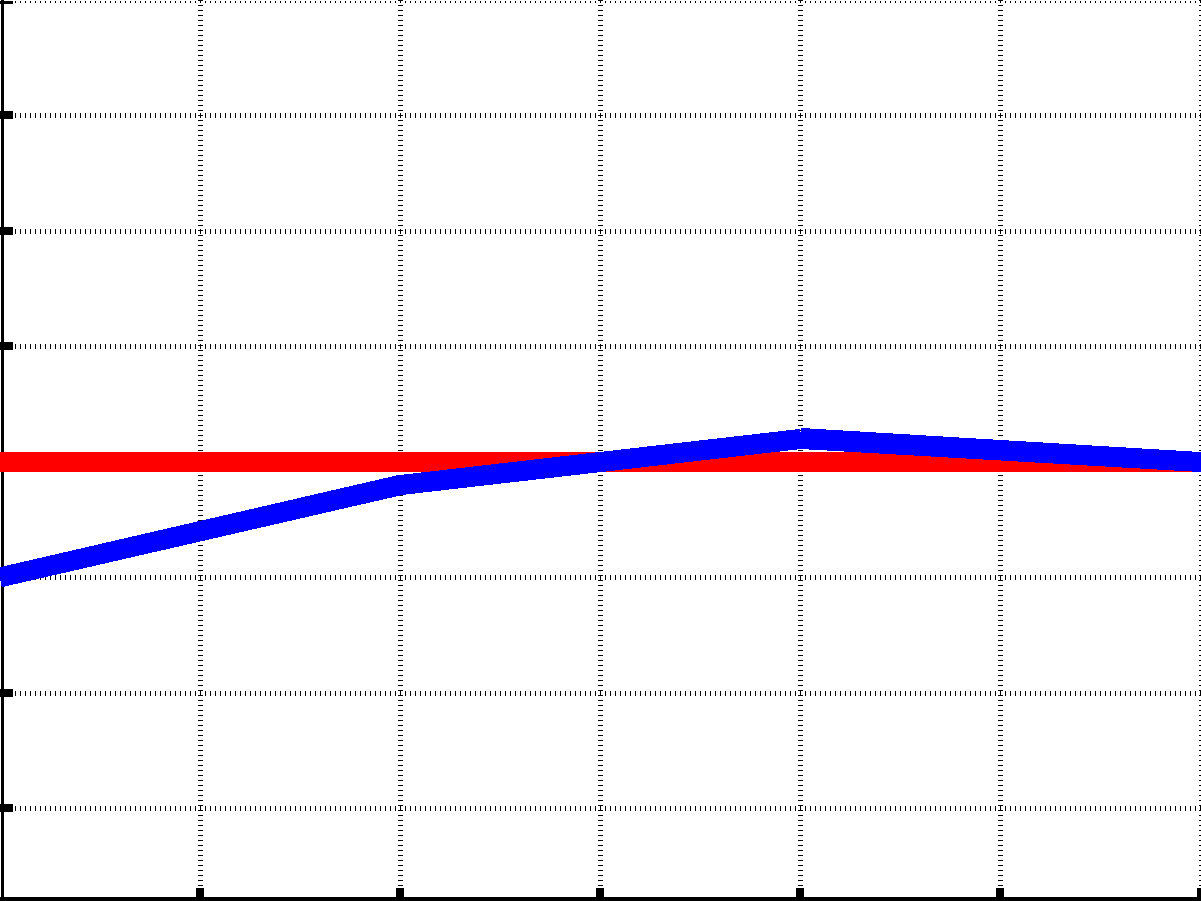} \\   
  	\end{tabular}
  } \\   
    \midrule
&&&&&&\multicolumn{4}{c}{Our approach} &&&&&&\\
\cmidrule(r){7-10} 
\includegraphics[width = \dbimsz]{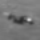} & 
  \hspace{-0.3cm}\includegraphics[width = \dbimsz]{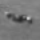} & 
  \hspace{-0.3cm}\includegraphics[width = \dbimsz]{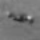} & 
  \hspace{-0.3cm}\includegraphics[width = \dbimsz]{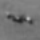} &
\includegraphics[width = \dbimsz]{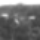} & 
  \hspace{-0.3cm}\includegraphics[width = \dbimsz]{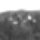} & 
  \hspace{-0.3cm}\includegraphics[width = \dbimsz]{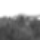} & 
  \hspace{-0.3cm}\includegraphics[width = \dbimsz]{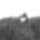} &
  \includegraphics[width = \dbimsz]{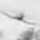} & 
  \hspace{-0.3cm}\includegraphics[width = \dbimsz]{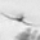} & 
  \hspace{-0.3cm}\includegraphics[width = \dbimsz]{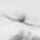} & 
  \hspace{-0.3cm}\includegraphics[width = \dbimsz]{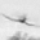} & 
  \includegraphics[width = \dbimsz]{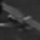} & 
  \hspace{-0.3cm}\includegraphics[width = \dbimsz]{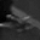} & 
  \hspace{-0.3cm}\includegraphics[width = \dbimsz]{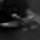} & 
  \hspace{-0.3cm}\includegraphics[width = \dbimsz]{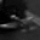} \\
    \multicolumn{4}{c}{
  \begin{tabular}{ccc}
\includegraphics[height = \dbimsz]{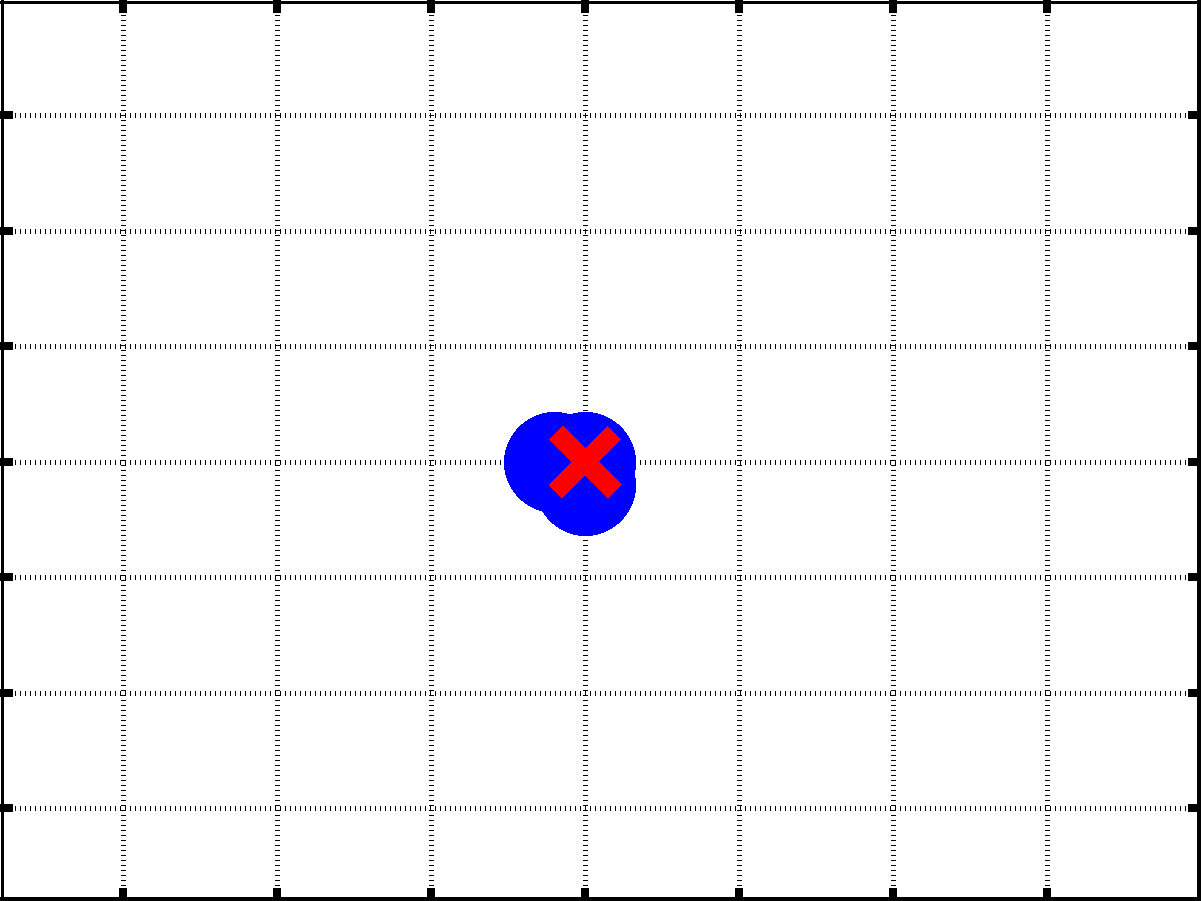} & 
  \hspace{-0.3cm}\includegraphics[height = \dbimsz]{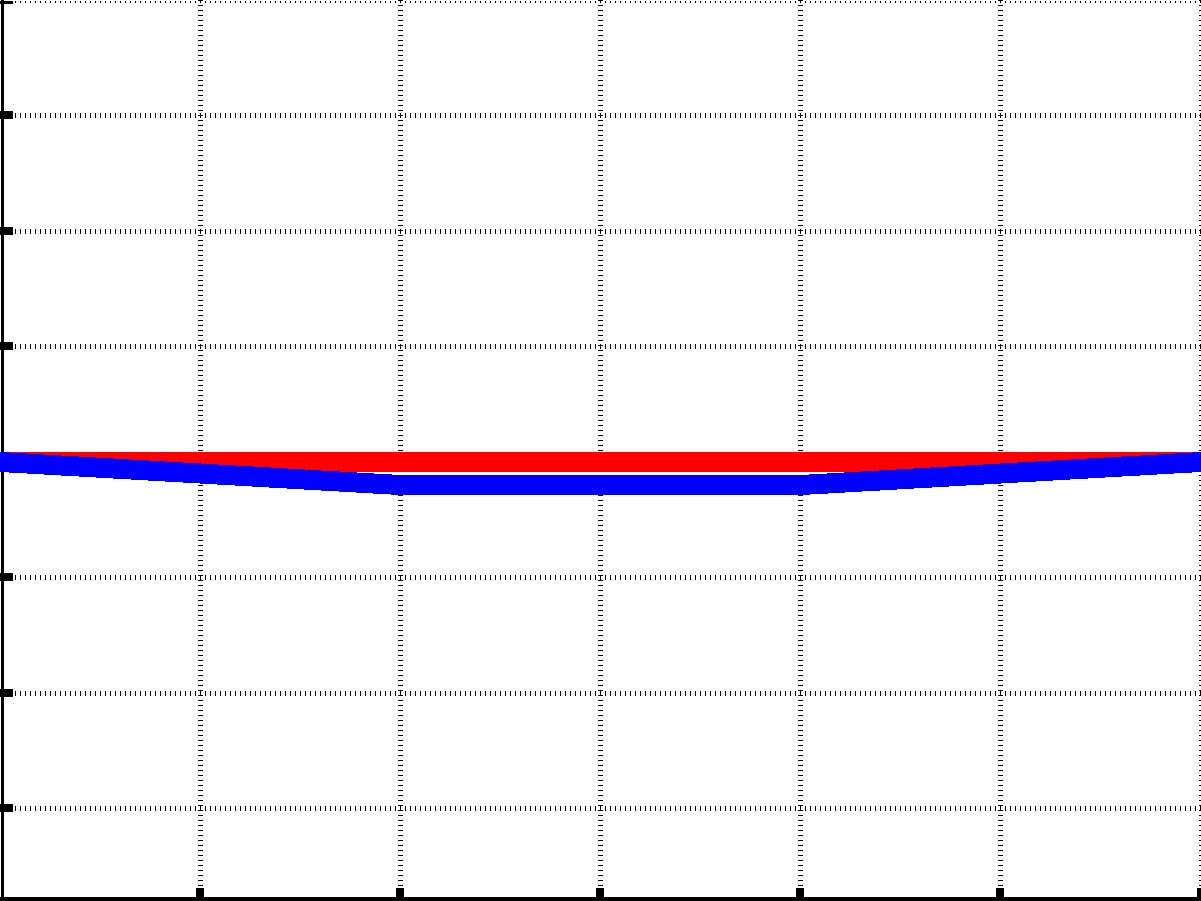} & 
  \hspace{-0.3cm}\includegraphics[height = \dbimsz]{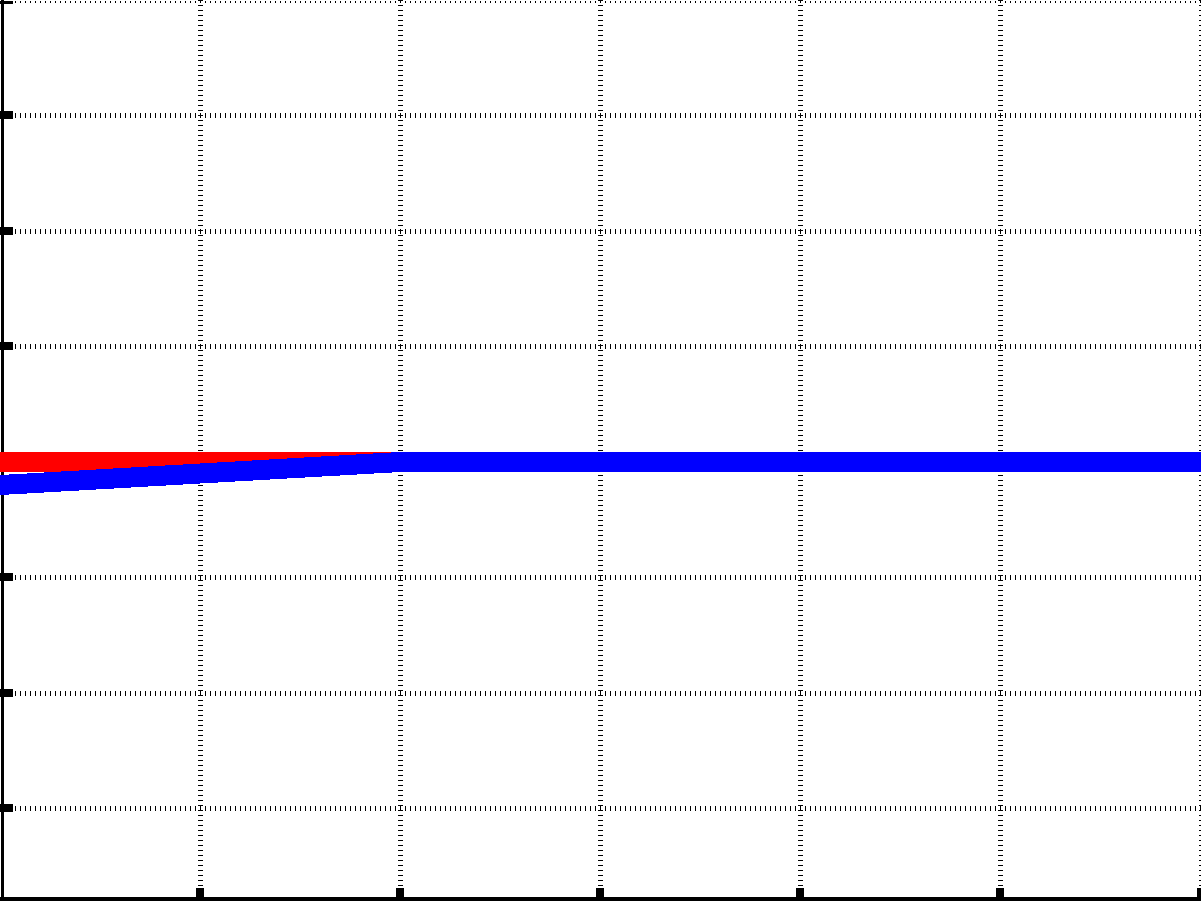} \\   
  \end{tabular}
    }&
    \multicolumn{4}{c}{
  \begin{tabular}{ccc}
\includegraphics[height = \dbimsz]{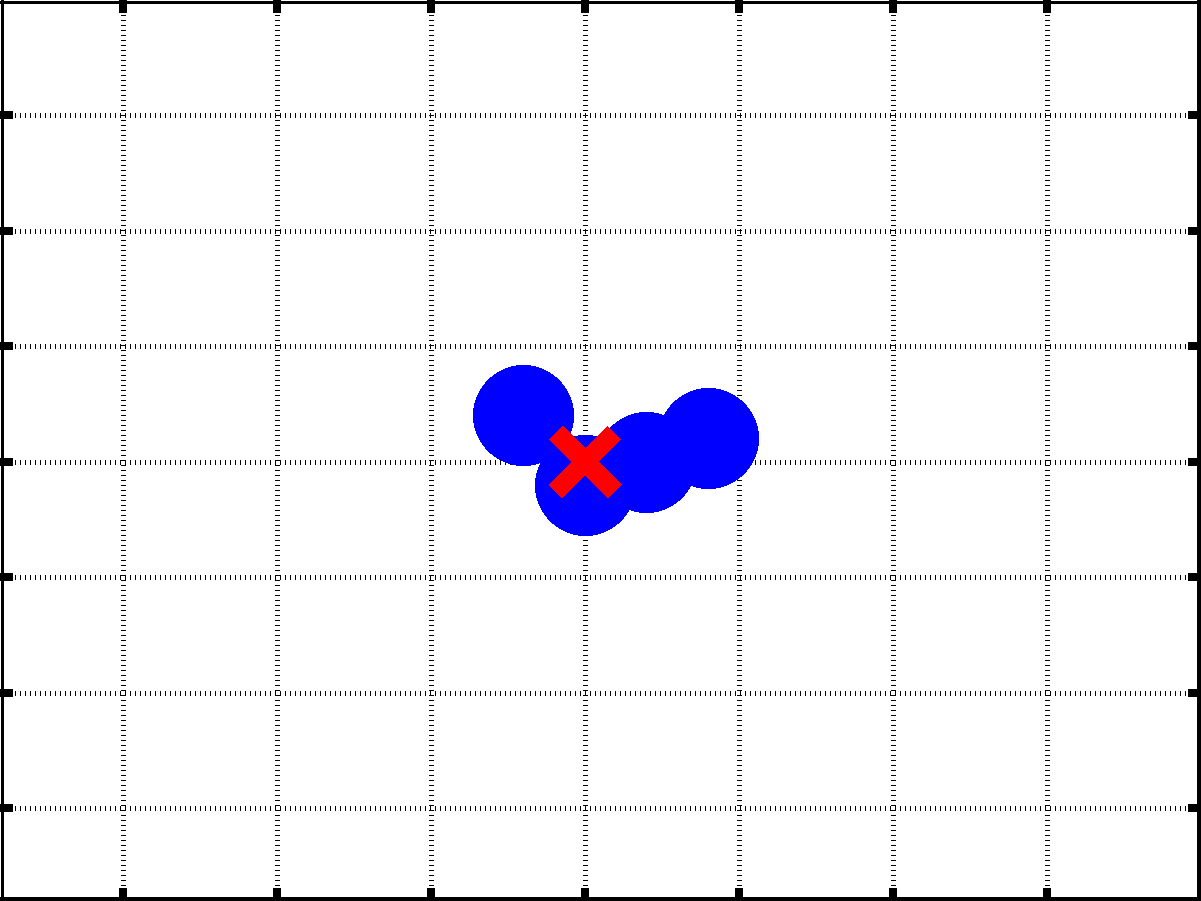} & 
  \hspace{-0.3cm}\includegraphics[height = \dbimsz]{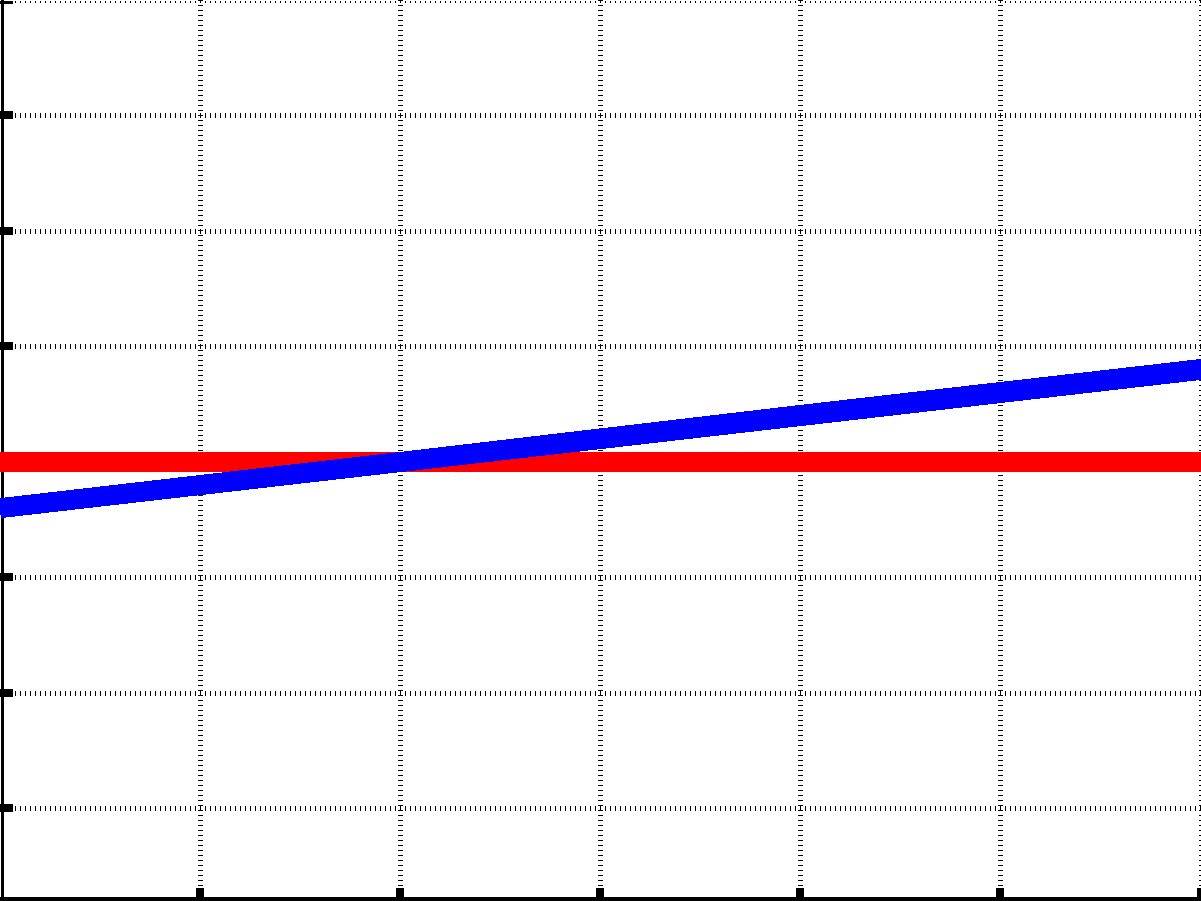} & 
  \hspace{-0.3cm}\includegraphics[height = \dbimsz]{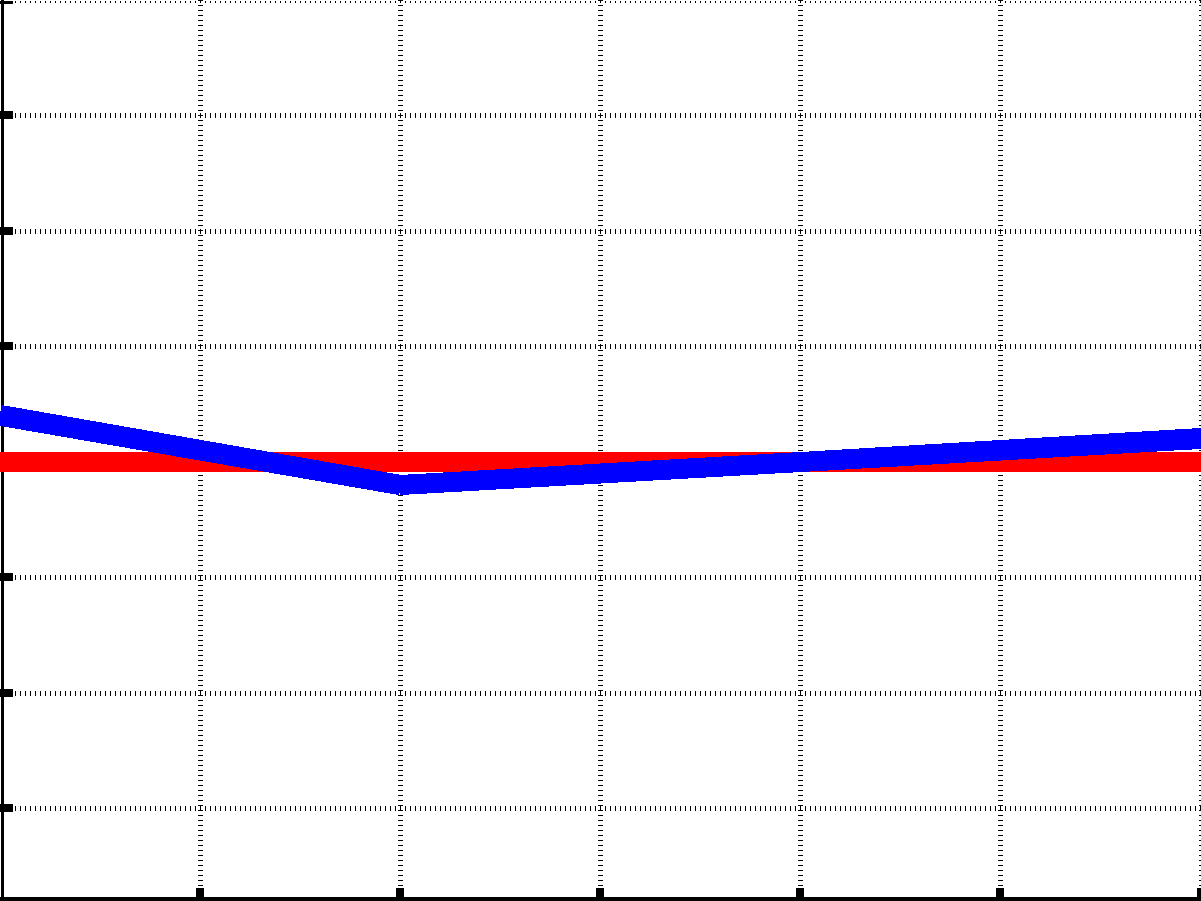} \\
    \end{tabular}
} &
  \multicolumn{4}{c}{
  	\begin{tabular}{ccc}
  		\includegraphics[height = \dbimsz]{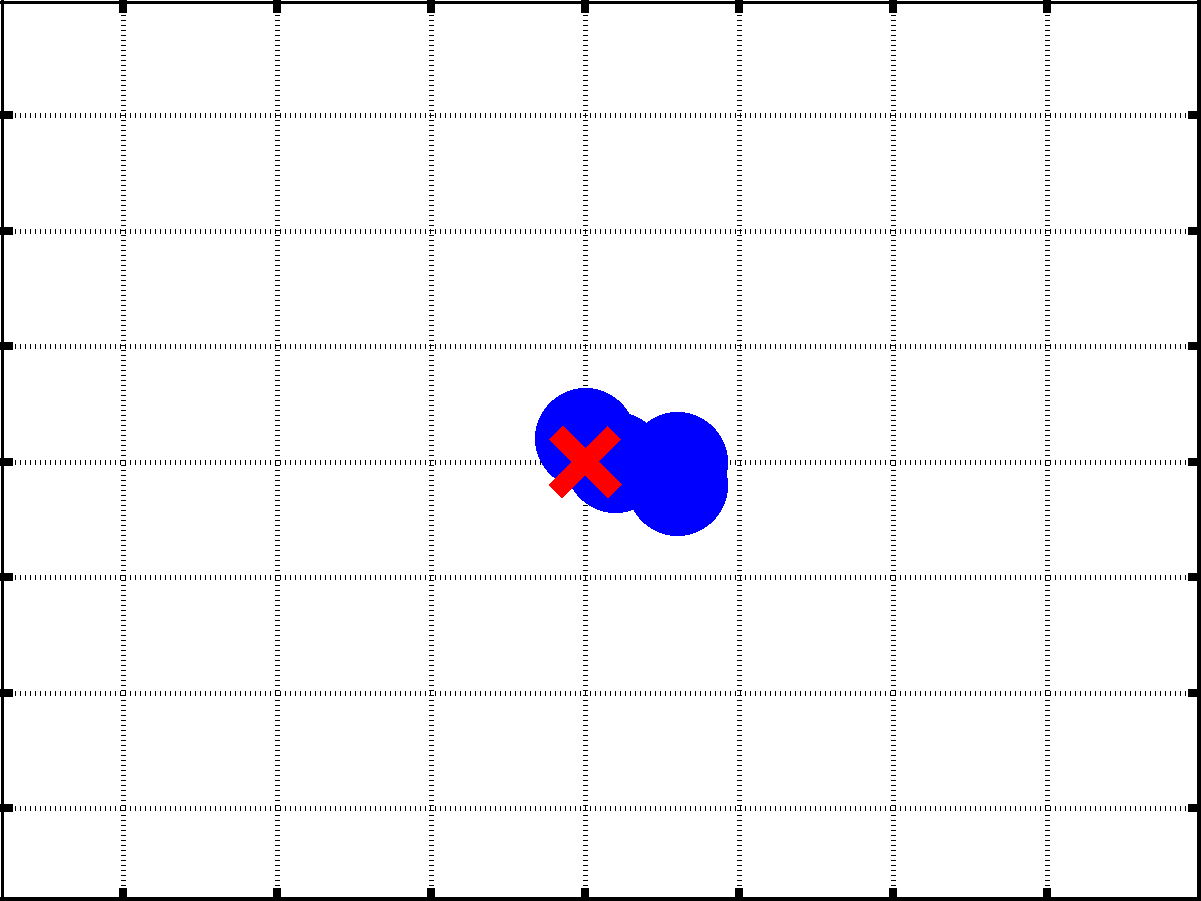} & 
  		\hspace{-0.3cm}\includegraphics[height = \dbimsz]{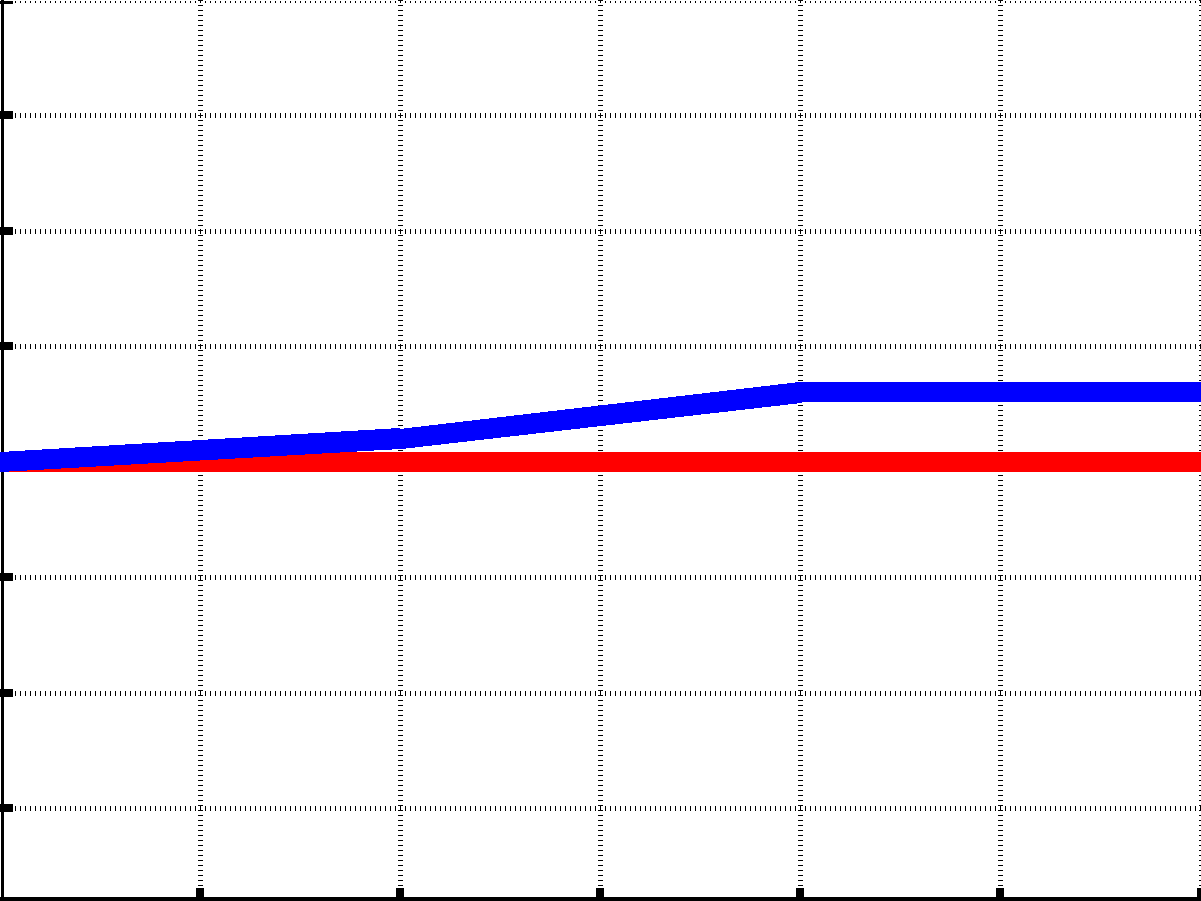} & 
  		\hspace{-0.3cm}\includegraphics[height = \dbimsz]{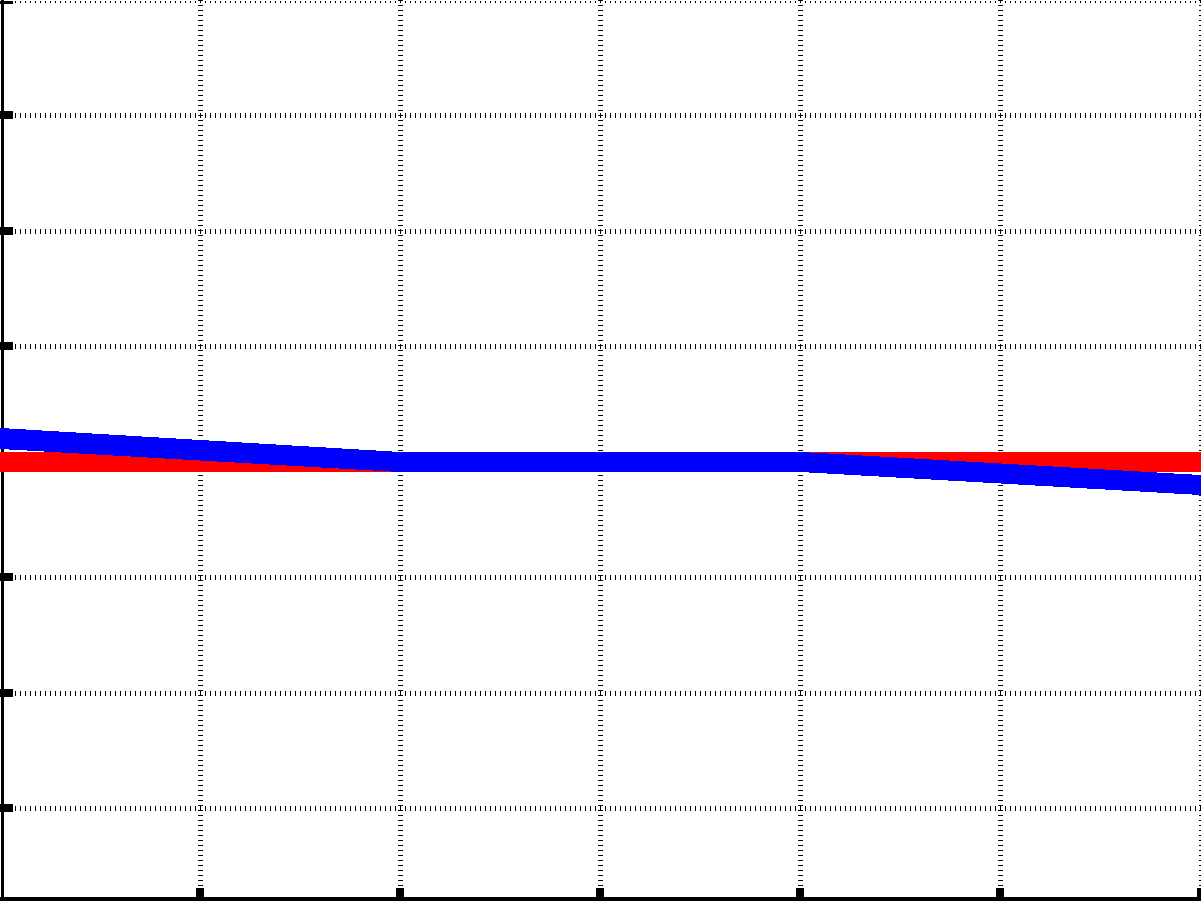} \\   
  	\end{tabular}
  }&
  \multicolumn{4}{c}{
  	\begin{tabular}{ccc}
  		\includegraphics[height = \dbimsz]{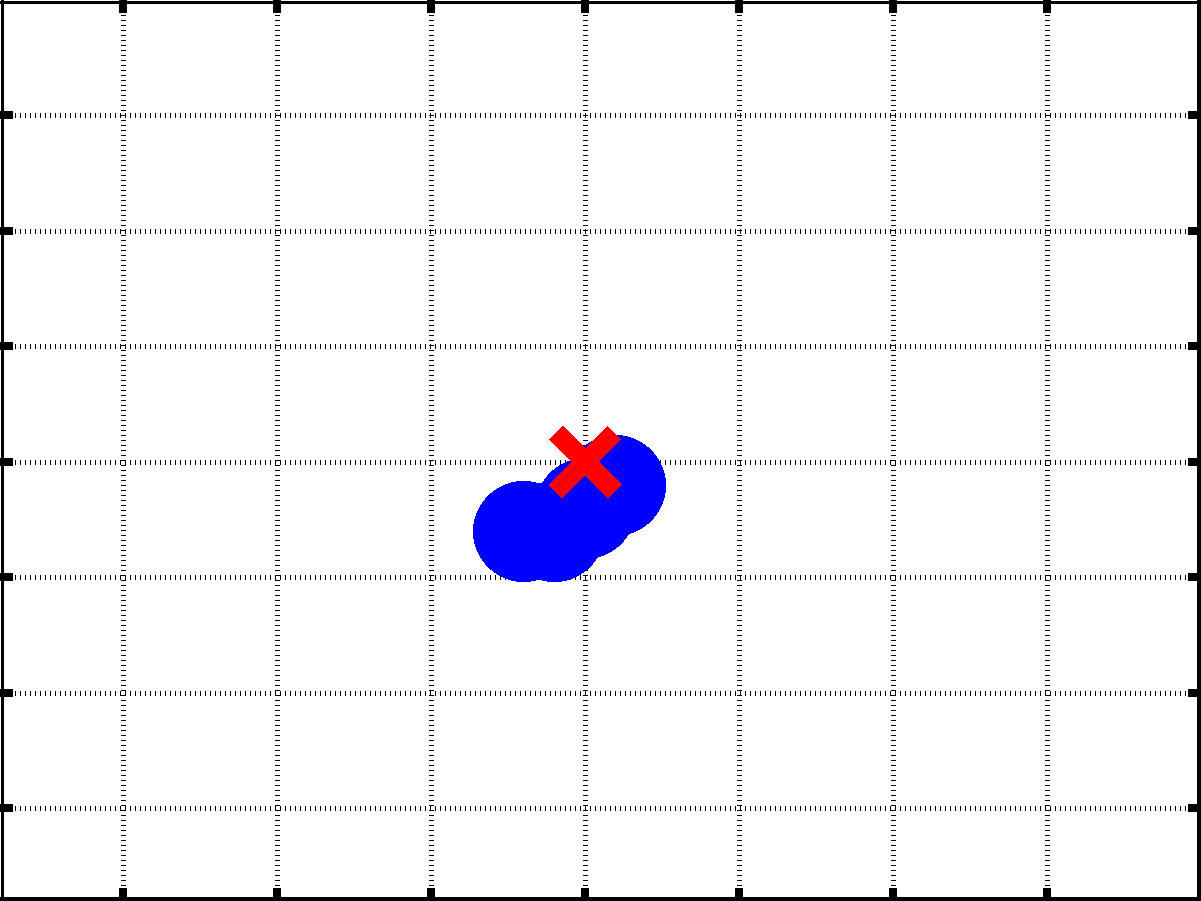} & 
  		\hspace{-0.3cm}\includegraphics[height = \dbimsz]{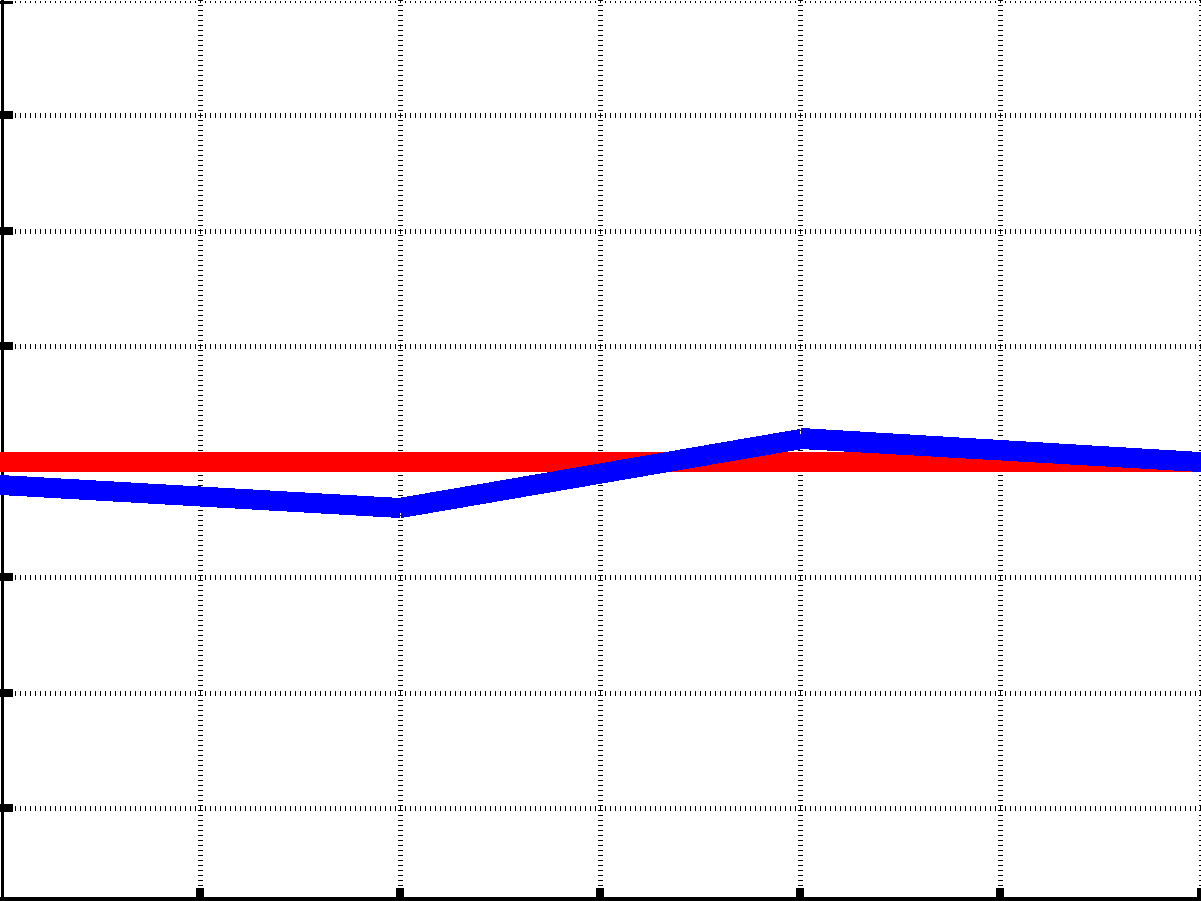} & 
  		\hspace{-0.3cm}\includegraphics[height = \dbimsz]{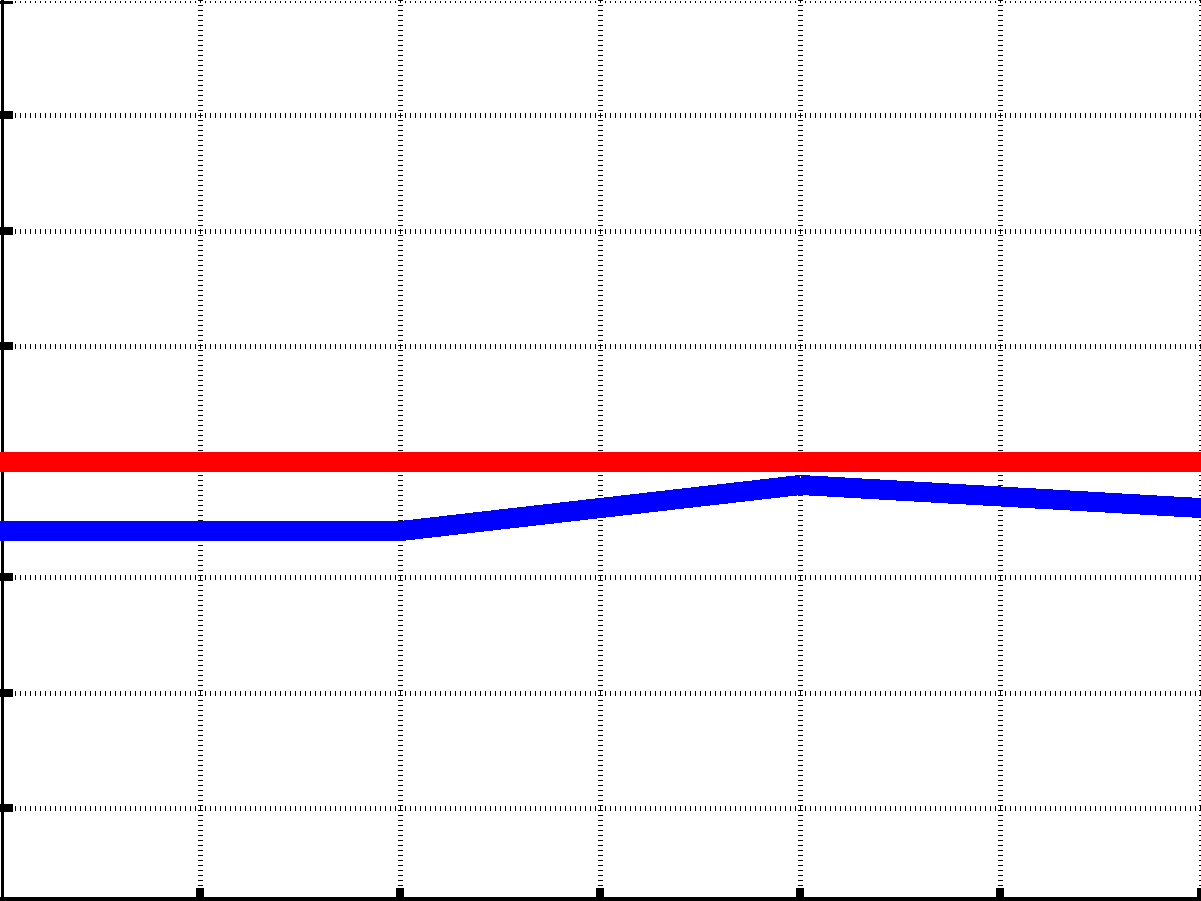} \\   
  	\end{tabular}
  } \\   
\bottomrule
\multicolumn{4}{c}{(a)} & \multicolumn{4}{c}{(b)} & \multicolumn{4}{c}{(c)}  & \multicolumn{4}{c}{(d)}\\
\end{tabular}
\vspace{-0.3cm}
\mycaption{Compensation for the apparent motion of different flying objects inside the st-cube allows to decrease in-class variation of the data, used  by the machine learning algorithms.  For each \stcb{},  we also  provide three  graphs:  The blue  dots in  the first  graph indicate the locations of the center  of the drone throughout the \stcb{}, the red cross indicates  the patch center. The  next two graphs plot the  variations of the $x$ and $y$  coordinates of the center of the  drone respectively, compared to the position of the center of the patch.  We can see that our method keeps the drone close to the center even for complicated backgrounds and when the drone is barely recognizable as in the right column. }
\label{fig:stabilize}
\end{figure*}

In this section, we first introduce a  basic approach to using \stcb{}s, that is, blocks of consecutive frames, for  object detection without first correcting for motion.    We   then  introduce   our   regression-based   approach  to   motion stabilization.  We will demonstrate in the result section that it brings a substantial performance improvement.

\subsection{Detection without Motion Stabilization}
\label{sec:basic}

\comment{\vincentrmk{I vote for changing $s_z$ into $s_t$.  But  that's a detail}}

Let $s_x$ and $s_y$ be spatial, and $s_t$ be temporal dimensions of a \stcb{}
such  as those  depicted  by \fig{st_cube}.   We  use a  training  set of  pairs $(b_i,y_i), i \in [1,N] $, where $ b_i  \in R^{s_x \times s_y \times s_t} $ is a \stcb{} and the  label $y_i \in [-1,1]$  indicates whether or not  it contains a target object. We then train an AdaBoost classifier:
\begin{equation}
  F: \mathbb{R}^{s_x \times s_y \times s_t} \rightarrow [0,1], \qquad F(b) = \mathop{\Sigma}\limits_{j=1}^T  \alpha_jf_j(b)    
  \label{eq:AdaBoost}
\end{equation}
where the $\alpha_j$  are learned weights and $T$ is the number of weak classifiers $f_j$ learned by the algorithm. We use $f_j$ of the form
\begin{equation}
\label{eq:wl}
f_{R,o,\tau}(b) = \left \{
   \begin{array}{ll}
       1 & \text{ if } E(b, R, o) > \tau, \\ 
       0 & \text{ otherwise.} \\
   \end{array}
   \right. 
\end{equation}
These weak learners are parametrized by  a box $R$ within $b$, an orientation
$o$ and a threshold $\tau$. $E(b, R, o)$ is the normalized image gradient energy
at orientation $o$ over the region $R$~\cite{Levi04}.

\comment{
\note{From  the description  above, I  would not  call these  Haar wavelets  and
  removed the term from the text. Can you check the text below?}
\ans{Yes, that's all correct.}
}

As a potential  alternative to these image  features, we tested a  3D version of
the HOG detector as in~\cite{Weinland10}. However, we found that its performance
depends critically on the  size of the bins used to compute  it. In practice, we
found it difficult to find sizes that consistently gave good results for objects
whose apparent shape can change dramatically.  The AdaBoost procedure solves this
problem by automatically selecting an appropriate range of sizes of the boxes
$R$ of Eq.~\ref{eq:wl}.

One  problem the AdaBoost procedure does not  address,  however, is  that the  orientations of  the gradients are biased  by the global object motion and that  this bias is independent of  object appearance.  This makes  the learning  task much  more difficult  and motion stabilization is required to eliminate this problem.

\subsection{Object-Centric Motion Stabilization}
\label{sec:stabilization}

\begin{figure}
\centering
\begin{tabular}{cccccc}
\toprule
\multicolumn{3}{c}{mUAVs} & \multicolumn{3}{c}{Aircrafts} \\
\midrule
\includegraphics[width = \dbimsz]{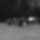} &
  \hspace{-0.3cm}\includegraphics[width = \dbimsz]{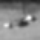} &
  \hspace{-0.3cm}\includegraphics[width = \dbimsz]{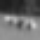} &
\includegraphics[width = \dbimsz]{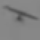} & 
  \hspace{-0.3cm}\includegraphics[width = \dbimsz]{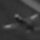} & 
  \hspace{-0.3cm}\includegraphics[width = \dbimsz]{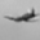} 
\\
\includegraphics[width = \dbimsz]{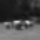} &
  \hspace{-0.3cm}\includegraphics[width = \dbimsz]{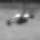} &
  \hspace{-0.3cm}\includegraphics[width = \dbimsz]{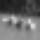} &
\includegraphics[width = \dbimsz]{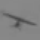} & 
  \hspace{-0.3cm}\includegraphics[width = \dbimsz]{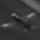} & 
  \hspace{-0.3cm}\includegraphics[width = \dbimsz]{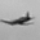} 
\\
\includegraphics[width = \dbimsz]{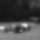} &
  \hspace{-0.3cm}\includegraphics[width = \dbimsz]{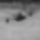} &
  \hspace{-0.3cm}\includegraphics[width = \dbimsz]{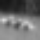} &
\includegraphics[width = \dbimsz]{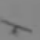} & 
  \hspace{-0.3cm}\includegraphics[width = \dbimsz]{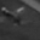} & 
  \hspace{-0.3cm}\includegraphics[width = \dbimsz]{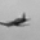} 
\\
\includegraphics[width = \dbimsz]{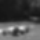} &
  \hspace{-0.3cm}\includegraphics[width = \dbimsz]{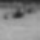} &
  \hspace{-0.3cm}\includegraphics[width = \dbimsz]{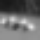} &
\includegraphics[width = \dbimsz]{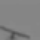} &
  \hspace{-0.3cm}\includegraphics[width = \dbimsz]{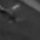} &
  \hspace{-0.3cm}\includegraphics[width = \dbimsz]{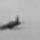} 
\\
\bottomrule
\end{tabular}
\mycaption{Sample patches of the mUAVs and aircrafts. Each column corresponds to a single \stcb{} and illustrates one kind from the variety of possible motions that an aircraft could have.}
\label{fig:st_cube}
\end{figure}

The best way  to avoid the above-mentioned  bias is to guarantee  that the target
object, if present in an \stcb{}, remains at the center of all spatial slices.

More specifically,  let $I_t$ denote the  $t^{th}$ frame of the  video sequence.
If we do not compensate for the motion, we can define the \stcb{} $b_{i,j,t}$ as the
3-D array of pixel intensities from $I_z, z \in [t-s_t+1, t]$ at image locations
$   (k,l),   k  \in   [i-s_x+1,i],   l   \in   [j-s_y+1,j]$,  as   depicted   by
\fig{st_cube}. Given these notations, correcting for motion can be formulated as
allowing  the $s_t$  spatial  slices  $m_{i,j,z}, z  \in  [t-s_t+1,t]$ to  shift
horizontally and vertically in individual images.

In~\cite{Park13}, these  shifts are computed  using flow information,  which has
been  shown to  be effective  in  the case  of  pedestrians who  occupy a  large
fraction of  the image and  move relatively slowly from  one frame to  the next.
However, as can  be seen in \fig{st_cube}  these assumptions do not  hold in our
case  and we  will  show in  the  result section  that  this negatively  impacts
the performance.

To  overcome this  difficulty, we  introduce  instead a  regression-based approach  to compensate  for motion  and keep  the object  in the  center of  the $m_{i,j,z}$ spatial slices even when the target object's appearance changes drastically.

\myparagraph{Training the regressors}

We propose  to train  two boosted  trees regressors~\cite{Sznitman13a},  one for
horizontal motion of the aircraft and one for its vertical motion.  The power of
this method is  that it does not use the  similarity between consecutive frames,
and is able to  predict how far the object is from the  center in the horizontal
or vertical directions, based just on a single patch.

We use  gradient boosting~\cite{Zheng07} to  learn regression  models for vertical $\phi_v(\cdot)$ and horizontal motion $\phi_h(\cdot)$. Each of these models $\phi_*:
\mathbb{R}^{s_x \times  s_y} \rightarrow \mathbb{R}$ can be represented in the form $  \phi_*(m) = \mathop{\Sigma}_{j  = 1}^T  \alpha_jh_j(m)$,  where  $\alpha_{j=1..T}$ are  real valued weights, $h_j:  \mathbb{R} ^{s_x \times s_y}  \rightarrow \mathbb{R}$ are weak learners  and $m \in \mathbb{R}^n$  is the input patch.   The GradientBoost approach  can  be  seen  as  extension of  the  classic  AdaBoost  algorithm  to real-valued weak learners and more general loss functions.


As  typically done  with gradient  boosting we  use regression  trees $h_j(m)  = T(\theta_j,HoG(m))$ as weak  learners for this approach,  where $\theta_j$ denotes the tree parameters.  $HoG(m)$ denotes the Histograms of Gradients for patch $m$. At every iteration $j$ the boosting approach finds the  weak learner $h_j(\cdot)$ that minimises the quadratic loss function
\begin{equation}
h_j(\cdot) = \mathop{argmin} \limits _{h(\cdot)} \left( \mathop{\Sigma}\limits_{i=1}^N w_i^j(h(x_i)-r_i)^2 \right),
\label{eq:quadratic_loss}
\end{equation}
where  $N$  is  the  number  of  training  samples  $m_i$  with  their
expected responses $r_i$. Weights  $w_i^j$ are estimated at every
iteration, by differentiating the loss function.


We used  the $HoG(\cdot)$ representation for  the patches $m_{i=1..N}$
because it is fast to compute  and proved to be robust to illumination
changes in many applications.  Therefore the regressor is able to perform in the outdoor environments, where illumination can  significantly change from one part of the video sequence to another.

\myparagraph{Motion compensation with regression}

After both  regressors for horizontal and  vertical motions are trained,  we use
them to compensate for the motion of the aircraft inside the \stcb{} $b_{i,j,t}$
in  an iterative  way.   \alg{regression}  outlines the  main  steps the  motion
compensation  approach takes  to  estimate  and correct  for  the  shift of  the
aircraft.  The  resulting  \stcb{}  keeps  the  aircraft  close  to  the  center
throughout   the  whole   sequence  of   patches   $m_{k=1..s_t}$   of
$b_{i,j,t}$. This approach provides not only  a better prediction, but also
allows  to estimate  the direction  of  motion of  the aircraft  and its  speed,
provided the  frame-rate of  the camera and  the size of  the target  object are
known.  This additional  information may be used by  various tracking algorithms
to improve their performance.

\begin{algorithm}[!t]
\caption{Regression based motion compensation.}
\label{alg:regression}
\begin{algorithmic}[0]
\STATE {\bf Input} 
\STATE\begin{enumerate}
\setlength\itemsep{0pt}
\setlength{\parskip}{0pt}
\item regressors $\phi_h(\cdot), \phi_v(\cdot)$ for horizontal and vertical motion respectively
\item \stcb{} $b_{i,j,t}$ with dimensions $s_x,s_y,s_t$
\item frames $I_p, p \in [t-s_t+1,t]$ of the video sequence
\end{enumerate}
\STATE
\STATE set $\epsilon = 1$
\FOR {$m_k, k \in [1,s_t]$}
\STATE set $ n = 1 $,  $(i_0,j_0) = (0,0)$ and $(i_1,j_1) = (i,j)$
\STATE
\STATE as it was previously defined, we refer to $m_k$ as the patch of the \stcb{} and to $ m_{i,j,p}, p = k+t-s_t$  as the patch extracted from the $I_p$ at the position $(i,j)$, so at the first iteration $m_k = m_{i_1,j_1,p}$
\STATE
\WHILE {$\left( (i_n-i_{n-1})^2+(j_n-j_{n-1})^2 \right) < \epsilon$}
\STATE $ n = n+1$
\STATE $ (sh_h,sh_v) = \left( \phi_h(m_p), \phi_v(m_p) \right)$
\STATE $ (i_n,j_n) = ( i_{n-1} - sh_v, j_{n-1} - sh_h )$
\STATE $ m_k = m_{i_n,j_n,p}$
\ENDWHILE
\ENDFOR
\end{algorithmic}
\end{algorithm}

\fig{stabilize}  show examples  of \stcb{}s  before and  after motion
compensation for different flying objects. For  each of the \stcb{}s $b$ and for
each patch $m_{k=1..s_t}$  inside $b$ we plot the position  of the actual center
of the flying object with respect to the center of the patch.

\comment{ We can see from these examples that the optical flow approach is more focused on the  background,  as in  the  case  where the  background  is  not uniform,  the positions of the drone over the patches are spread across the patch, rather then located  close to  each other,  as in  the case  of our  regression-based motion compensation.} We can see from these examples that the optical flow approach is more focused on the  background,  as in  the  case  where the  background  is  not uniform,  the positions of the drone over the patches are spread across the patch. However, in  the case  of our  regression-based motion compensation the center positions of the drone are located  close to each other and to the center of the patch. Moreover if  the  appearance  of the  drone  changes inside  the \stcb{} (e.g.  due to the lighting  changes) optical flow based method is unable to correctly estimate the shift of the object.  On the other hand our regression approach is capable of identifying the correct shift even in the situations when the  outlines of  the object  are heavily  corrupted by  noise, coming  from the background. \fig{stabilize} illustrates this fact for different flying objects and various background complexity levels. Note  also that our regressor generalizes well to different objects that were not used for training.

\comment{This  is illustrated  by  the  fact  that  for the  \stcb{}s  with regression-based motion compensation, $d_b(\cdot)$ shows  that part of the patch occupied by the  drone is in general  static and moved to the  center, while the background is allowed to change.} 

Provided  regressors are  estimated, we  use them  for motion compensation of the flying objects inside the \stcb{}s of the training  dataset.  This allows us to train the AdaBoost classifier from~\eqt{AdaBoost}, on the data with  much  less  in-class variation and thus it is easier for the machine learning algorithm to fit a proper model to it.

\begin{figure*}[ht!]
\centering
\begin{tabular}{ccccccc}
\toprule
\multicolumn{7}{c}{Original frames from the video sequences} \\
\includegraphics[height = \imsz]{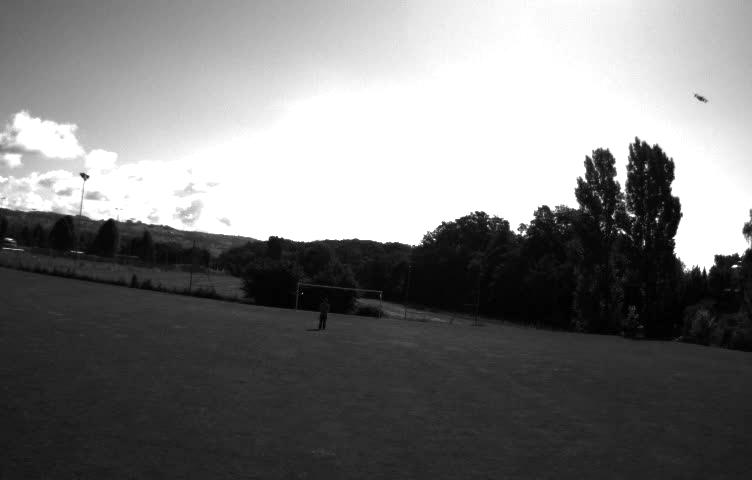} &
\hspace{-0.3cm}\includegraphics[height = \imsz]{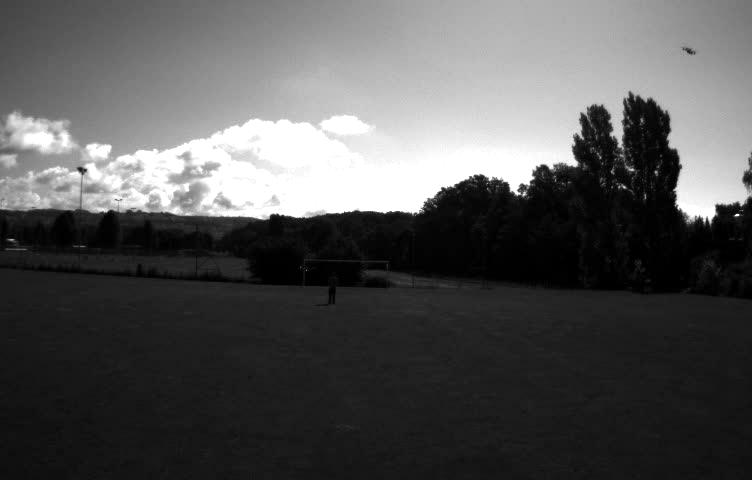} &
\hspace{-0.3cm}\includegraphics[height = \imsz]{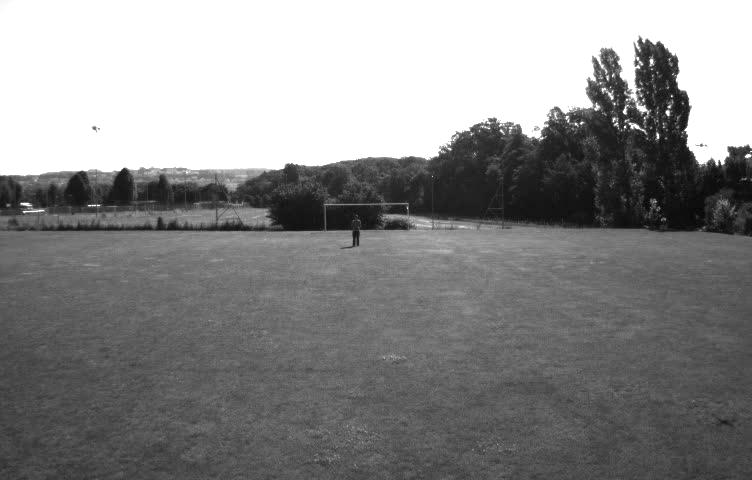} &
& \hfill
\includegraphics[height = \imsz]{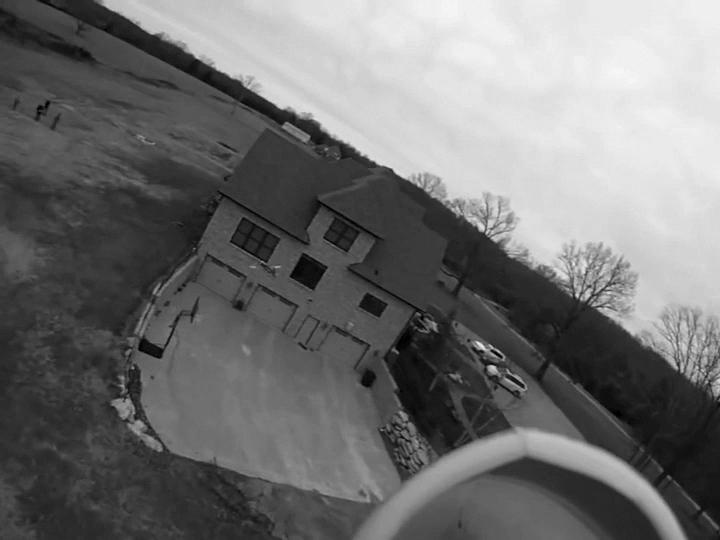} &
\hspace{-0.3cm}\includegraphics[height = \imsz]{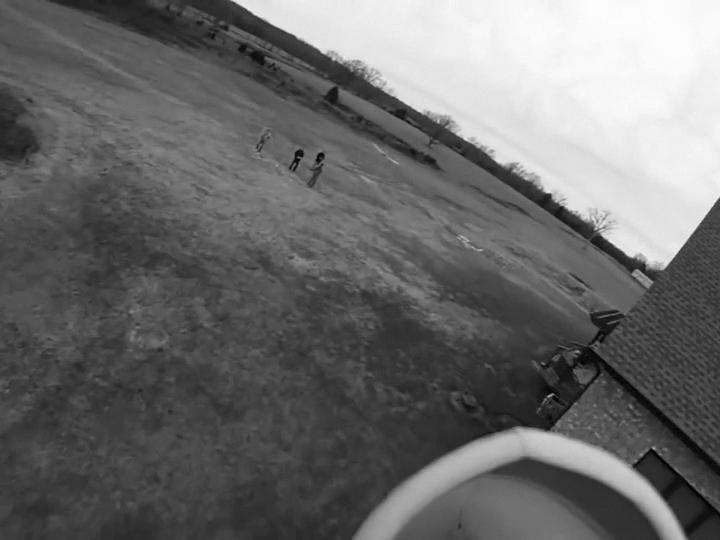} &
\hspace{-0.3cm}\includegraphics[height = \imsz]{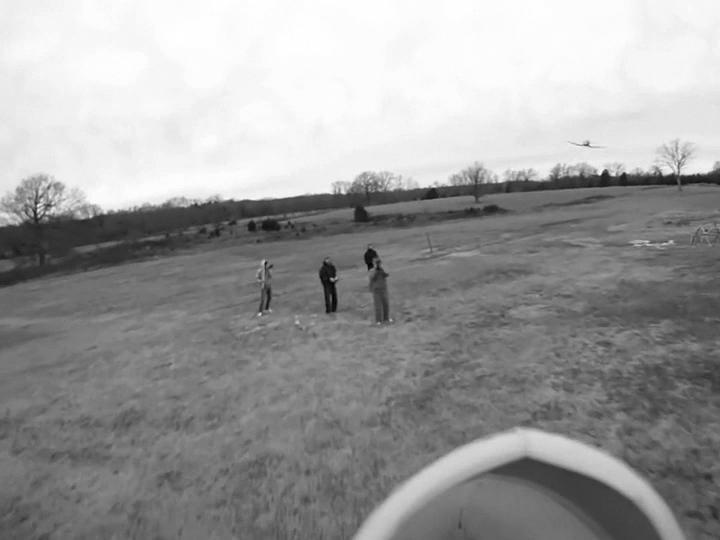} \\
\midrule
\multicolumn{7}{c}{Background subtraction} \\
\includegraphics[height = \imsz]{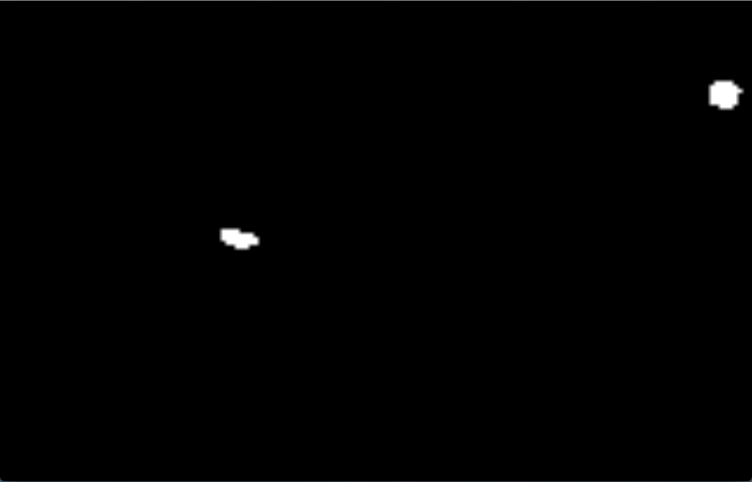} &
  \hspace{-0.3cm}\includegraphics[height = \imsz]{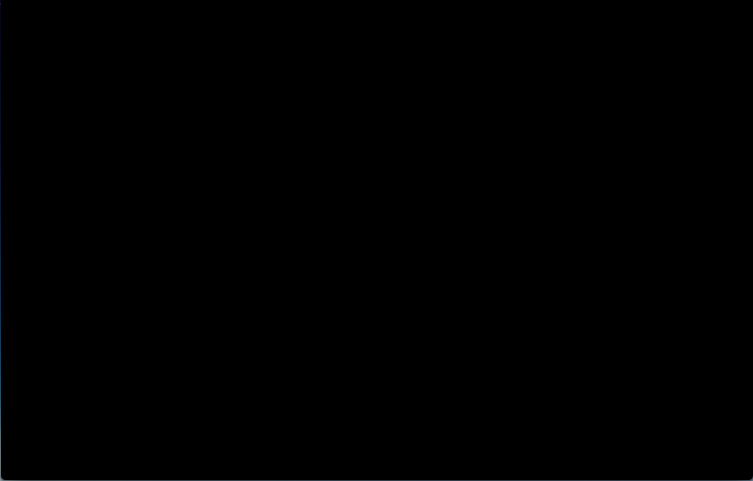} &
  \hspace{-0.3cm}\includegraphics[height = \imsz]{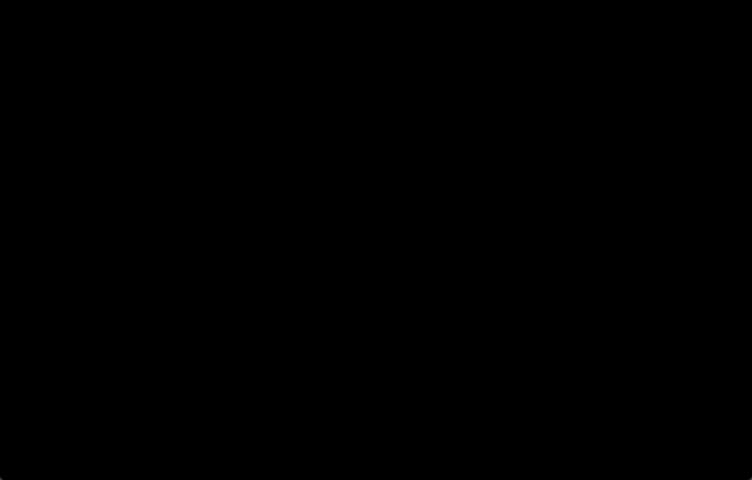} &
  & \hfill
\includegraphics[height = \imsz]{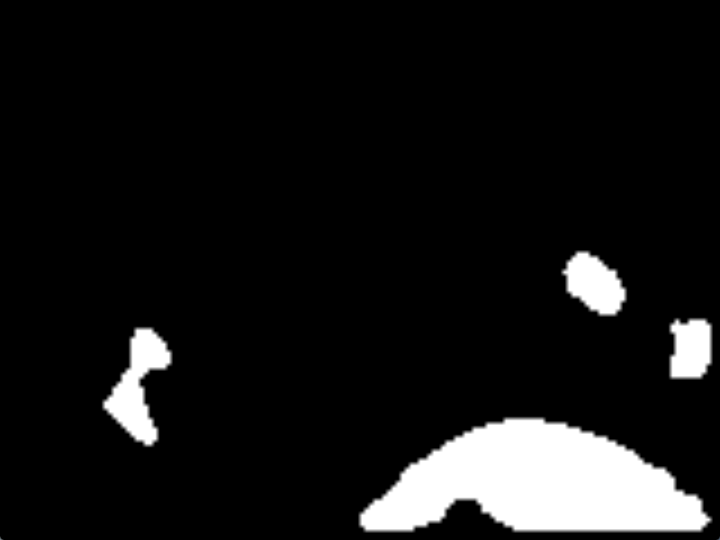} &
  \hspace{-0.3cm}\includegraphics[height = \imsz]{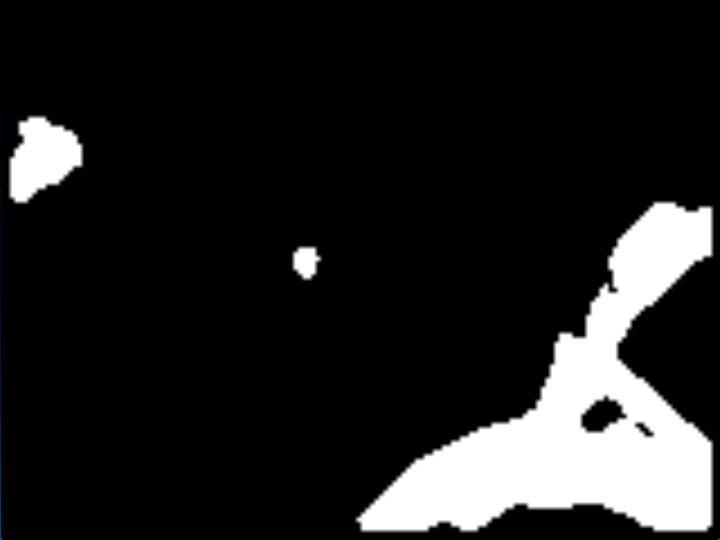} &
  \hspace{-0.3cm}\includegraphics[height = \imsz]{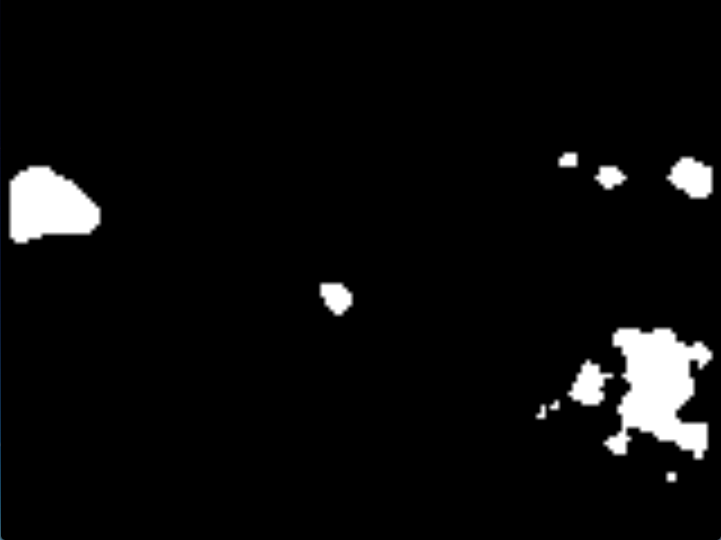} \\
  \midrule
  \multicolumn{7}{c}{Optical flow} \\
  \includegraphics[height = \imsz]{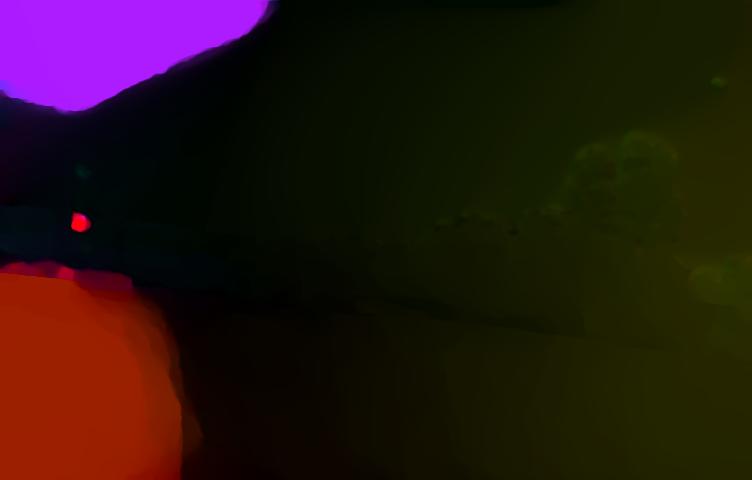} &
  \hspace{-0.3cm}\includegraphics[height = \imsz]{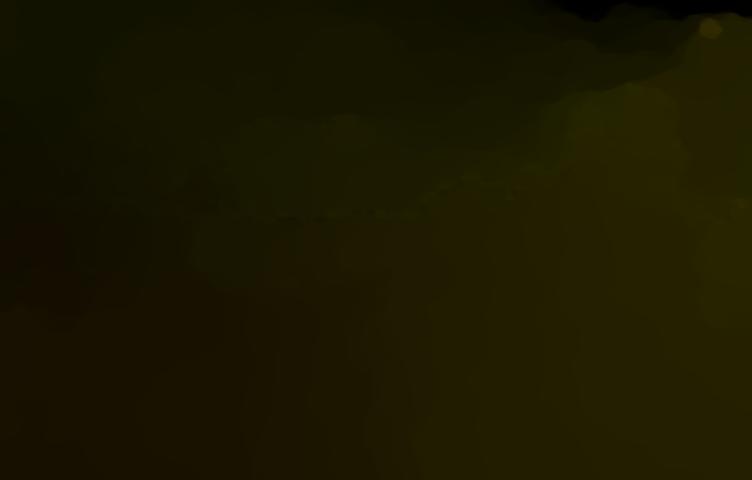} &
  \hspace{-0.3cm}\includegraphics[height = \imsz]{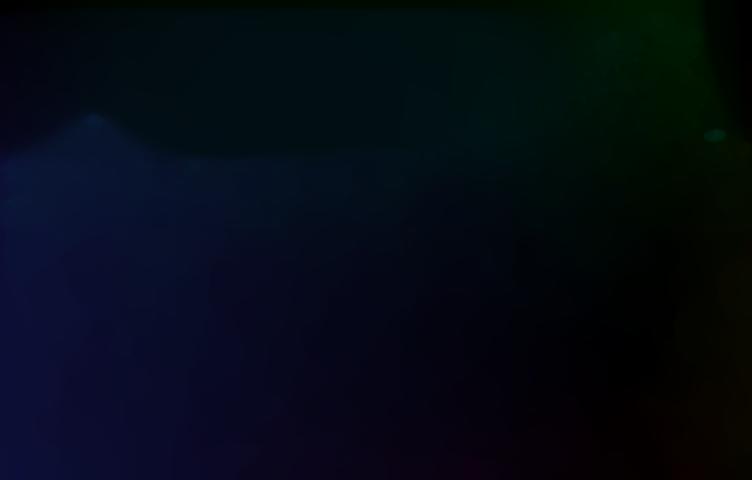} &
  & \hfill
\includegraphics[height = \imsz]{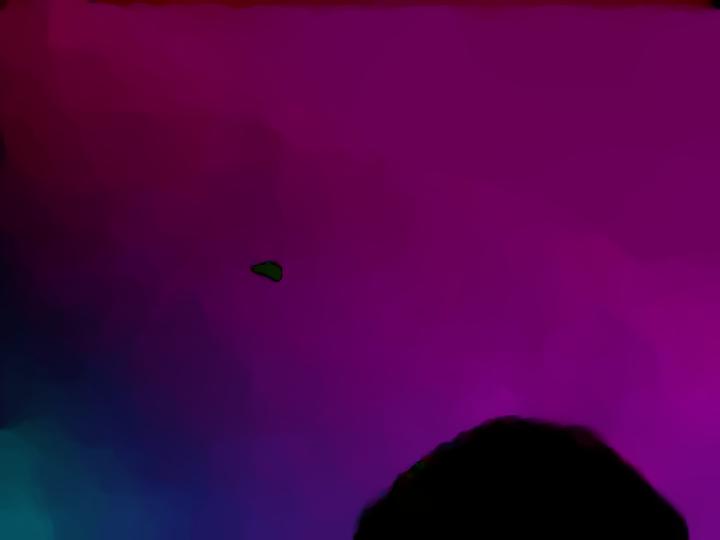} &
  \hspace{-0.3cm}\includegraphics[height = \imsz]{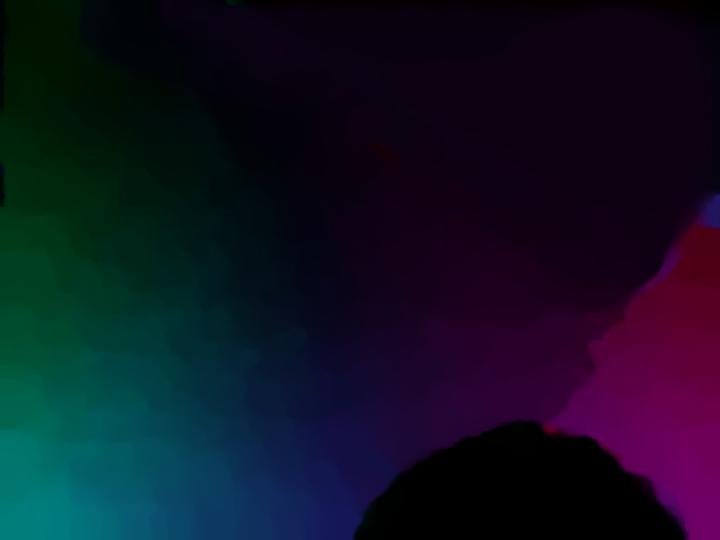} &
  \hspace{-0.3cm}\includegraphics[height = \imsz]{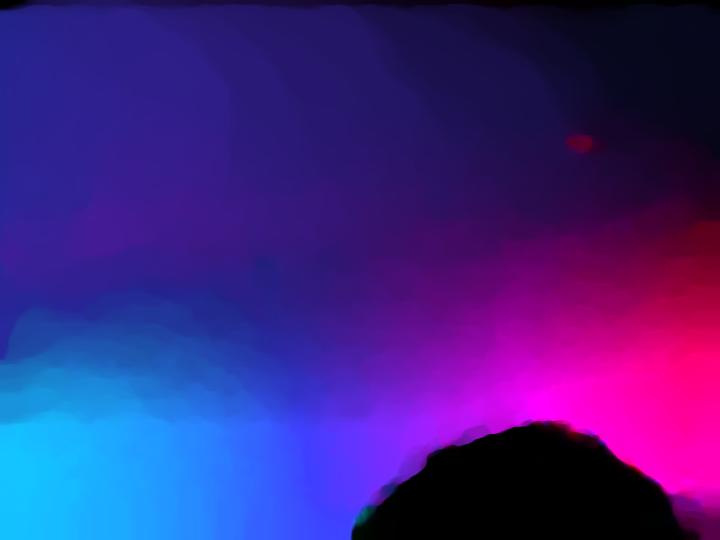} \\
  \midrule
\multicolumn{7}{c}{Our approach} \\
\includegraphics[height = \imsz]{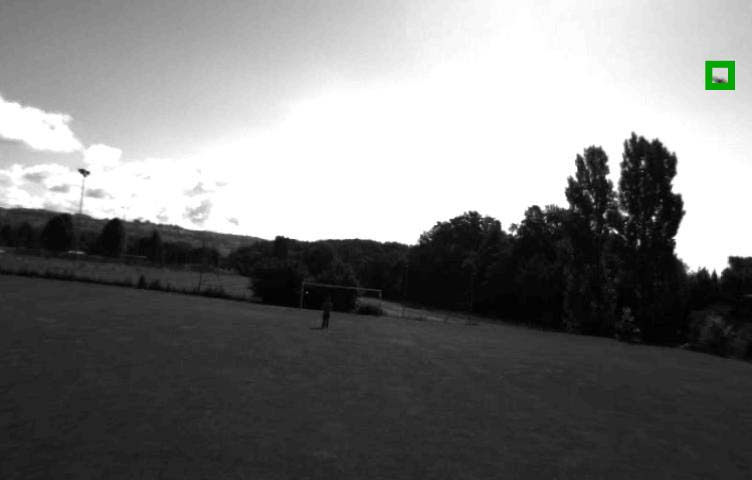} &
  \hspace{-0.3cm}\includegraphics[height = \imsz]{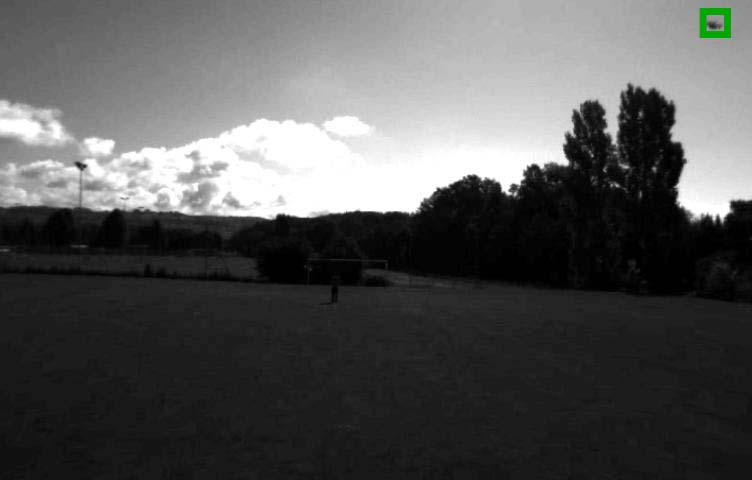} &
  \hspace{-0.3cm}\includegraphics[height = \imsz]{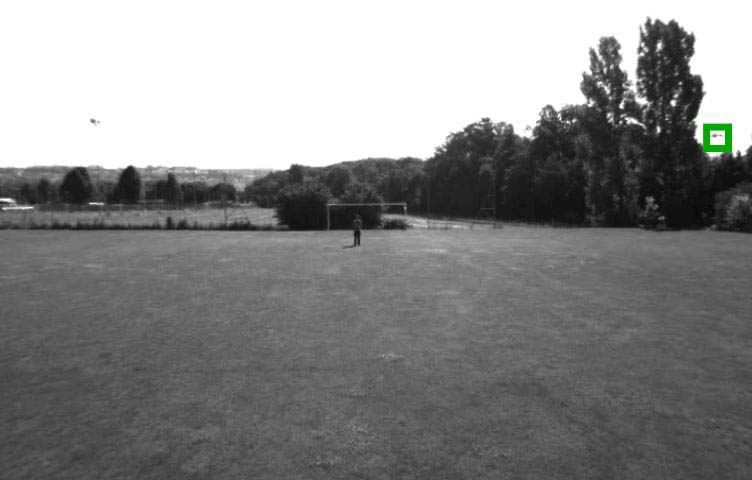} &
  & \hfill
\includegraphics[height = \imsz]{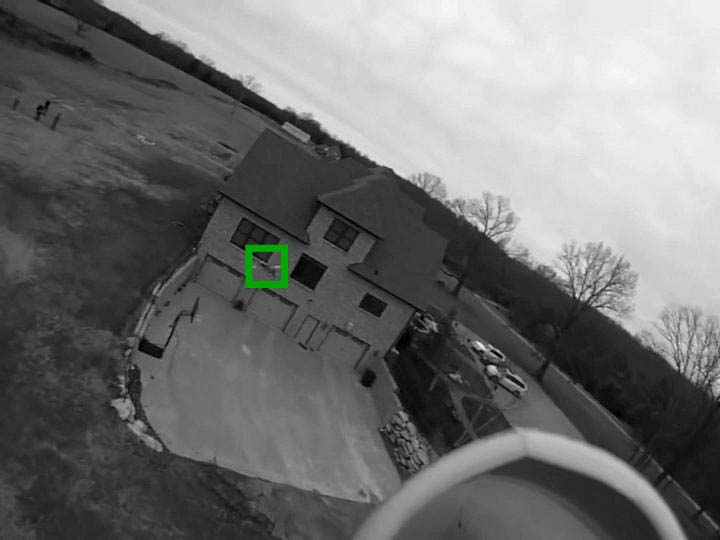} &
  \hspace{-0.3cm}\includegraphics[height = \imsz]{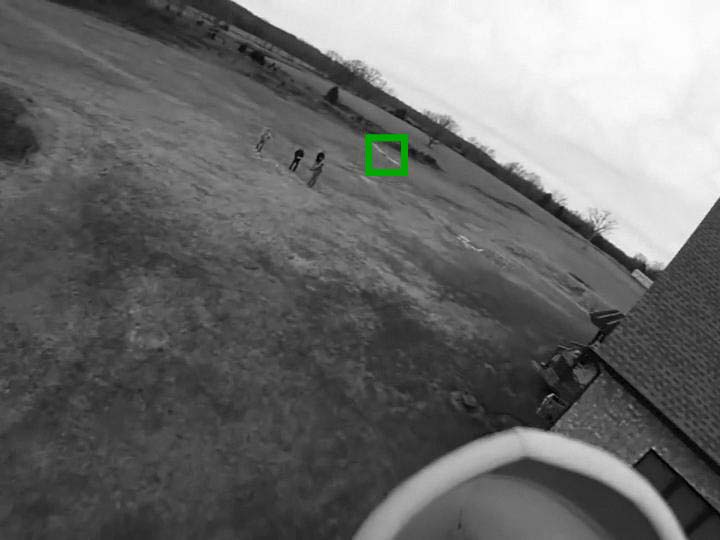} &
  \hspace{-0.3cm}\includegraphics[height = \imsz]{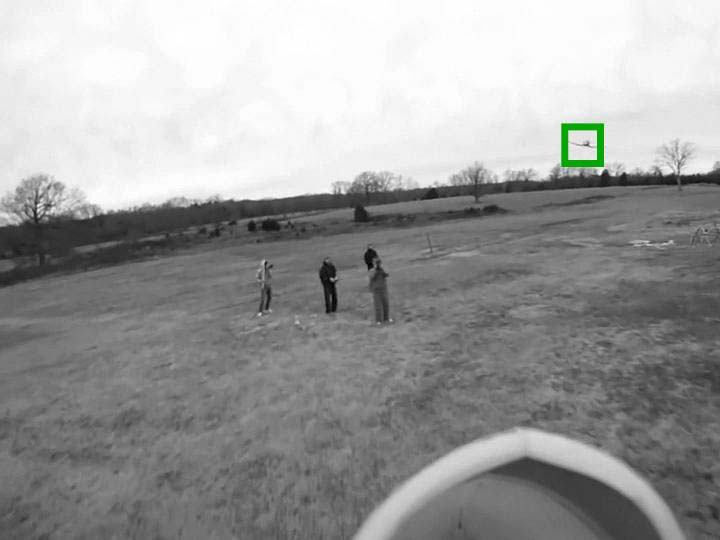} \\ 
  \bottomrule
 \multicolumn{3}{c}{UAV dataset} & \hfill & \multicolumn{3}{c}{Aircraft dataset} \\
\end{tabular}
\mycaption{Comparison of our approach  with motion-based methods. {\bf First row:} Original frames from the video sequences. \confirmed{{\bf
      Second     row:}      Using     a     state-of-the      art     subtraction
    algorithm~\cite{SeungJong12} is not sufficient  to detect the target objects
    as the  camera is moving  and the background can  vary because of  trees and
    grass moving with the wind.  The UAV is detected only in one image, together
    with a  false detection. The  plane is detected in  only one image  as well,
    together with large  errors.  {\bf Third row:} The task is  also very difficult
    for a state-of-the-art  optical flow approach~\cite{Brox11}. The  UAV is not
    revealed in  the optical flow  images, the plane is  visible in only  two of
    them.  {\bf Bottom row:}  Our detector can detect the target  objects by relying
    on motion and appearance. (best seen in color) }}
\label{fig:bg_comp}
\end{figure*}

\section{Results}

In  this  section,   we  evaluate  the  performance  of   our  approach  against
state-of-the-art  ones~\cite{Dollar09b,Park13} on  two  challenging datasets.   They include many real-world challenges such as fast illumination changes and complex backgrounds, created by  moving treetops seen against a changing  sky.  They are as follows:

\begin{figure}
\centering
\begin{tabular}{cccccccc}
\includegraphics[width = \mdsz]{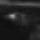} &
\hspace{-0.3cm}\includegraphics[width = \mdsz]{./fig/mc/nmc_res_88_2.png} &
\hspace{-0.3cm}\includegraphics[width = \mdsz]{./fig/mc/nmc_res_285_2.png} &
\hspace{-0.3cm}\includegraphics[width = \mdsz]{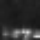} &
\includegraphics[width = \mdsz]{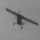} & 
  \hspace{-0.3cm}\includegraphics[width = \mdsz]{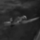} & 
  \hspace{-0.3cm}\includegraphics[width = \mdsz]{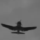} & 
  \hspace{-0.3cm}\includegraphics[width = \mdsz]{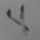} \\
\includegraphics[width = \mdsz]{./fig/mc/nmc_res_802_2.png} &
\hspace{-0.3cm}\includegraphics[width = \mdsz]{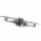} &
\hspace{-0.3cm}\includegraphics[width = \mdsz]{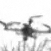} &
\hspace{-0.3cm}\includegraphics[width = \mdsz]{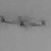} &
\includegraphics[width = \mdsz]{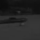} & 
  \hspace{-0.3cm}\includegraphics[width = \mdsz]{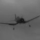} & 
  \hspace{-0.3cm}\includegraphics[width = \mdsz]{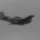} & 
  \hspace{-0.3cm}\includegraphics[width = \mdsz]{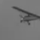} \\
\multicolumn{4}{c}{(a) UAV dataset} & \multicolumn{4}{c}{(b) Aircraft dataset} \\
\end{tabular}
\vspace{-0.2cm}
\mycaption{Sample image windows containing aircrafts or UAVs from our datasets.}
\label{fig:drone_db}
\end{figure}

\begin{itemize}[topsep=3pt, partopsep=5pt]
	\setlength\itemsep{5pt}
	\setlength{\parskip}{0pt}
        
\item {\bf UAV dataset} It comprises 20 video sequences of 4000 $752 \times 480$
  frames each  on average.  They  were acquired by a  camera mounted on  a drone
  filming   similar  ones   while  flying   outdoors.    As  can   be  seen   in
  \fig{drone_db}(a),  there appearance  is  extremely variable  due to  changing
  attitudes and lighting conditions.
  
\item {\bf Aircraft dataset} It consists of 20 YouTube videos of planes or radio
  controlled plane-like drones.  Some videos  were acquired by a camera
    on the  ground and the rest  was filmed by  a camera on board of an aircraft.
  These  videos vary  in length  from  hundreds to  thousands of  frames and  in
  resolution  from $640  \times 480$  to $1280  \times 720$.   \fig{drone_db}(b)
  depicts the variety of plane types that can be seen in them.
    
\end{itemize}
\noindent
\artem{We will make these datasets, together with the ground-truth annotations, publicly available as a new challenging benchmark for aerial objects detection and visual-guided collision avoidance.}

\comment{
We further split the Aircraft dataset  into two separate subsets. The first one,
mostly includes sequences  filmed by people standing on the  ground and features
the planes seen against the sky,  which means that the background changes little
from one frame to the next. The second  one is made of sequences filmed from the
on-board camera  of another aircraft, which  means that the backgrounds  are far
more complex and can  change abruptly from one frame to the  next. We will refer
to  the first  as  {\it Simple  Aircraft}  and  to the  second  as {\it  Complex
  Aircraft} datasets.}

\comment{\note{It's not  clear if  you trained  separate classifiers  for the  simple and
  complex aircrafts. Did you?}
  
\ans{Yes, I have trained separate classifiers for simple and complex datasets}
}

\subsection{Training and Testing}

In  all  cases  we used  half  of  the  data  to  train both  the  regressor  of
Eq.~\ref{eq:quadratic_loss}  and  the  classifier of  Eq.~\ref{eq:AdaBoost}.  We
manually supplied 8000 bounding boxes centered on a UAV and 4000 on a plane.

\comment{\note{I imagine  that you  had to train  the regressor first  so that  you could
  correct for motion and only then train  the classifier. Is that so? If yes, it
  should be clearly explained in the method section.}
  
  \ans{Yes }
}
\myparagraph{Training the Regressors}

To provide labeled examples,  where the aircraft or UAV is not  in the center of
the  patch but  still at  least  partially within  it, we  randomly shifted  the
manually supplied bounding boxes by distances of up to half of their size.  This step is repeated for every second frame of the training database to cover the variety of shapes and backgrounds in front of which the aircraft might appear.

\comment{\note{If  you train  on $40  \times 40$  patches, how  do you  handle the  scale
  problem? Can you find objects that are smaller or larger?}}

The apparent size of  the objects in the UAV and Aircraft  datasets vary from 10
to 100 pixels on the image plane. To train the regressor, we used $40 \times 40$
patches containing the UAV or aircraft  shifted from the center.  We have chosen
this  size because  smaller ones  will result  in fewer  features available  for
gradient boosting, while  bigger ones will introduce noise and  take more time to
analyze. We detect the targets at different scales by running the
  detector on the image at different resolutions.

\myparagraph{Training the Classifiers}

We used the \stcb{}s of size $(s_x,s_y,s_z) = (40,40,4)$, the spatial dimensions
being  the same  as  for regression.   The  choice  of $s_z  =  4$ represents  a
compromise between being able to detect far away objects by increasing $s_z$ and
closer ones that require a smaller $s_z$ because the frame-to-frame motion might
be too big for our motion-compensation mechanism.

\myparagraph{Evaluation Metric}
We report  precision-recall curves. Precision is  computed as the number of true positives, detected by  the algorithm divided by the total number of detections. Recall is the number of true positives  divided by the number of the positively  labeled test examples.  Additionally we use  the {\it Average Precision} (AveP) measure, which we take to be the integral $\int_0^1p(r)dr$, where $p$ is the precision, and $r$ the recall.

\comment{
\note{You should really give AveP numbers for everything.}}

\subsection{Baselines}

To  demonstrate  the  effectiveness  of  our approach,  we  compare  it  against
state-of-the-art algorithms.  We chose  them to be  representative of  the three
different ways the problem of detecting  small moving objects can be approached,
as discussed in Section~\ref{sec:related}.

\begin{itemize}[topsep=3pt, partopsep=5pt]
	\setlength\itemsep{5pt}
	\setlength{\parskip}{0pt}

  \item {\bf Appearance-Based  Approaches} that rely on  detection in individual
    frames.       We     will      compare      against     Deformable      Part
    Models~(DPMs)~\cite{Felzenszwalb10},           Convolutional          Neural
    Networks~(CNN)~\cite{Serre07}, Random Forests~\cite{Breiman01}, and \comment{Boosting
    approaches using} Aggregate Channel Features method~(ACF)~\cite{Dollar09b}, the latter being widely considered to be among the best.
    Since our  algorithm labels  \stcb{}s as  positive or  negative, for  a fair
    comparison with  these single frame  algorithms, we proceed as  follows.  If
    they label the middle frame of the  cube as positive, then the whole \stcb{}
    is regarded as a positive detection and otherwise not.

  \item {\bf Motion-based Approaches} do  not use any appearance information and
    rely purely on the correct estimation  of the background motion. Among those
    we          experimented          with          MultiCue          background
    subtraction~\cite{SeungJong12,bgslibrary13}  and large  displacement optical
    flow~\cite{Brox11}.

  \item {\bf  Hybrid approaches} are closest  in spirit to ours  and correct for
    motion using image-flow. Among those,  the one presented in~\cite{Park13} is
    the most recent one we know of and the one we compare against. To ensure
    fair  comparison, we used the same size \stcb{}s for both.

\end{itemize}

\begin{figure}
\centering
\begin{tabular}{cc}
\includegraphics[width = \halfimsz]{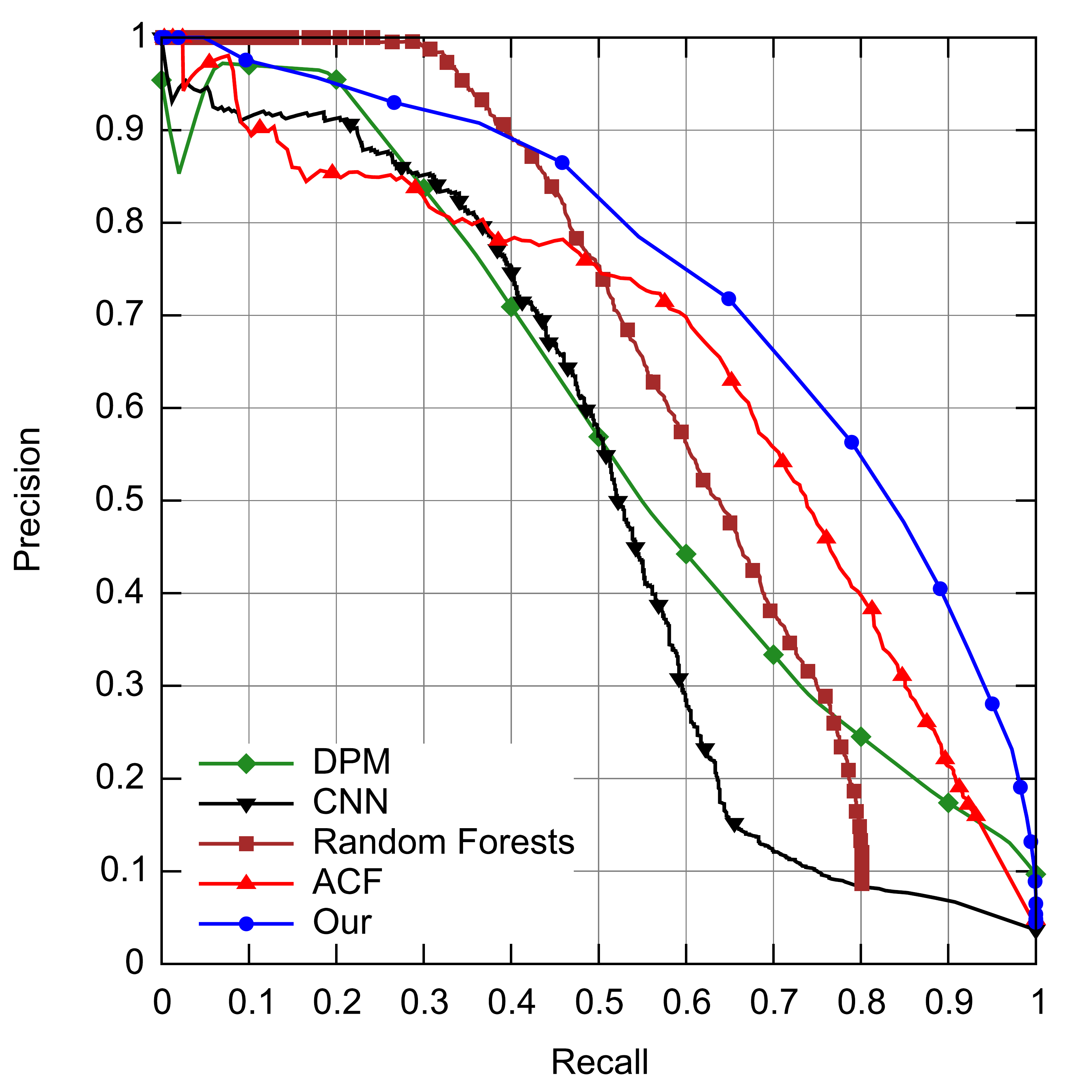} &
\hspace{-0.3cm}\includegraphics[width = \halfimsz]{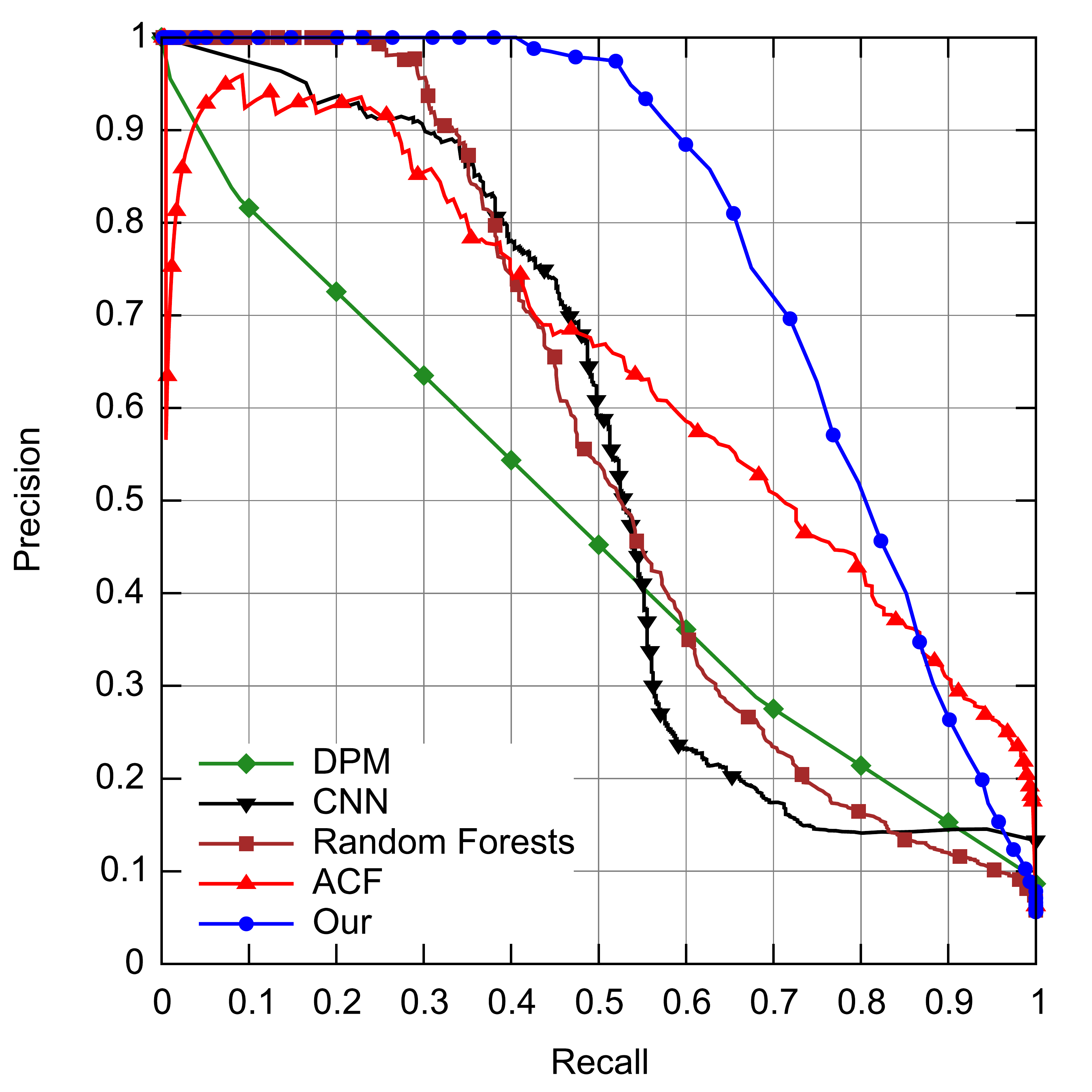}\\
UAV dataset & Aircraft dataset \\
\end{tabular}
\mycaption{Comparison against apperance-based approaches.   For both the UAV and
  Aircraft  datasets,  our   approach  achieves  about  a   $10\%$  increase  of
  performance compared to the state-of-the-art ACF method.}
\label{fig:comparison_vs_single}
\end{figure}

\begin{table}
\centering
\begin{tabular}{lcc}
\toprule
& \multicolumn{2}{c}{Average Precision} \\
\cmidrule(r){2-3} 
Method & UAV dataset & Aircraft dataset \\
\midrule
DPM~\cite{Felzenszwalb10} & 0.573 & 0.470 \\
CNN~\cite{Serre07} & 0.504 & 0.547 \\
Random Forests~\cite{Breiman01} & 0.618 & 0.563 \\
ACF~\cite{Dollar09b} & 0.652 & 0.648 \\
\Stcb{}s without & \multirow{2}{*}{0.485} & \multirow{2}{*}{0.497} \\
motion compensation &  & \\
\Stcb{}s+optical flow & 0.540 & 0.652 \\
Park~\cite{Park13} & 0.568 & 0.705 \\
Our & { \bf 0.751} & {\bf 0.789} \\
\bottomrule
\end{tabular}
\mycaption{Average precision of detection methods on our datasets. We can see that in both cases our approach with regression-based motion compensation is able to outperform both purely appearance based methods and state-of-the-art hybrid approach.}
\label{tbl:average_precision}
\end{table}

\subsection{Evaluation against Competing Approaches}

Here  we compare  our  regression-based approach  against the three
classes of methods discussed above.
\comment{
\note{Is this $10\%$ a measure of AveP? Is so, say it and put the full tables in
  the figure. When we talk about aircraft are we talking about the simple of the
  complex ones? Or both?}}


\myparagraph{Appearance-Based    Methods.}

\fig{comparison_vs_single} compares our  method with  appearance-based approaches
  on  our  two  datasets.    \tbl{average_precision}  summarizes the results in terms of Average Precision. For both the UAV  and Aircraft datasets we
  can  achieve on  average  around  $10\%$ improvement,  in terms of this
  measure, over the ACF method, which itself outperforms the others. The DPM and CNN
  methods perform  the worst on  average. Most likely, this happens because the
  first one depends  on using the correct size  of the bins for  HoG estimation, which  makes it  hard to  generalize for a  large variety of  flying objects  and the second one requires much more training samples than our detector does.

\comment{\ans{\fig{bg_comp} is the way to compare our method with motion-based one,
  because it is not clear how to make a precision-recall curve for them, still
  we need to provide some comparison I think. \vincent{For me, it's obvious that
    motion-based methods are not possible here. I changed the caption of Fig4,
    but we can also remove it}. }}

\myparagraph{Motion-Based Methods.}  \fig{bg_comp} shows  that state-of-the-art
background       subtraction~\cite{SeungJong12}      and       optical      flow
computation~\cite{Brox11} do not  work well enough for detecting  UAVs or planes
in the challenging conditions that we consider.

We  do  not  provide  precision-recall curves  for  motion-based  methods
  because it it not  clear how big the moving part of the  frame should be to be
  considered as  an aircraft.  We  have tested  several potential sizes  and the
  average precision was much lower  than those in \tbl{average_precision} in all
  cases. \comment{We will provide these figures as a supplementary material.}

\myparagraph{Motion compensation approaches.}

\begin{figure}
\centering
\begin{tabular}{cc}
\includegraphics[width = \halfimsz]{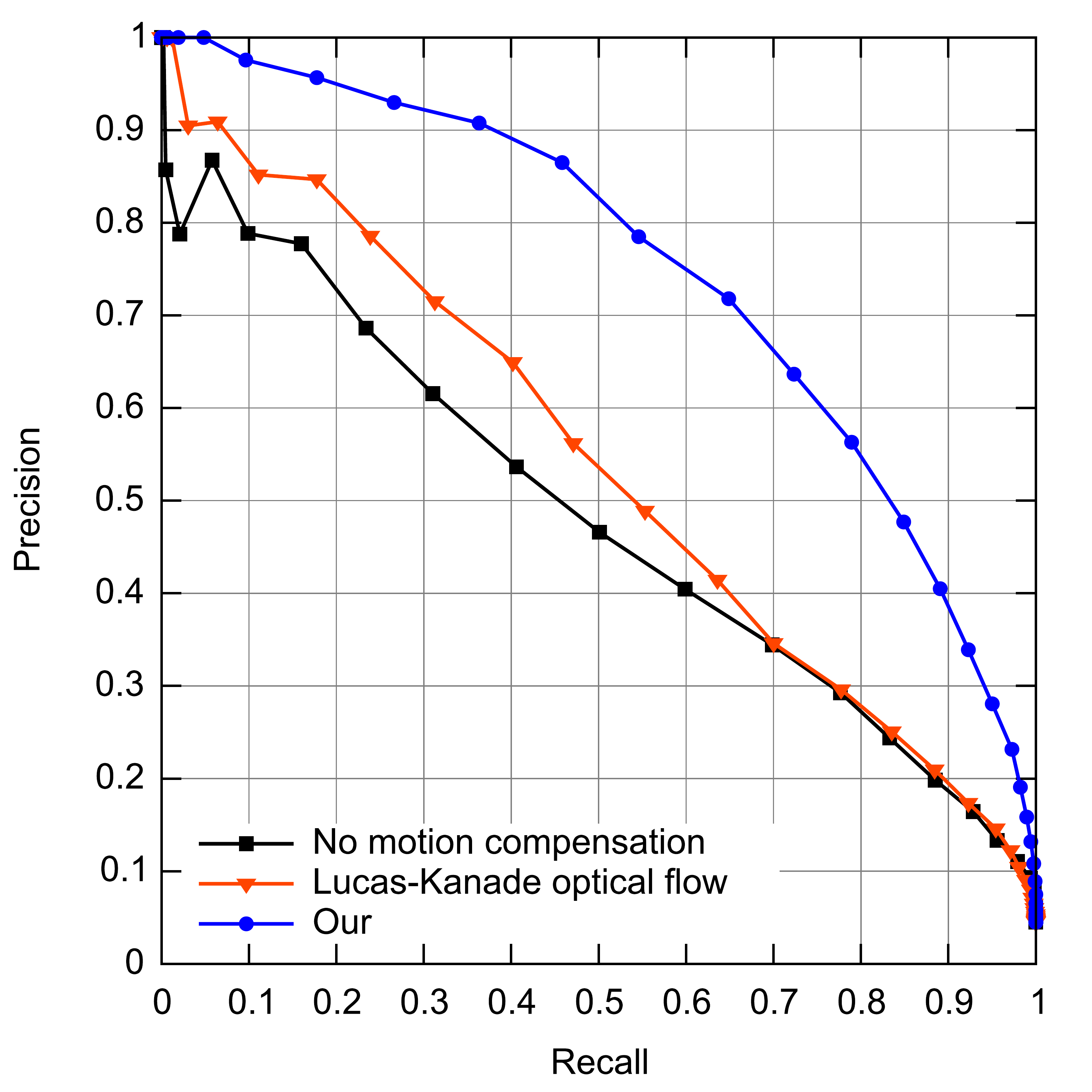} &
\hspace{-0.3cm}\includegraphics[width = \halfimsz]{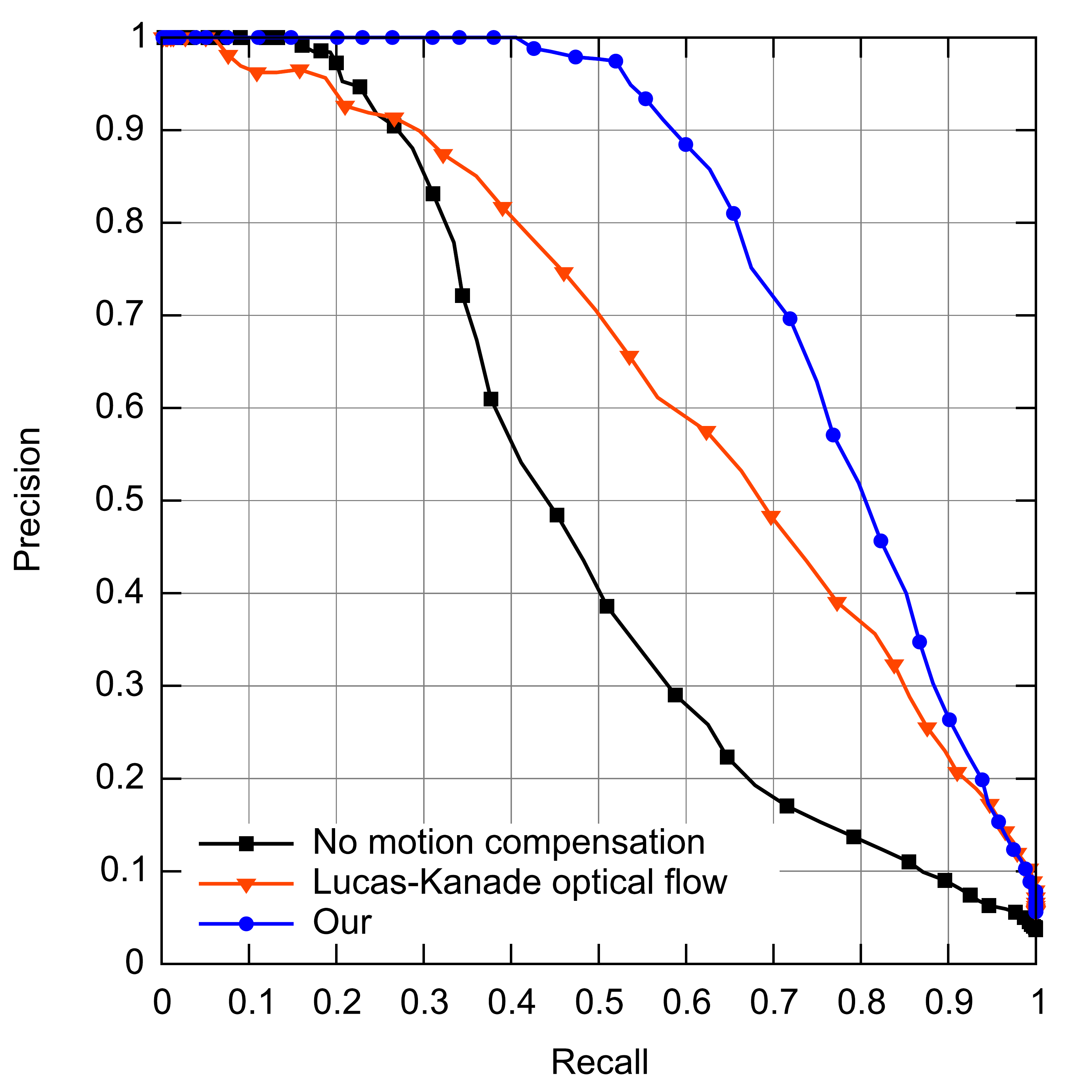} \\
(a) UAV dataset & (b) Aircraft dataset \\
\end{tabular}
\mycaption{Evaluation of the motion compensation methods on our datasets. Unlike other motion compensation algorithms, our regression-based method is able to properly identify the shift in object position and correct for it, even in the situation, when the background is complex and the outlines of the object are barely visible, which leads to significant improvement in the detection accuracy.}
\label{fig:dr_mc}
\end{figure}

\comment{We first compare here our regression-based motion compensation approach with the optical flow one. Then we present the comparison with the recent hybrid approach for object detection~\cite{Park13}.}
\fig{dr_mc} compares our motion compensation algorithm with
  the optical flow-based one used in~\cite{Park13} for both UAV and
  Aircraft datasets.
Using motion compensation for alignment of the \stcb{}s results into
higher performance of the detectors, as the in-class variation of the
data is decreased. \tbl{average_precision} shows that we can achieve
at least $15\%$ improvement in average precision on both datasets
using our motion compensation algorithm.


Among the  motion compensation  approaches our  regression-based method
  outperforms the optical flow-based one of~\cite{Park13}, because it is able to
  correctly  compensate  for  the  mUAV  motion even  in  the  cases  where  the
  background is  complex and the  drone might not be  visible even to  the human
  eye.  \fig{stabilize}(b,d)  illustrates this  hard situation with  an example.
  On the  contrary, the optical flow  method is more focused  on the background,
  which decreases  its performance.   \fig{stabilize}(b) shows  an example  of a
  relatively  easy situation,  when the  aircraft  is clearly  visible, but  the
  optical flow  algorithm fails to correctly  compensate for its shift  from the
  center, while our regression-based approach succeeds.



Our regression-based motion compensation algorithm allows us to significantly reduce the in-class variation of the data, which results into $30\%$ boost in performance, as given by the Average precision measure.

\myparagraph{Hybrid approaches.}

\begin{figure}
\centering
\begin{tabular}{cc}
\includegraphics[width = \halfimsz]{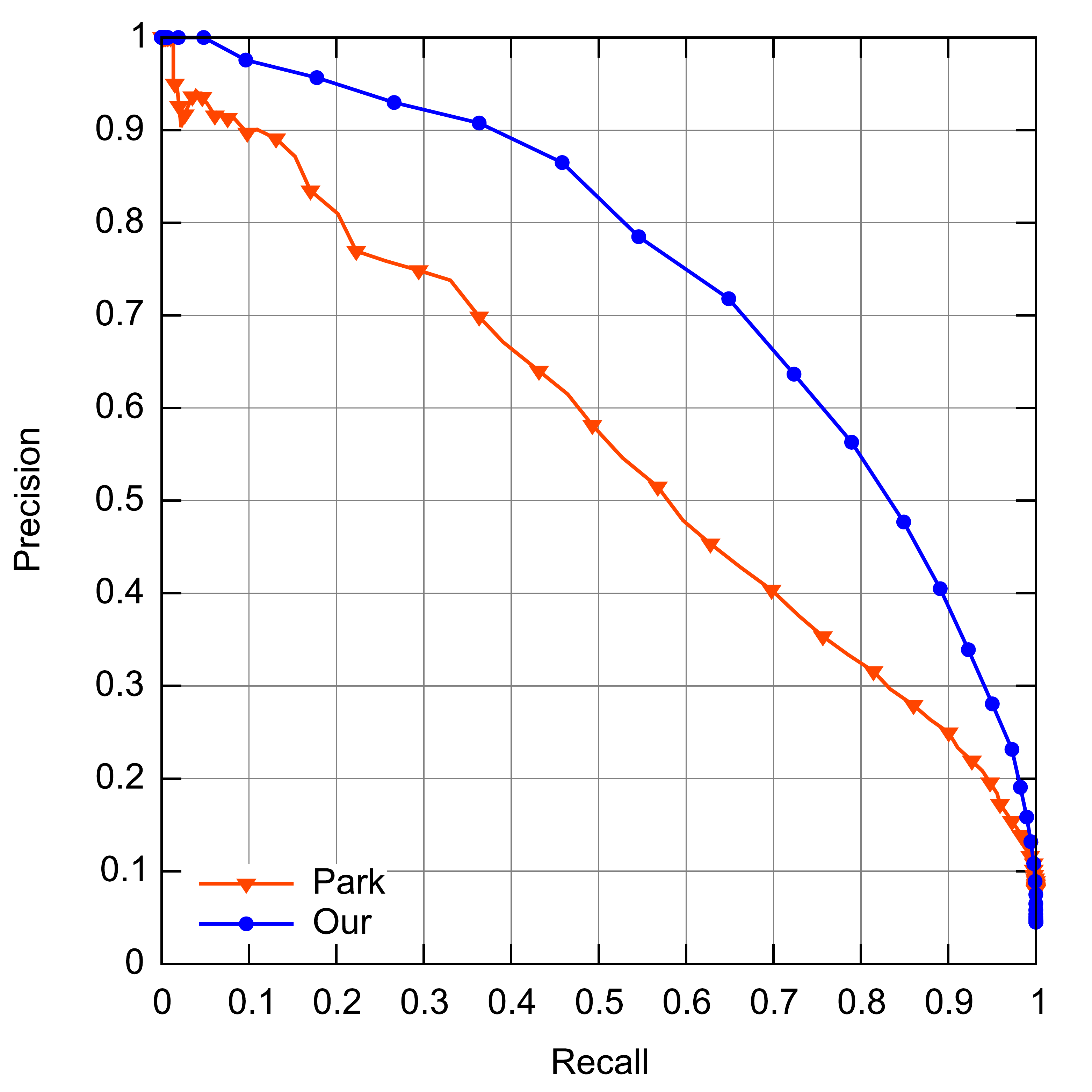} &
\hspace{-0.3cm}\includegraphics[width = \halfimsz]{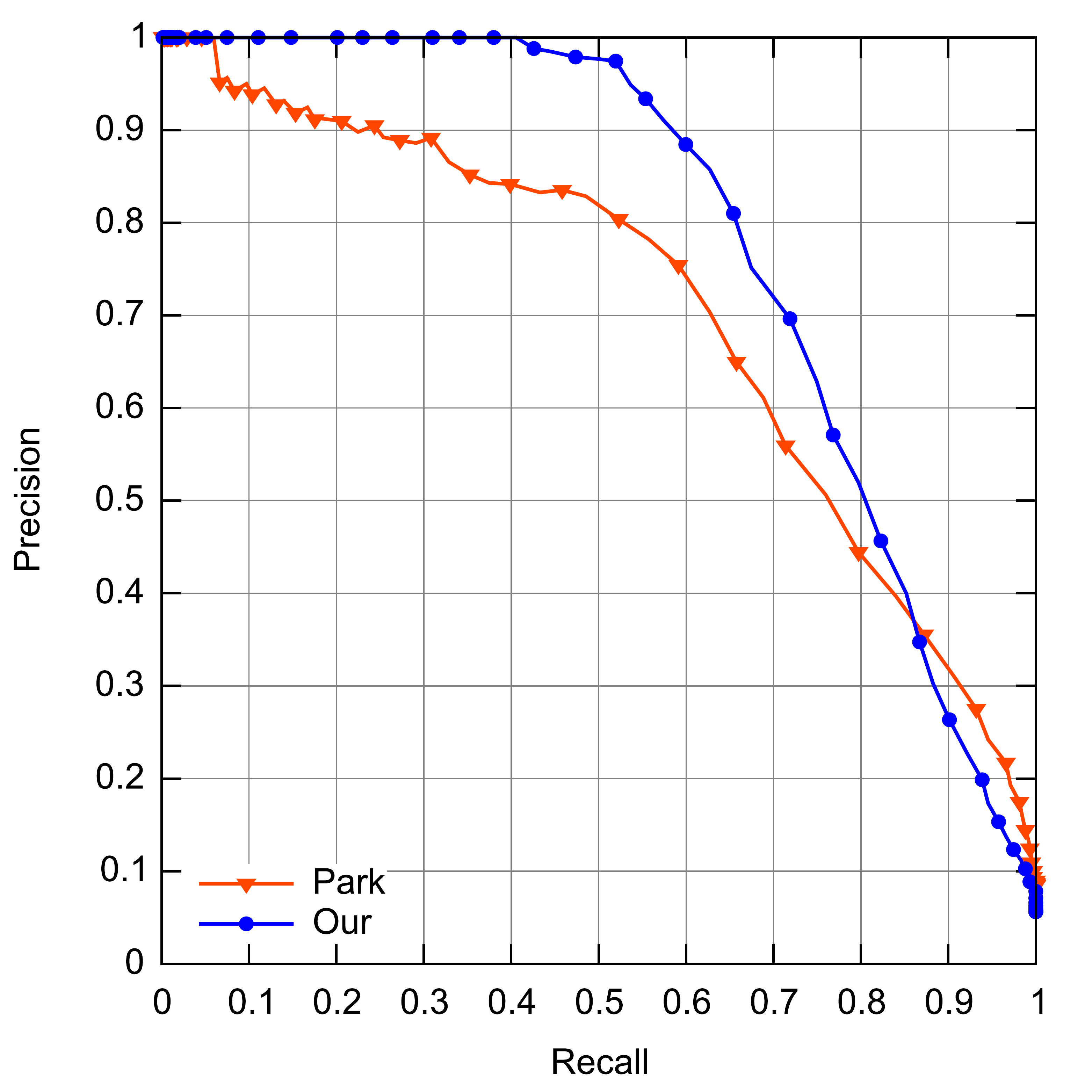} \\
(a) UAV dataset & (b) Aircraft dataset\\
\end{tabular}
\mycaption{Comparison of our approach to the hybrid method (Park). Our method is able to show higher performance for both of the datasets, due to the regression-based motion compensation algorithm used.}
\label{fig:park_comp}
\end{figure}

\fig{park_comp} illustrates the comparison of our method to the hybrid approach~\cite{Park13}, which relies on motion compensation using Lucas-Kanade optical flow method, and yields state-of-the-art performance for pedestrian detection. For both UAV and Aircraft datasets our method is able to achieve higher performance, due to our regression-based approach for compensating motion that allows to properly identify and correct for the shift of the aircraft inside the block of patches, used for detection.

\subsection{Collision Courses}
\label{sec:collision}

\begin{figure}
\centering
\begin{tabular}{cc}
  \hspace{-0.3cm}\includegraphics[height = 2.2cm]{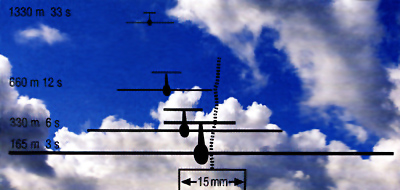}&
  \hspace{-0.3cm}\includegraphics[height = 2.2cm]{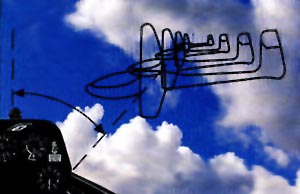}
\end{tabular}
\mycaption{Collision courses.  (Left) The apparent size of a standard glider and
  its 15~m wingspan  flying towards another aircraft at a  relatively slow speed
  (100 km/h) \confirmed{is very small 33s before impact, but the glider completely
    fills  the field  of  view only  half  a minute  later,  3s before  impact.}
  (Right) An aircraft on a collision course  is seen in a constant direction but
  its apparent size grows, slowly at first and then faster.}
\label{fig:glider}
\end{figure}

\comment{fills the field   of  view   3s  before  impact   but  still   seems  quite  small   33s  before  impact. }

Detecting  another aircraft  on a  potential  collision course  is an  important
sub-case of the more generic detection  problem we are addressing in this paper.
As shown in Fig.~\ref{fig:glider}(b), the hallmark of a collision course is that
the object on such a course is always seen at a constant angle and that its size
increases slowly,  at least at  first.

This means that motion stabilization is less important in this case and that the
temporal gradients have  a specific distribution.  In other  words, the in-class
variation for the positive examples should be much smaller in this scenario than
in  the  general  case  and  could  be potentially  be  captured  by  a  3D  HoG
descriptor~\cite{Weinland10}.   This gives  us a  good way  to test  whether our
motion-stabilization mechanism  negatively impacts performance in  this specific
case, as do most mechanisms that  enforce invariance when such invariance is not
required.

\begin{figure}
\centering
\begin{tabular}{c}
\includegraphics[width = 2.0in]{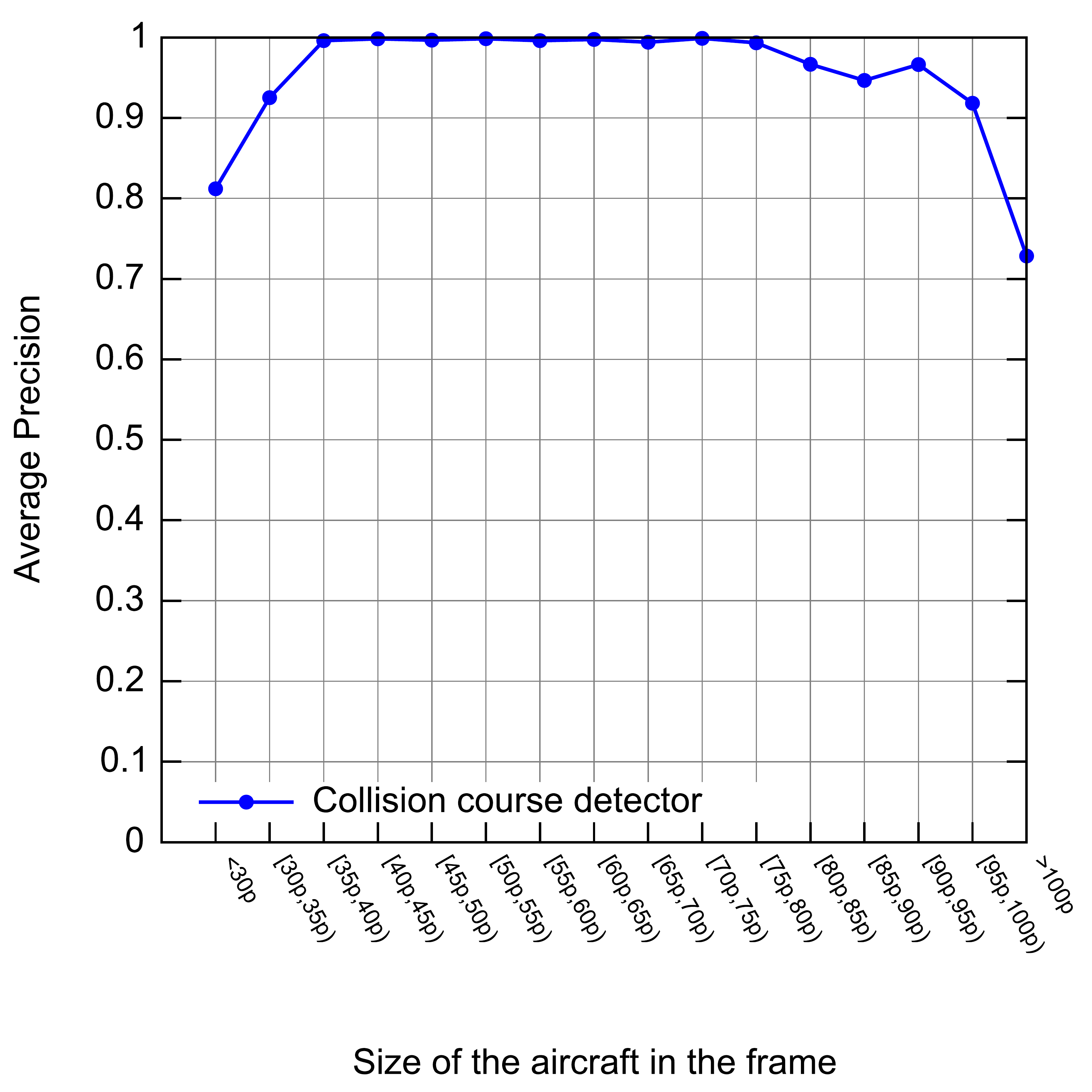} \\
\begin{tabular}{lc}
	\toprule
	& AveP \vspace{-0.5em}\\
	\scriptsize{Detector:} &  \scriptsize{(Average Precision)} \\
	3D HOG & 0.907 \\
	Our    & 0.904 \\
	\bottomrule
\end{tabular}\\
\end{tabular}
\caption{Performance for aircrafts on a  collision course. (Top) Distribution of
  the average precision we can achieve as a function of the size of the aircraft
  in the video  frame. It is close to  100\% for sizes between 35  pixels and 75
  pixels,  which  translates  to  a  useful range  of  distances  for  collision
  avoidance purposes. (Bottom)  The Average Precision of our  method compared to
  using a 3D HOG detector.}
\label{fig:Collision_course}
\end{figure}

To this end, we therefore searched YouTube for a set of video sequences in which
airplanes appear to be on a collision  course for substantial amount of time. We
selected 14 videos that vary in length  from tens to several hundreds of frames.
As before, we used  half of them for training the  collision course detector and
the other to test it.  In \fig{Collision_course}, we compare our results against
those    obtained     using    classification     based    on    a     3D    HoG
descriptor~\cite{Weinland10}  without motion  compensation, as  suggested above.
This corresponds to the method  labeled ``St-cubes without motion compensation''
in \tbl{average_precision}.   As expected, even  though it did not  perform very
well in  the general case, it  turns out to  be very effective in  this specific
scenario.   Our approach  is  very  slightly less  precise,  which reflects  the
phenomenon discussed above.

Furthermore, the  curve at the  top of  \fig{Collision_course} shows that  it is
only when the aircraft is either very small in the image ($< 30$ pixels) or very
close that  the average precision  of our  detector slightly decreases.   In the
first case, this happens  because the object is too far and  the increase of its
apparent size is hardly perceptible. In  the second case, the appearance changes
very   significantly   for   different   types   of   aircrafts,   which   harms
performance. However the goal of a  collision avoidance system is to avoid these
kinds of situations and  to detect the aircraft at a safe  distance.  We can see
that our  approach allows us  to achieve close  to $100\%$ performance  within a
large range and could therefore be used for this purpose.

\section{Conclusion}

We showed that  temporal information from a sequence of  frames plays a vital role in  detection  of   small  fast  moving  objects like UAVs or aircrafts in   complex  outdoor environments.   We therefore  developed  an  object-centric motion  compensation approach that is robust to changes of the appearances of both the object and the background. This approach allows us to outperform state-of-the-art techniques on two challenging datasets. Motion  information  provided  by  our method  has  a  variety  of applications, from detection of potential collision situations to improvement of vision-guided tracking algorithms.

\artem{We collected two challenging datasets for UAVs and Aircrafts. These datasets can be used as a new benchmark for flying objects detection and visual-based aerial collision avoidance.}
{\small
\bibliographystyle{ieee}
\bibliography{string,vision,learning,misc,biomed}
}

\end{document}